\documentclass[10pt,twocolumn,letterpaper]{article}
\usepackage[T1]{fontenc}
\usepackage{lmodern}
\usepackage[letterpaper,margin=0.8in,columnsep=0.3in]{geometry}
\usepackage{graphicx}
\usepackage{amsmath,amssymb}
\usepackage{booktabs}
\usepackage[font=small,labelfont=bf]{caption}
\usepackage[font=footnotesize,skip=3pt,subrefformat=parens]{subcaption}
\usepackage[dvipsnames]{xcolor}
\usepackage{xspace}
\usepackage[numbers,sort&compress]{natbib}
\usepackage{dsfont}
\usepackage[export]{adjustbox}
\usepackage{listings}
\usepackage{multirow}
\usepackage{dblfloatfix}
\usepackage{placeins}
\usepackage[hyphens]{url}
\definecolor{linkblue}{rgb}{0.15,0.35,0.55}
\usepackage[breaklinks,colorlinks,allcolors=linkblue]{hyperref}
\hypersetup{pdftitle={Region Aware Squeeze-and-Excitation Networks for Image Segmentation},
  pdfauthor={Jack Roberts, Praneeth Tummala, Kavi Sarna, Jeova Farias}}

\newcommand{\codefigsize}{\fontsize{5.8pt}{6.5pt}\selectfont}

\lstdefinestyle{pytorchcode}{
    language=Python,
    basicstyle=\ttfamily\codefigsize,
    keywordstyle=\color{blue!70!black},
    commentstyle=\color{green!40!black},
    stringstyle=\color{red!60!black},
    showstringspaces=false,
    breaklines=true,
    frame=single,
    rulecolor=\color{black!30},
    columns=fullflexible,
    keepspaces=true,
    tabsize=2,
    xleftmargin=0pt,
    xrightmargin=0pt,
    framexleftmargin=1pt,
    framexrightmargin=1pt,
    aboveskip=2pt,
    belowskip=2pt
}

\date{}
\title{Region Aware Squeeze-and-Excitation Networks for Image Segmentation}

\author{%
\normalsize Jack Roberts\textsuperscript{1}\quad
Praneeth Tummala\textsuperscript{1}\quad
Kavi Sarna\textsuperscript{2}\quad
Jeova Farias\textsuperscript{2}\\[0.5em]
\small \textsuperscript{1}Department of Computer Science, Brown University, Providence, RI, USA\\
\small \textsuperscript{2}Department of Computer Science, Bowdoin College, Brunswick, ME, USA\\[0.3em]
\small\texttt{\{jack\_roberts2,praneeth\_tummala\}@brown.edu}\\
\small\texttt{kavs@capgroup.com, j.farias@bowdoin.edu}
}

\newcommand{\softmax}{\operatorname{Softmax}}

\newcommand{\convOneD}{\operatorname{Conv1D}}

\newcommand{\cbam}{{CBAM}\xspace}
\newcommand{\cse}{{cSE}\xspace}
\newcommand{\scse}{{scSE}\xspace}
\newcommand{\sse}{{sSE}\xspace}
\newcommand{\ema}{{EMA}\xspace}
\newcommand{\ca}{{CA}\xspace}

\newcommand{\se}{{SE}\xspace}
\newcommand{\eca}{{ECA}\xspace}
\newcommand{\base}{\texttt{BASE}\xspace}
\newcommand{\spase}{\texttt{SPASE}\xspace}
\newcommand{\please}{\texttt{PLEASE}\xspace}
\newcommand{\mapse}{\texttt{MAPSE}\xspace}

\begin{document}

\maketitle

\begin{abstract}
Squeeze-and-Excitation (\se) modules improve convolutional networks by modeling channel-wise dependencies, but collapsing all spatial structure into a single per-channel descriptor via global average pooling limits their effectiveness for dense prediction tasks such as image segmentation. We generalize the \se framework by replacing global pooling with spatially guided pooling, enabling region-aware channel recalibration across a family of modules with complementary cost profiles: some add essentially no parameters, others no measurable latency. Within this framework, we formalize four modules of increasing region flexibility: \base, a formalization of Tiled Squeeze-and-Excite~\cite{vosco2021tiled}, partitions the feature map into a regular grid; \spase aggregates over SLIC superpixels~\cite{achanta2012slic} for image-adaptive regions; and, our two most novel modules, \mapse sources its regions from a semantic map produced by a frozen, general-purpose network, while \please learns soft region assignments end-to-end via a lightweight assignment network. In every variant, each channel receives one gate per region, so the attention at each pixel reflects the region it belongs to, bridging channel and spatial attention; \please further yields its soft assignments as a free byproduct of its forward pass, giving structured spatial interpretability without post-hoc analysis. Across binary segmentation (CUB~\cite{wahCUB2011}, DUTS~\cite{wang2017}, ECSSD~\cite{shi2015hierarchical}), semantic segmentation (PASCAL VOC 2012~\cite{everingham2010pascal}, ADE20K~\cite{zhou2017scene}), and classification (mini-ImageNet~\cite{vinyals2016matching}), our modules outperform \se and related spatial-channel attention variants on most benchmarks, holding the majority of best results.

\end{abstract}


\section{Introduction}
Image segmentation is a fundamental problem in computer vision and medical imaging where every pixel must be assigned a semantic label. Fully convolutional networks~\cite{long2015fcn} and encoder--decoder architectures such as U-Net~\cite{ronneberger2015unet} are the de facto standard for this task, learning hierarchical representations via skip connections. Nevertheless, these architectures often treat each feature channel uniformly and rely on large receptive fields for global context.

Squeeze-and-Excitation (\se) modules~\cite{hu2018squeeze} improve convolutional networks by modelling inter-channel dependencies: global average pooling (\emph{squeeze}) yields a channel descriptor that is used to reweight each channel (\emph{excitation}). This significantly improves classification performance, but collapses spatial structure into a single scalar per channel, limiting effectiveness for dense prediction tasks such as segmentation~\cite{roy2018concurrent}. Prior work introduced spatial extensions such as spatial \se (\sse) and concurrent spatial--channel \se (\scse), incorporating spatial attention alongside channel-wise recalibration~\cite{roy2018concurrent,roy2019recalibrating}. These approaches still assign a single weight per channel or per spatial location, and therefore cannot adaptively model variation across semantically meaningful regions.

We address this gap by formalizing a family of \emph{region-aware} Squeeze-and-Excitation modules that incorporate spatially coherent region information directly into channel recalibration. We explore four variants of increasing region flexibility, the first three each a strict generalization of the previous: \base, a formalization of TSE~\cite{vosco2021tiled}, uses a regular $k \times k$ grid; \spase aggregates over SLIC superpixels~\cite{achanta2012slic} to produce image-adaptive regions; \mapse sources its regions from a semantic map inherited from a frozen, general-purpose network never trained specifically for the segmentation task, at no additional training cost; and \please learns soft region assignments end-to-end via a lightweight assignment network embedded in its pipeline.

The key idea shared across all four variants is to compute region-specific channel descriptors and project the resulting gating signals back to pixel space. Unlike prior \se variants, each channel receives multiple gating values (one per region) and the attention applied at each pixel reflects the region it belongs to. This formulation bridges channel and spatial attention, enabling adaptive feature recalibration aligned with the spatial structure of the scene. Beyond recalibration, \please produces soft input-dependent region assignments as a byproduct of its forward pass, providing spatial interpretability without any post-hoc analysis.   

\section{Related Work}

\subsection{Channel, Spatial, and Region Attention}
Channel attention mechanisms have become a standard tool for improving convolutional representations. The \se block~\cite{hu2018squeeze} introduced global channel recalibration via GAP and a learned MLP gate. Later variants enrich either the squeeze or excitation step: SRM~\cite{lee2019srm} augments GAP with global standard deviation to capture style-based statistics; FcaNet~\cite{qin2021fcanet} reframes GAP as a special case of DCT-based frequency decomposition and proposes multi-spectral descriptors; MCA~\cite{jiang2024mca} aggregates higher-order statistical moments via a Cross Moment Convolution module; \eca~\cite{wang2020eca} replaces the fully connected bottleneck with local cross-channel 1D convolutions; GCT~\cite{yang2020gated} adopts a non-parametric Gaussian excitation using L2-normalized channel contexts; and EAN~\cite{li2024ean} unifies normalization and attention via GroupNorm and L2 Norm.

Other approaches combine channel and spatial attention. \cbam~\cite{woo2018cbam} applies channel recalibration followed by a sequential spatial attention map. Gather-Excite~\cite{hu2018gather} aggregates feature responses from large spatial neighbourhoods via a gather operator and redistributes them to local features, showing that richer spatial context in the squeeze step yields consistent gains. Coordinate Attention (\ca)~\cite{hou2021coordinateattentionefficientmobile} factorizes 2D global pooling into two 1D encodings along the horizontal and vertical axes, producing direction- and position-sensitive maps effective for dense prediction. \ema~\cite{Ouyang_2023} avoids channel dimensionality reduction by grouping channels into parallel sub-features and fusing global and local spatial context via a cross-dimension interaction. SCSA~\cite{si2024scsaexploringsynergisticeffects} similarly combines spatial and channel attention by decomposing the feature map along its horizontal and vertical axes, applying multi-scale 1D convolutions to construct a dense spatial attention map, and subsequently using self-attention to compute a single image-level channel gate.

For segmentation specifically, Guha Roy \emph{et al.}~\cite{roy2018concurrent,roy2019recalibrating} proposed \cse, \sse, and \scse. SANet~\cite{SANet} similarly introduces spatial awareness through an average-pooled but not fully squeezed attention-convolution channel, whose output is upsampled to provide dense pixel-group-aware recalibration. PRM~\cite{ma2019prm} incorporates positional information and feature propagation into channel attention, while Project-and-Excite~\cite{rickmann2020project} extends \se to volumetric data by aggregating along spatial axes rather than collapsing it entirely.

Soft region cues have also been used to guide pixel-level attention. SNA~\cite{SNA} learns soft superpixel assignments but uses them to modulate local grid-based neighbourhood attention: each pixel remains a weighted combination of neighbouring pixel features rather than an explicitly pooled region representation. RANet~\cite{RANet} instead learns object-region construction and models information exchange between regions to enhance pixel representations, but as a dedicated semantic-segmentation architecture that uses regions for relational context exchange, rather than a drop-in recalibration module that pools and excites an independent channel descriptor for each region. Despite these novelties, none of the above methods pools features into a set of explicit region descriptors and independently excites each descriptor to produce a separate channel-gating vector per region.

A notable limitation shared across \se and its variants is the absence of built-in spatial interpretability: their channel-wise gates carry no spatial semantics, and interpretability must be applied post-hoc via external tools such as Grad-CAM~\cite{selvaraju2017grad}. \cbam~\cite{woo2018cbam} is most commonly associated with interpretability, owing to its explicit spatial attention map; however, that map is channel-agnostic --- a single scalar per spatial location shared across all channels --- and is not tied to any semantic decomposition of the image.

\subsection{Region Pooling and Channel Recalibration}

Tiled Squeeze-and-Excite (TSE)~\cite{vosco2021tiled} replaces global average pooling with average pooling over a regular spatial grid: each pooled tile descriptor passes independently through a shared excitation function, producing a channel vector broadcast over its tile. TSE therefore preserves multiple spatially indexed channel descriptors through the full squeeze-and-excitation operation, rather than collapsing the feature map into one global descriptor. We formalize this fixed-region pipeline as \base.

Region-wise feature aggregation has also been explored through superpixel pooling: Schuurmans \emph{et al.}~\cite{schuurmans2018efficientsemanticimagesegmentation} pool feature vectors over externally computed superpixel regions for semantic-segmentation prediction branches, establishing superpixel pooling as an existing image-adaptive aggregation method. \spase applies this idea to produce $K$ distinct gating values per channel that vary meaningfully across the feature map, integrating it into our generalised region-aware SE framework.

Our framework unifies these operations under a common formulation in which $K$ distinct regions are aggregated into $K$ region-specific channel descriptors, each excited independently, with the resulting $K$ channel-gating vectors projected back to pixel space to recalibrate the feature map. \base formalizes the fixed-grid case, \spase uses SLIC computed superpixel regions, \mapse sources its regions from a semantic map inherited from a frozen, general-purpose network, and \please learns soft region assignments end-to-end.

Beyond recalibration, \please produces its soft region assignments $P \in \mathbb{R}^{K \times h \times w}$ as a byproduct of its forward pass. Unlike CBAM's channel-agnostic spatial map, these $K$ maps form a learned, spatially coherent decomposition of the feature map into semantically meaningful regions, providing structured interpretability without post-hoc analysis.

\section{Background}

\subsection{Notation and Definitions}
A tensor $X \in \mathbb{R}^{c \times h \times w}$ has $c$ channels and spatial dimensions $h, w$. Scalars are unbolded (e.g. $x$), vectors bolded (e.g. $\mathbf{x}$), and tensors with spatial dimensions capitalized (e.g. $X$). Square brackets denote indexing (e.g. $X[c, h, w]$) and ``$:$'' denotes Python-style slicing, e.g. $X[:, h, w]$ for all channels at spatial location $(h,w)$. The notation $[k]$ denotes the index set $\{0, 1, \dots, k-1\}$ on a 0-based convention.

Given an image $I \in \mathbb{R}^{3 \times H \times W}$, a superpixelation algorithm partitions $[H] \times [W]$ into non-overlapping regions called \textit{superpixels}, encoded as a label map $S \in [K]^{H \times W}$, where $S[i,j] = k$ indicates pixel $(i,j)$ belongs to superpixel $s_k = \{(i,j) : S[i,j] = k\}$. We use SLIC~\cite{achanta2012slic} to find $S$, which clusters pixels in joint color-spatial space and exposes two knobs: the number of superpixels $K$ and a compactness $m$ trading off color similarity and spatial proximity, with larger $m$ yielding more regular, grid-like regions.

$\convOneD^{k}$ and $\operatorname{Conv2D}^{k \times k}$ denote 1D and 2D convolutions with kernel size $k$ and $\softmax$ denotes the usual normalized exponential over a vector's components.

\subsection{Squeeze-and-Excite Networks}
The Squeeze-and-Excite (SE) block~\cite{hu2018squeeze} recalibrates features by learning inter-channel dependencies. For input $X \in \mathbb{R}^{c \times h \times w}$, the block proceeds in two phases.

The \textit{squeeze} phase compresses each channel into a scalar via global average pooling (GAP), producing a descriptor $\mathbf{z}$:
\begin{equation}
    \mathbf{z} = \operatorname{GAP}(X), \quad \mathbf{z}[r] = \frac{1}{hw} \sum_{i \in [h]} \sum_{j \in [w]} X[r, i, j].
\end{equation}

Then the \textit{excite} phase follows. The descriptor $\mathbf{z}$ is transformed by a learnable module $\phi(\cdot)$, producing a channel-wise gate $\mathbf{w} \in \mathbb{R}^c$ of the same dimension: $\mathbf{w} = \phi(\mathbf{z})$.

In this work, we consider two variants for $\phi(\cdot)$:
\begin{itemize}
    \item \textit{MLP}: the original formulation~\cite{hu2018squeeze} processes $\mathbf{z}$ with a two-layer MLP and sigmoid gate $\sigma$:
    \begin{equation}\label{eq:se_mlp}
        \phi(\mathbf{x}) = \sigma(\Theta_{2}\, \operatorname{ReLU}(\Theta_{1}\mathbf{x})),
    \end{equation}
    where $\Theta_1, \Theta_2$ are learned weight matrices.
    \item \textit{Conv1D}: ECA~\cite{wang2020eca} captures local cross-channel interactions via a lightweight 1D convolution:
    \begin{equation}\label{eq:se_conv}
        \phi(\mathbf{x}) = \sigma(\convOneD^k(\mathbf{x})),
    \end{equation}
    where padding preserves the input size and $k = \left| (\log_2(c) + 1)/2 \right|_{\text{odd}}$ rounds to the nearest odd integer.
\end{itemize}

\begin{figure*}[t]
    \centering
    \begin{subfigure}[t]{0.4\linewidth}
        \centering
        \includegraphics[trim={24pt 9pt 26pt 10pt}, width=\linewidth]{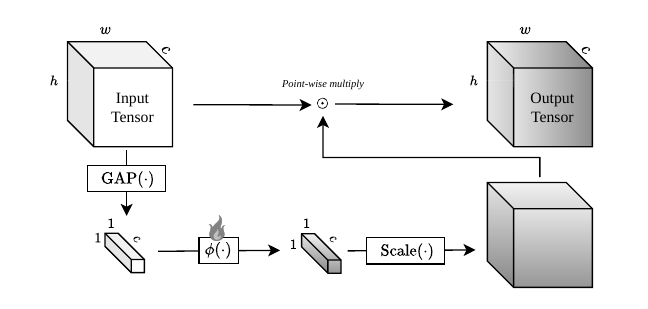}
        \caption{\se (Original)}\label{fig:SE}
    \end{subfigure}%
    \hspace{20pt}
    \begin{subfigure}[t]{0.4\linewidth}
        \centering
        \includegraphics[trim={26pt 9pt 26pt 10pt}, width=\linewidth]{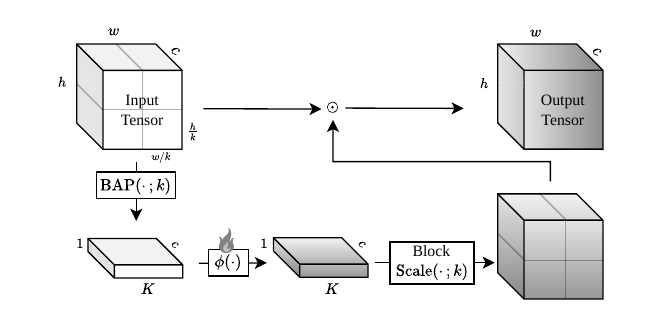}
        \caption{\base (TSE~\cite{vosco2021tiled} Formalization)}\label{fig:BASE}
    \end{subfigure}

    \vspace{10pt}

    \begin{subfigure}[t]{0.45\linewidth}
        \centering
        \includegraphics[trim={24pt 14pt 26pt 12pt},clip, width=\linewidth]{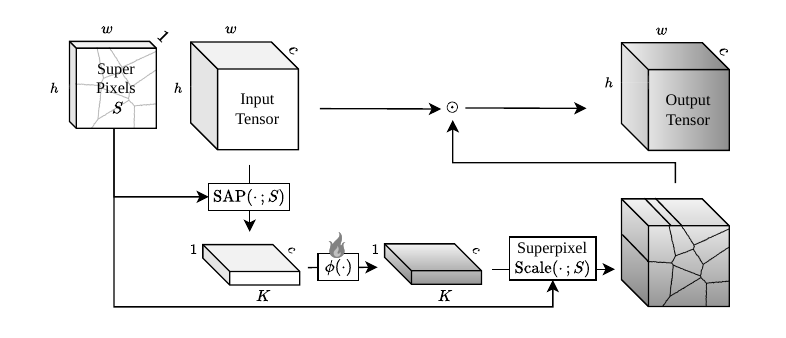}
        \caption{\spase (Proposed)}\label{fig:SPASE}
    \end{subfigure}%
    \hfill
    \begin{subfigure}[t]{0.524\linewidth}
        \centering
        \includegraphics[trim={24pt 13pt 24pt 10pt},clip, width=\linewidth]{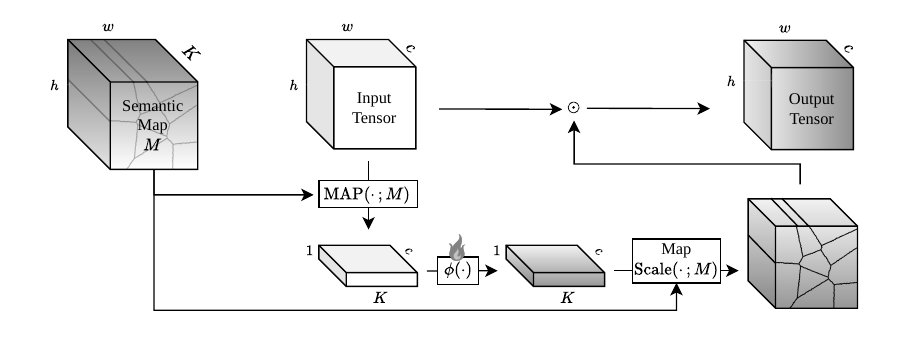}
        \caption{\mapse (Proposed)}\label{fig:MAPSE}
    \end{subfigure}

    \vspace{10pt}
    
    \begin{subfigure}[t]{0.524\linewidth}
        \centering
        \includegraphics[trim={24pt 13pt 24pt 10pt},clip, width=\linewidth]{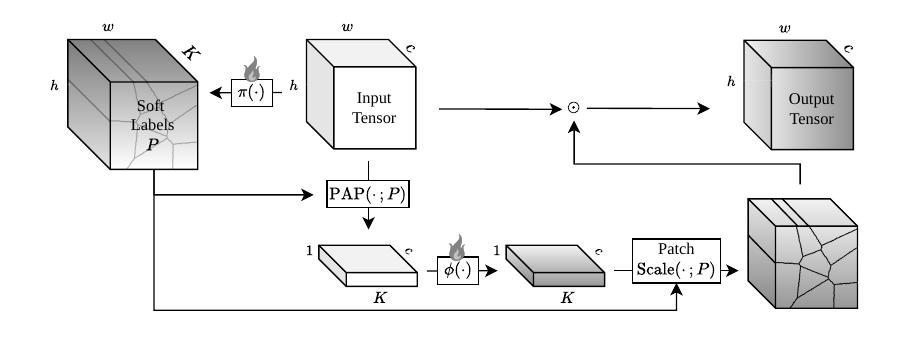}
        \caption{\please (Proposed)}\label{fig:PLEASE}
    \end{subfigure}

    \caption{Relevant Squeeze-and-Excite Modules. Learnable operations are marked with\includegraphics[height=10pt]{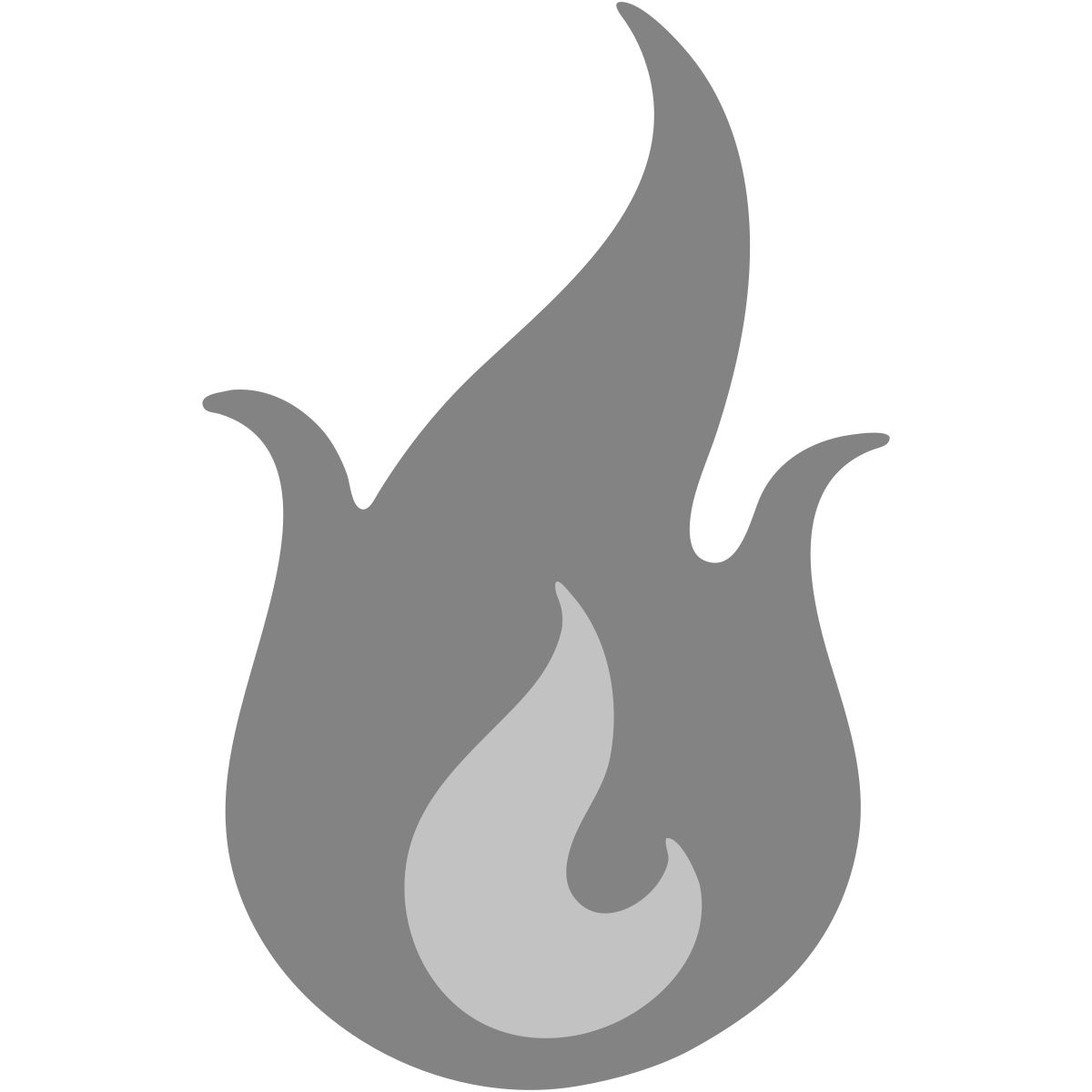}.} \label{fig:modules}
\end{figure*}

The gate $\mathbf{w}$ is broadcast (``scaled''~\cite{hu2018squeeze}) across spatial dimensions and applied to $X$ via pointwise multiplication, with broadcast tensor $W \in \mathbb{R}^{c \times h \times w}$ and block output $\widetilde{X}_{\text{SE}}$:
\begin{equation}\label{eq:excite_op}
    W[:, i, j] = \mathbf{w}, \; \forall\, i \in [h],\, j \in [w], \quad \widetilde{X}_{\text{SE}} = X \odot W.
\end{equation}
The input shape is preserved, allowing \se modules to be inserted anywhere in an existing architecture (Fig.~\ref{fig:SE}).

\section{Region-Aware Squeeze-and-Excite Blocks}
We generalize the \se framework along a single axis: the source of its pooling regions. First, we formalize TSE as a patchified version of \se that captures local features within a regular grid. Then we generalize this rigid spatial tensor partitioning to an input dependent regional splitting via superpixelation, fixed and computed offline at the network input level. Next, we replace this classical, task-agnostic partitioning with a semantic map inherited from a frozen, general-purpose deep network, allowing regions to reflect learned high-level structure at no additional training cost. We finally generalize this approach by learning soft regions at each \se module, enabling optimization-driven spatial partitioning. Each variant is described below and is easily implementable and with simple PyTorch code provided in the supplementary material. 




\subsection{Block Aggregation for SE}

Standard \se modules discard all spatial structure via GAP. TSE~\cite{vosco2021tiled} preserves coarse spatial structure by operating on a rigid $k \times k$ spatial grid, where aggregation occurs within each cell independently. We formalize this as the first, simplest member of our \textit{region-aware} SE family, which we call \textbf{Block Aggregation for Squeeze-and-Excite (BASE)}, depicted in Fig.~\ref{fig:BASE}.

The input tensor is divided into a $k \times k$ grid of non-overlapping blocks, each of shape $k' \times k'$, where $k' = h/k = w/k$ (assuming $k$ evenly divides $h$ and $w$). Average pooling is performed within each block, and the resulting per-block averages are concatenated into a single vector per channel. Formally, for $X \in \mathbb{R}^{c \times h \times w}$, Block Average Pooling with parameter $k$ ($\operatorname{BAP}$) yields $Z_{\text{block}} \in \mathbb{R}^{c \times k^2}$:
\[
Z_{\text{block}} = \operatorname{BAP}(X;\, k), 
\]
\begin{equation*}
    Z_{\text{block}}[r,\, uk + v] = \frac{1}{k'^2} \sum_{i \in \Omega_u} \sum_{j \in \Omega_v} X[r, i, j],
\end{equation*}
where $u, v \in \{0, \ldots, k-1\}$, $\Omega_u = \{uk', \ldots, (u+1)k' - 1\}$, and $\Omega_v = \{vk', \ldots, (v+1)k' - 1\}$. Note that $\operatorname{BAP}(\cdot;\, 1) = \operatorname{GAP}(\cdot)$, so \base strictly generalizes \se.

The excitation step applies $\phi(\cdot)$ independently to each column of $Z_{\text{block}}$, yielding $W_{\text{block}} \in \mathbb{R}^{c \times k^2}$. Because the same $\phi(\cdot)$ is shared across all $k^2$ block descriptors, \base adds no more parameters than a single \se excitation whatever the value of $k$ --- a property every variant below inherits.
Each scalar in $W_{\text{block}}$ is then broadcast uniformly over its corresponding spatial block to produce $W \in \mathbb{R}^{c \times h \times w}$, which is applied to $X$ via pointwise multiplication:
$$ W[:, i, j] = W_{\text{block}}[:,\, uk + v], \quad \forall\, i \in \Omega_u,\, j \in \Omega_v.$$

\begin{equation}
\widetilde{X}_{\text{BASE}} = X \odot W,
\end{equation}
where $u = \lfloor i/k' \rfloor$, $v = \lfloor j/k' \rfloor$. 

\subsection{Superpixel Aggregation for SE}

\base still aggregates input information over fixed, image-independent regions. We propose \textbf{Superpixel Aggregation for Squeeze-and-Excite (SPASE}), depicted in Fig.~\ref{fig:SPASE}, which replaces the rigid grid with image-adaptive and boundary preserving regions obtained via superpixels.

Let $S \in [K]^{H \times W}$ be the superpixelation of the input image into $K$ superpixels, computed once per image and reused at every layer the module is applied. At each layer, $S$ is resized to match the spatial resolution of the feature map via nearest-neighbor interpolation, yielding $S \in [K]^{h \times w}$. Formally, for $X \in \mathbb{R}^{c \times h \times w}$, Superpixel Average Pooling ($\operatorname{SAP}$) yields $Z_{\text{sp}} \in \mathbb{R}^{c \times K}$:

\begin{equation*}
    \begin{gathered} Z_{\text{sp}} = \operatorname{SAP}(X;\, S), \\ Z_{\text{sp}}[:, k] = \frac{1}{|s_k|} \sum_{(i,j) \in s_k} X[:, i, j],\end{gathered}
\end{equation*}
where $s_k = \{(i,j) : S[i,j] = k\}$ is the set of spatial locations assigned to superpixel $k$. Unlike the soft assignments of \mapse and \please below, which always place mass on every region, nearest-neighbor resizing can leave a superpixel empty at coarse resolutions, so we skip any region with $|s_k| = 0$. Note that when all superpixels are rectangular (i.e., high compactness) and arranged in a $\sqrt{K} \times \sqrt{K}$ grid, $\operatorname{SAP}$ reduces to $\operatorname{BAP}$ with $k = \sqrt{K}$.

Excitation applies $\phi(\cdot)$ independently to each column of $Z_{\text{sp}}$, yielding a gate matrix $W_{\text{sp}} \in \mathbb{R}^{c \times K}$. Each gate scalar is then broadcast over its corresponding superpixel to produce $W \in \mathbb{R}^{c \times h \times w}$, which is then applied to $X$:
$$W[:, i, j]\! =\!\! \sum_{k \in [K]}\! W_{\text{sp}}[:, k]\, \mathds{1}\{S[i,j]\! =\! k\}, \; \forall\, i \in [h],\, j \in [w].$$

\begin{equation}
\widetilde{X}_{\text{SPASE}} = X \odot W.
\end{equation}
Note that we can also write $W[:, i, j] = W_{\text{sp}}[:,\, S[i,j]], \forall i,j$. This means that each feature in the input tensor is reweighted by the gate corresponding to the superpixel it belongs to. Unlike \base, \spase uses image-dependent regions rather than drawing arbitrary divisions in the image. Because superpixel boundaries tend to align with object and color boundaries, each region descriptor is less likely to mix features from semantically distinct structures --- unlike a fixed grid cell, which can straddle an object boundary regardless of image content.

\subsection{Map Aggregation for SE}

\spase computes its regions with a classical, task-agnostic superpixel algorithm that operates purely on low-level color and spatial proximity, with no notion of the semantic content of the scene. We propose \textbf{Map Aggregation for Squeeze-and-Excite (MAPSE)}, which instead sources its regions from a \emph{semantic map}: a $K$-channel tensor that broadly localizes semantically meaningful structure within the image, produced once by a frozen, general-purpose deep network that was not trained for the segmentation task at hand. Because this semantic map already encodes learned, high-level structure rather than only low-level pixel similarity, \mapse's regions can be more meaningful than \spase's, while still avoiding the training cost of a fully learned partitioning.

As with $S$ in \spase, let $M \in \mathbb{R}^{K \times H \times W}_{\geq 0}$ be such a semantic map, normalized (via $\ell_1$ normalization) so that $\sum_{k \in [K]} M[k, i, j] = 1$ at every spatial location $(i,j)$, computed once per input image $I$ and bilinearly resized to $M \in \mathbb{R}^{K \times h \times w}$ at each layer the module is applied.

Map Average Pooling ($\operatorname{MAP}$) analogously yields $Z_{\text{map}} \in \mathbb{R}^{c \times K}$ for $X \in \mathbb{R}^{c \times h \times w}$:

\[
Z_{\text{map}} = \operatorname{MAP}(X;\, M),
\]
$$Z_{\text{map}}[:, k] = \frac{1}{m_k} \sum_{i \in [h]} \sum_{j \in [w]} M[k, i, j]\, X[:, i, j],$$
where $m_k = \sum_{i \in [h]} \sum_{j \in [w]} M[k, i, j]$ is the total assignment mass of region $k$; $m_k > 0$ is guaranteed whenever region $k$ receives nonzero evidence anywhere in the image.

As before, $\phi(\cdot)$ excites each column of $Z_{\text{map}}$ independently into a gate matrix $W_{\text{map}} \in \mathbb{R}^{c \times K}$, projected back to pixel space via $M$:
$$
\begin{gathered}W[:, i, j] = \sum_{k \in [K]} W_{\text{map}}[:, k]\, M[k, i, j], \\ \forall\, i \in [h],\, j \in [w].\end{gathered}
$$

\begin{equation}
\widetilde{X}_{\text{MAPSE}} = X \odot W.
\end{equation}
Unlike \spase, \mapse's regions are not tied to any particular geometry or hand-crafted algorithm: any tensor that broadly localizes semantic content in the image can serve as $M$. In practice we instantiate $M$ with two frozen, off-the-shelf sources: (1) the Grad-CAM activations~\cite{selvaraju2017grad} of the $K$ highest-scoring classes of an ImageNet-pretrained classifier, one channel per class, which for a classifier ending in global average pooling and a linear layer can be computed without a backward pass; and (2) the rectified projection of self-supervised DINOv3~\cite{simeoni2025dinov3} patch tokens onto their leading $K$ principal components, grouping patches by learned visual similarity rather than class evidence. Neither source is trained for, or even aware of, the downstream segmentation task, distinguishing \mapse from \please (described next), whose regions are optimized jointly with the rest of the network. The \mapse architecture is depicted in Fig.~\ref{fig:MAPSE}.

\begin{table*}[t]
\centering
\scriptsize
\setlength{\tabcolsep}{1.5pt}

\caption{Interpretability visualizations for representative good and bad segmentation predictions. The top two rows correspond to examples selected from the top 5\% of examples by mIoU, while the bottom two rows correspond to examples selected from the bottom 5\% by mIoU. CUB/DUTS/ECSSD use U-Net, while ADE20K uses Semantic U-Net.}

\resizebox{\textwidth}{!}{%
\begin{tabular}{lccccccccccc}
\toprule
Dataset & Input & GT & Prediction & E1 Argmax & E1 PCA & E2 Argmax & E2 PCA & E3 Argmax & E3 PCA & E4 Argmax & E4 PCA \\
\midrule

\textcolor{green!60!black}{ECSSD} &
\includegraphics[width=0.075\textwidth,height=0.50in,keepaspectratio]{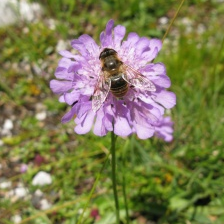} &
\includegraphics[width=0.075\textwidth,height=0.50in,keepaspectratio]{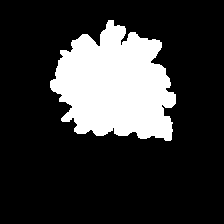} &
\includegraphics[width=0.075\textwidth,height=0.50in,keepaspectratio]{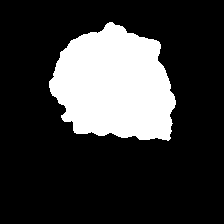} &
\includegraphics[width=0.075\textwidth,height=0.50in,keepaspectratio]{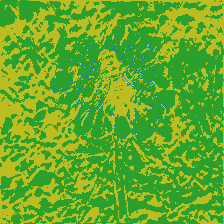} &
\includegraphics[width=0.075\textwidth,height=0.50in,keepaspectratio]{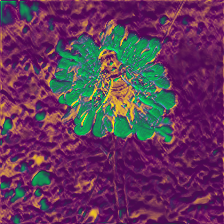} &
\includegraphics[width=0.075\textwidth,height=0.50in,keepaspectratio]{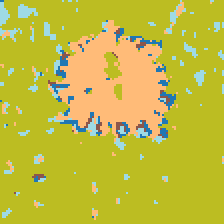} &
\includegraphics[width=0.075\textwidth,height=0.50in,keepaspectratio]{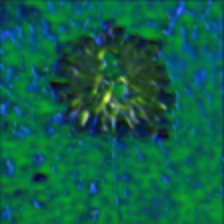} &
\includegraphics[width=0.075\textwidth,height=0.50in,keepaspectratio]{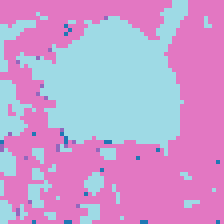} &
\includegraphics[width=0.075\textwidth,height=0.50in,keepaspectratio]{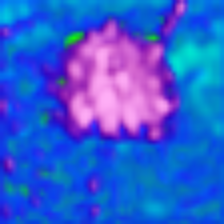} &
\includegraphics[width=0.075\textwidth,height=0.50in,keepaspectratio]{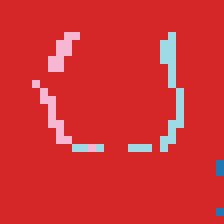} &
\includegraphics[width=0.075\textwidth,height=0.50in,keepaspectratio]{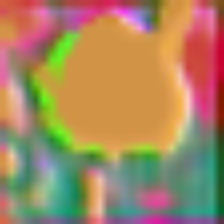} \\

\textcolor{green!60!black}{ADE20K} &
\includegraphics[width=0.075\textwidth,height=0.50in,keepaspectratio]{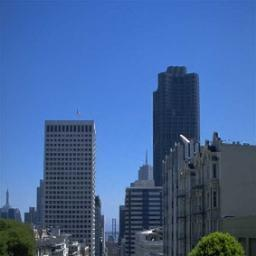} &
\includegraphics[width=0.075\textwidth,height=0.50in,keepaspectratio]{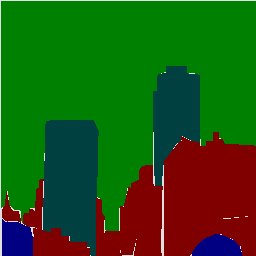} &
\includegraphics[width=0.075\textwidth,height=0.50in,keepaspectratio]{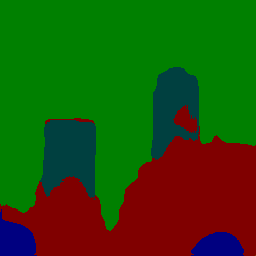} &
\includegraphics[width=0.075\textwidth,height=0.50in,keepaspectratio]{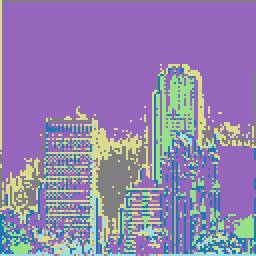} &
\includegraphics[width=0.075\textwidth,height=0.50in,keepaspectratio]{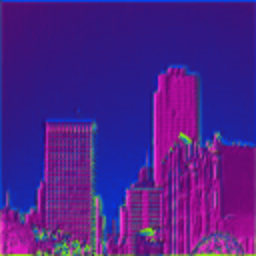} &
\includegraphics[width=0.075\textwidth,height=0.50in,keepaspectratio]{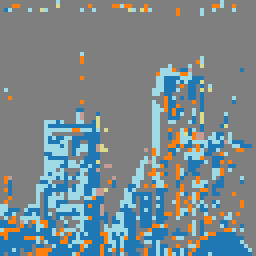} &
\includegraphics[width=0.075\textwidth,height=0.50in,keepaspectratio]{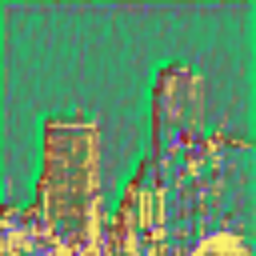} &
\includegraphics[width=0.075\textwidth,height=0.50in,keepaspectratio]{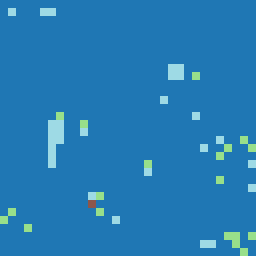} &
\includegraphics[width=0.075\textwidth,height=0.50in,keepaspectratio]{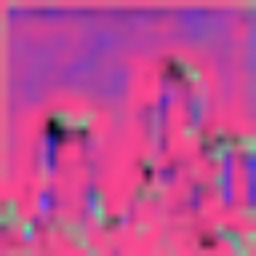} &
\includegraphics[width=0.075\textwidth,height=0.50in,keepaspectratio]{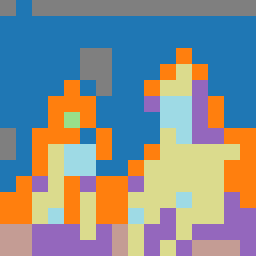} &
\includegraphics[width=0.075\textwidth,height=0.50in,keepaspectratio]{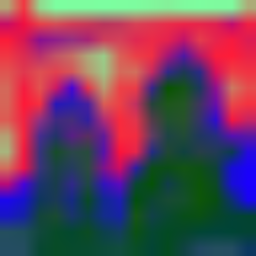} \\

\midrule

\textcolor{red!70!black}{CUB} &
\includegraphics[width=0.075\textwidth,height=0.50in,keepaspectratio]{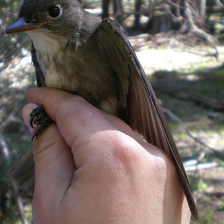} &
\includegraphics[width=0.075\textwidth,height=0.50in,keepaspectratio]{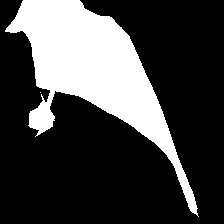} &
\includegraphics[width=0.075\textwidth,height=0.50in,keepaspectratio]{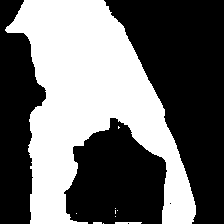} &
\includegraphics[width=0.075\textwidth,height=0.50in,keepaspectratio]{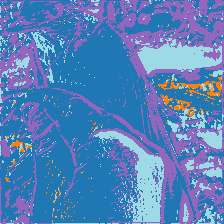} &
\includegraphics[width=0.075\textwidth,height=0.50in,keepaspectratio]{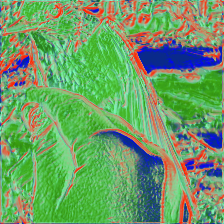} &
\includegraphics[width=0.075\textwidth,height=0.50in,keepaspectratio]{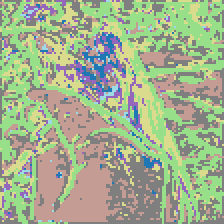} &
\includegraphics[width=0.075\textwidth,height=0.50in,keepaspectratio]{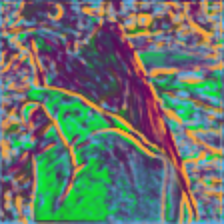} &
\includegraphics[width=0.075\textwidth,height=0.50in,keepaspectratio]{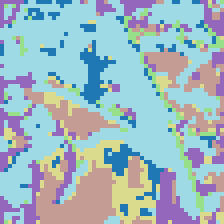} &
\includegraphics[width=0.075\textwidth,height=0.50in,keepaspectratio]{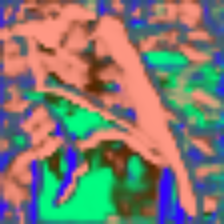} &
\includegraphics[width=0.075\textwidth,height=0.50in,keepaspectratio]{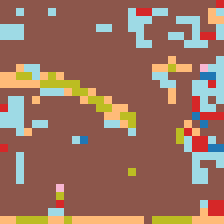} &
\includegraphics[width=0.075\textwidth,height=0.50in,keepaspectratio]{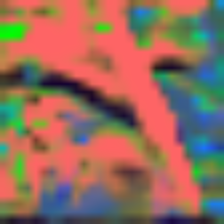} \\

\textcolor{red!70!black}{ADE20K} &
\includegraphics[width=0.075\textwidth,height=0.50in,keepaspectratio]{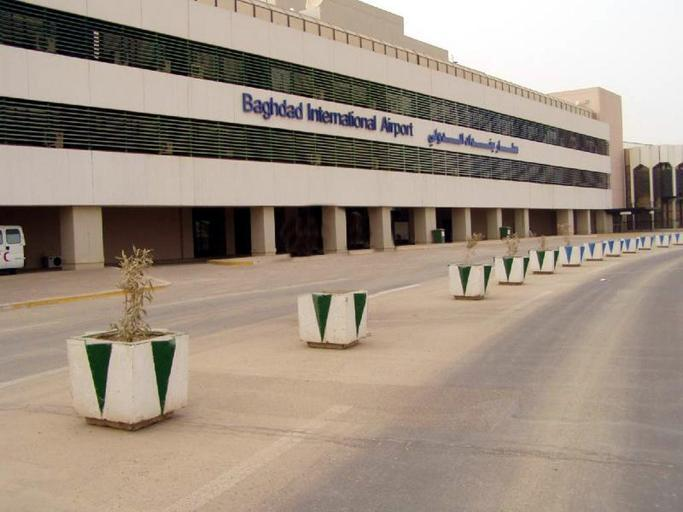} &
\includegraphics[width=0.075\textwidth,height=0.50in,keepaspectratio]{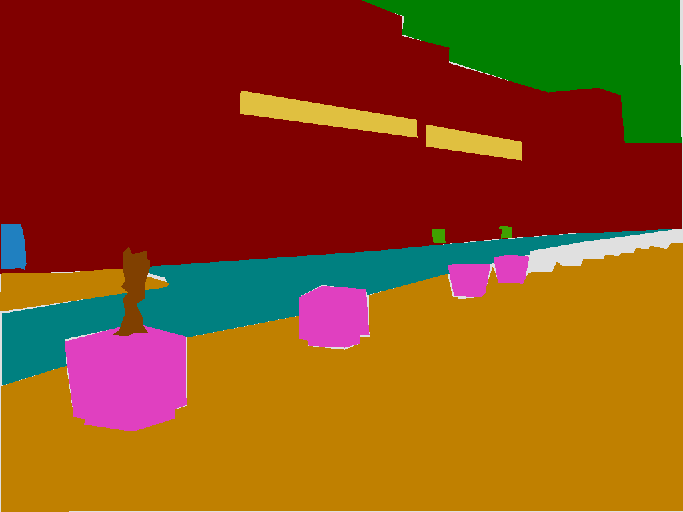} &
\includegraphics[width=0.075\textwidth,height=0.50in,keepaspectratio]{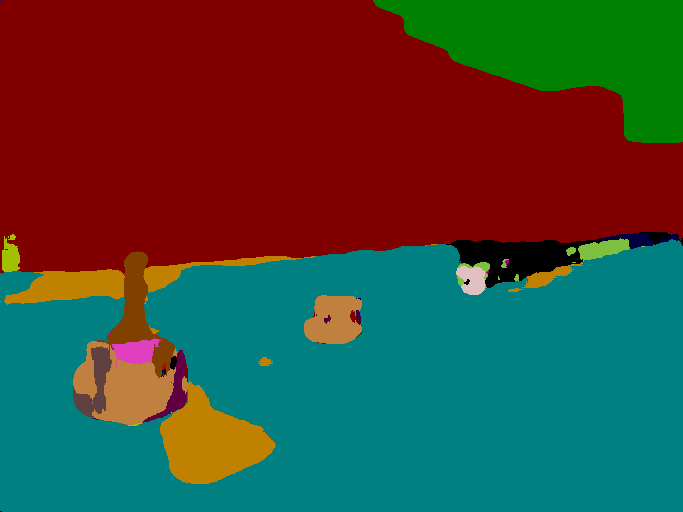} &
\includegraphics[width=0.075\textwidth,height=0.50in,keepaspectratio]{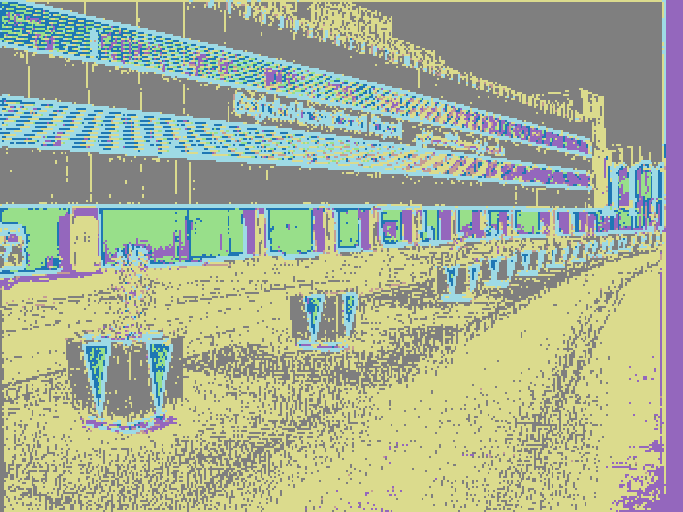} &
\includegraphics[width=0.075\textwidth,height=0.50in,keepaspectratio]{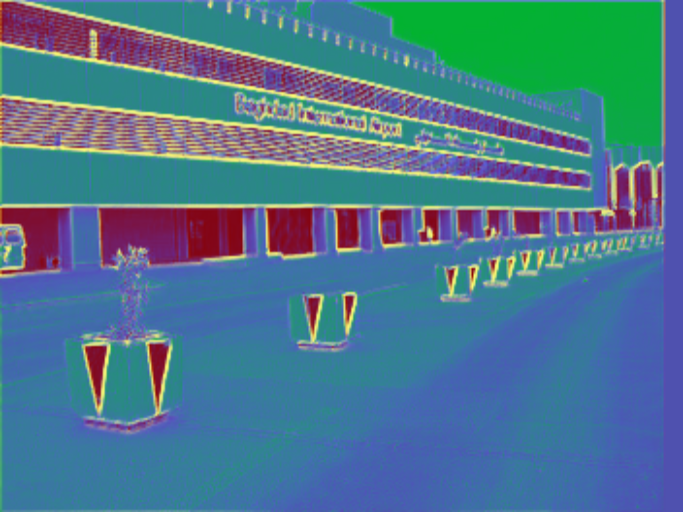} &
\includegraphics[width=0.075\textwidth,height=0.50in,keepaspectratio]{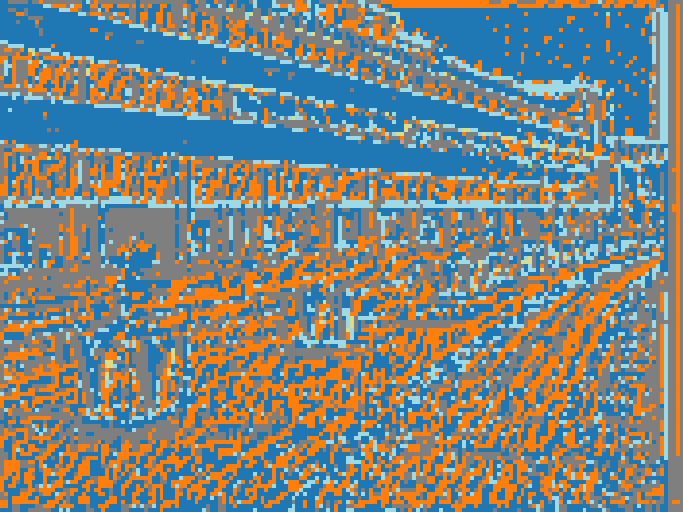} &
\includegraphics[width=0.075\textwidth,height=0.50in,keepaspectratio]{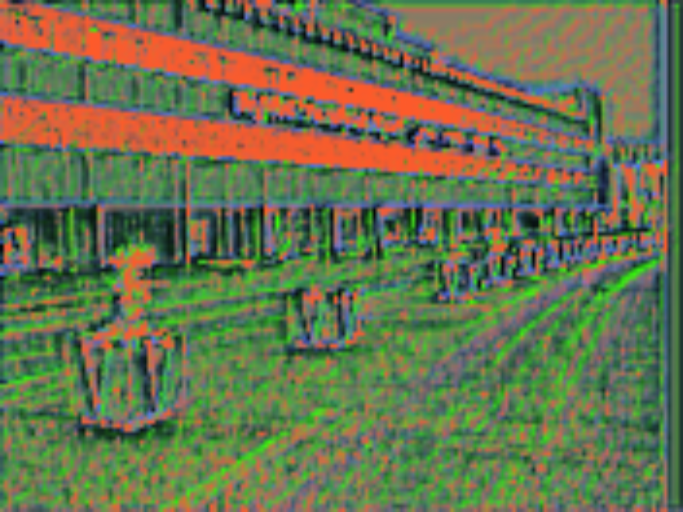} &
\includegraphics[width=0.075\textwidth,height=0.50in,keepaspectratio]{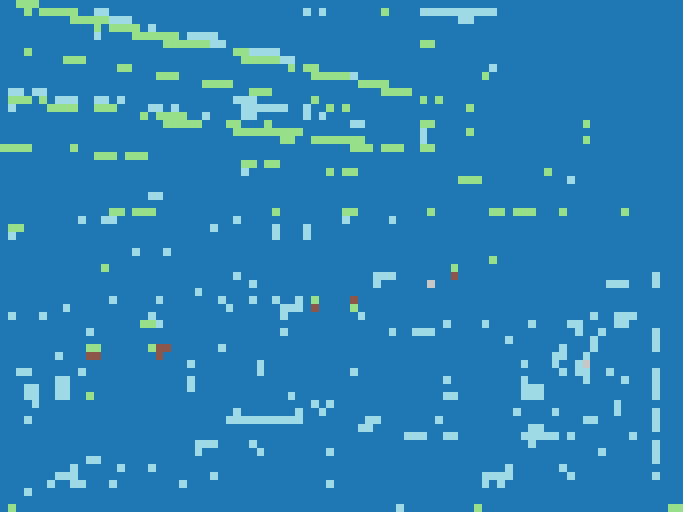} &
\includegraphics[width=0.075\textwidth,height=0.50in,keepaspectratio]{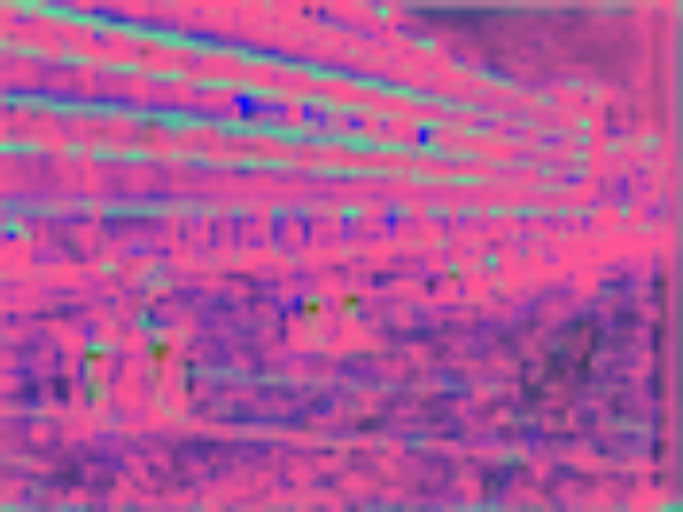} &
\includegraphics[width=0.075\textwidth,height=0.50in,keepaspectratio]{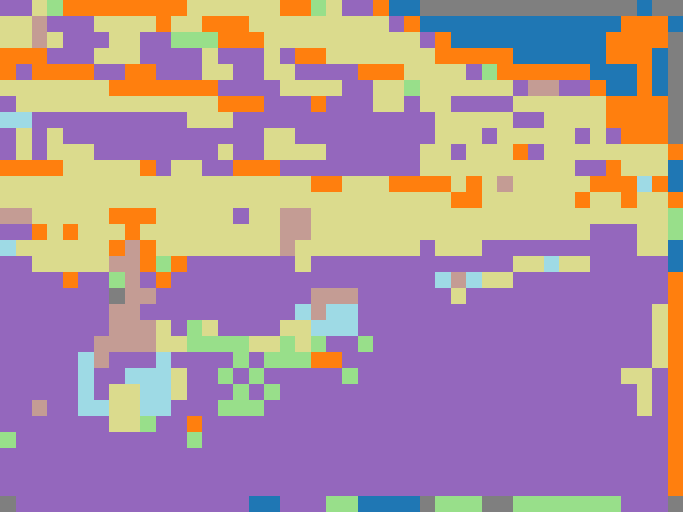} &
\includegraphics[width=0.075\textwidth,height=0.50in,keepaspectratio]{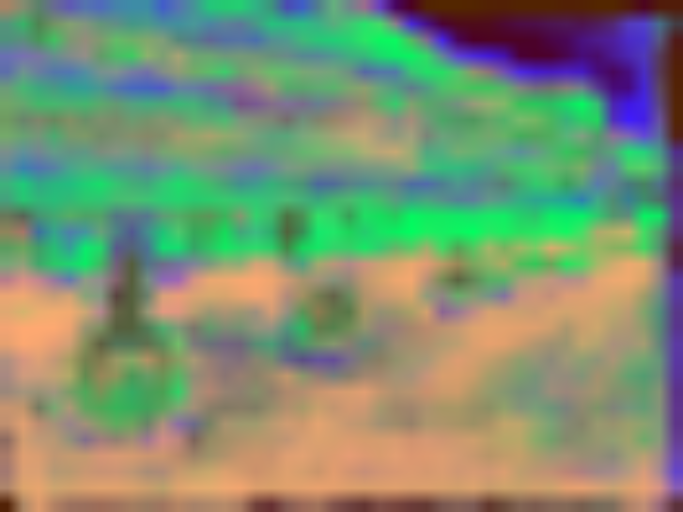} \\

\bottomrule
\end{tabular}%
}
\label{tab:please_condensed_interpretability_visuals}
\end{table*}

\subsection{Patch Learnable Aggregation for SE}

Both \spase and \mapse rely on a region partition that is fixed before training, whether computed by a classical superpixel algorithm (e.g., SLIC~\cite{achanta2012slic}) or inherited from a frozen external model. This introduces a non-learnable component, notable computational or memory overhead, and a partition that cannot adapt to the task or to the representation at the depth at which it is applied. To address these limitations, we propose \textbf{Patch Learnable Aggregation for Squeeze-and-Excite} (\textbf{PLEASE}), which replaces externally computed region assignments with learned soft assignments trained end-to-end with the network.

The module begins with a learnable soft-assignment layer $\pi(\cdot): \mathbb{R}^{c \times h \times w} \rightarrow \mathbb{R}^{K \times h \times w}$, followed by a softmax over regions. For $X \in \mathbb{R}^{c \times h \times w}$:
$$\pi(X)\! = \!\softmax\!\left(\operatorname{Conv2D}_{2}^{1\times 1}\!\left(\operatorname{ReLU}\!\left(\operatorname{Conv2D}_{1}^{1\times 1}(X)\!\right)\!\right)\!\right),$$
where the softmax is applied along the $K$ dimension, so each spatial location $(i,j)$ holds a probability distribution over $K$ learned regions. Define $P = \pi(X)$.

Patch Average Pooling ($\operatorname{PAP}$) then computes $Z_{\text{patch}} \in \mathbb{R}^{c \times K}$ exactly as $\operatorname{MAP}$, with $P$ in place of $M$:
\[
Z_{\text{patch}} = \operatorname{PAP}(X;\, P),
\]
\begin{equation}
    Z_{\text{patch}}[:, k] = \frac{1}{m_k} \sum_{i \in [h]} \sum_{j \in [w]} P[k, i, j]\, X[:, i, j],
\label{eq:PAP}
\end{equation}
where $m_k = \sum_{i \in [h]} \sum_{j \in [w]} P[k, i, j]$; since $P$ is a softmax output, $m_k > 0$ is guaranteed for all $k$, so the normalization is always well-defined.

As before, $\phi(\cdot)$ excites $Z_{\text{patch}}$ into a gate matrix $W_{\text{patch}} \in \mathbb{R}^{c \times K}$, projected back to pixel space via $P$:
 $$ \begin{gathered}W[:, i, j] = \sum_{k \in [K]} W_{\text{patch}}[:, k]\, P[k, i, j], \\ \forall\, i \in [h],\, j \in [w].\end{gathered}$$

\begin{equation}
    \widetilde{X}_{\text{PLEASE}} = X \odot W.
\end{equation}
Unlike \base, which partitions the feature map into fixed grid cells, \spase, which relies on an external segmentation algorithm, or \mapse, which relies on a frozen external semantic map, \please learns a lightweight soft partitioning end-to-end, allowing the regions themselves to adapt to the task. Note that \please reduces to \spase when the soft assignments $P$ are replaced by hard assignments, i.e., when $P[k, i, j] = \mathds{1}\{S[i,j] = k\}$ for some label map $S \in [K]^{h \times w}$. The architecture is depicted in Fig.~\ref{fig:PLEASE}.

Note \please's soft assignment module learns spatially coherent regions despite a $1\times 1$ conv filter, which has no receptive field of its own. That coherence therefore cannot originate in $\pi$'s own spatial support; it must come from the signal $\pi$ reads. This is possible because $\pi(\cdot)$ operates on $X$, which already carries spatial correlations from the receptive fields of the preceding convolutional layers; \please only needs to classify this locally smooth signal, not to introduce spatial coherence itself.

\section{Experiments}

\subsection{Experimental Setup}

We evaluate on binary segmentation benchmarks CUB~\cite{wahCUB2011}, DUTS~\cite{wang2017}, and ECSSD~\cite{shi2015hierarchical}; semantic segmentation benchmarks PASCAL VOC 2012~\cite{everingham2010pascal} and ADE20K~\cite{zhou2017scene}; and a classification benchmark, mini-ImageNet~\cite{vinyals2016matching}, derived from ImageNet~\cite{imagenet15russakovsky}. We compare against the dominant channel and spatial attention baselines --- \se~\cite{hu2018squeeze}, \scse~\cite{roy2018concurrent}, \eca~\cite{wang2020eca}, Coordinate Attention (CA)~\cite{hou2021coordinateattentionefficientmobile}, Efficient Multi-Scale Attention (EMA)~\cite{Ouyang_2023}, and \cbam~\cite{woo2018cbam} --- focusing on CNNs, where channel attention was developed and is most widely benchmarked. Further results appear in the supp. material.

Binary experiments use the U-Net~\cite{ronneberger2015unet} and FCN~\cite{long2015fcn} architectures detailed in the supp. material, trained on an 80/10/10 train/validation/test split with $224 \times 224$ inputs for 50 epochs via AdamW (learning rate $10^{-3}$, weight decay $10^{-4}$, batch size 16), using random crops and horizontal flips ($p=0.5$); evaluation uses center crops with ImageNet normalization.

Semantic experiments use an ImageNet-pretrained ResNet-34~\cite{DBLP:journals/corr/HeZRS15} Semantic U-Net with dataset-specific prediction heads: VOC models train on the SBD-extended VOC \textit{trainaug} split~\cite{hariharan2011semantic} and evaluate on the VOC2012 validation split; ADE20K models use the official train/validation splits. Training uses random scaling in $[0.5, 2.0]$, padding, random $512 \times 512$ crops, horizontal flips, and ImageNet normalization; validation uses full-resolution images.

Classification experiments use ResNet-50 on the 100-class mini-ImageNet dataset, with attention modules inserted into every residual module before the residual addition, trained for 100 epochs with SGD (learning rate $0.1$, momentum $0.9$, weight decay $10^{-4}$, batch size 128) and a multistep schedule decayed by $0.1$ at epochs 30, 60, and 90. Training uses random resized $224 \times 224$ crops and horizontal flips; evaluation resizes to 256 and center-crops to $224 \times 224$.

All configurations are run with three random seeds; we report mean $\pm$ standard deviation for mIoU~\cite{everingham2010pascal} and Boundary F1~\cite{martin2004learning} (2-pixel tolerance) for segmentation, and top-1/top-5 accuracy for classification.

Modules are inserted at the U-Net encoder activations before downsampling~\cite{KC-UNet,coronary-unet}, the FCN skip tensors before decoder fusion, the Semantic U-Net's pretrained-encoder feature maps, and the ResNet-50 residual branch before the identity addition.

Demo code for the U-Net experiments accompanies the supp. material, which also states the full configuration; the codebase will be released upon acceptance.

\begin{table}[!h]
\centering
\footnotesize
\caption{U-Net complexity and latency at input size $3{\times}224{\times}224$. Latency is reported in ms/image for FP32 inference at batch size 1, using 50 warm-up and 200 measured iterations. GPU latency was measured on a single NVIDIA GeForce RTX 3090 (24\,GB); CPU latency was measured single-threaded on one core of a dual-socket AMD EPYC 7532 node ($2{\times}32$ physical cores; SMT disabled).}

\label{tab:unet-complexity}
\setlength{\tabcolsep}{4pt}
\renewcommand{\arraystretch}{1.15}
\resizebox{\linewidth}{!}{%
\begin{tabular}{lrrrrrr}
\toprule
Model & Params & \shortstack{Added\\ params} & GFLOPs & \shortstack{Added \\ GFLOPs} & \shortstack{CPU \\ latency} & \shortstack{GPU \\ latency} \\
\midrule
Baseline & 7.763M & 0 & 18.551 & 0.000 & 237.69 & 2.10 \\
SE  & 7.774M & +11,390 & 18.557 & +0.006 & 245.33 & 2.83 \\
\scse  & 7.775M & +11,874 & 18.569 & +0.018 & 244.90 & 3.16 \\
ECA  & 7.763M & +16 & 18.557 & +0.006 & 241.73 & 2.58 \\
CA  & 7.782M & +18,704 & 18.565 & +0.015 & 243.62 & 3.42 \\
EMA & 7.764M & +910 & 18.738 & +0.187 & 270.51 & 4.04 \\
CBAM & 7.775M & +11,782 & 18.582 & +0.031 & 246.20 & 4.73 \\
\midrule
\mapse (GradCAM) & 7.763M & +16 & 28.445 & +9.894 & 436.72 & 9.72 \\
\mapse (DINO) & 7.763M & +16 & 28.209 & +9.658 & 402.80 & 24.13 \\
\base & 7.763M & +16 & 18.557 & +0.006 & 249.39 & 2.87 \\
\spase & 7.763M & +16 & 18.592 & +0.041 & 260.73 & 8.28 \\
\please & 7.786M & +22,768 & 18.766 & +0.216 & 247.41 & 3.68 \\
\bottomrule
\end{tabular}%
}
\end{table}

\begin{table*}[t]

\centering
\scriptsize
\caption{U-Net comparison across the binary segmentation datasets. All values in $\%$. Best results are in \textbf{bold}, 2nd best \underline{underlined}.}

\label{tab:main-results}
\setlength{\tabcolsep}{4pt}
\renewcommand{\arraystretch}{1.15}
\begin{adjustbox}{max width=\textwidth}
\begin{tabular}{lccccccc}
\toprule
& & \multicolumn{2}{c}{CUB} & \multicolumn{2}{c}{DUTS} & \multicolumn{2}{c}{ECSSD} \\
\cmidrule(lr){3-4} \cmidrule(lr){5-6} \cmidrule(lr){7-8}
Model & Added params & mIoU & Boundary F1 & mIoU & Boundary F1 & mIoU & Boundary F1 \\
\midrule
Baseline
& $0.000$
& 91.66 $\pm$ 0.15 & 76.56 $\pm$ 0.02
& 82.06 $\pm$ 0.57 & 59.54 $\pm$ 1.08
& 64.54 $\pm$ 2.28 & 28.43 $\pm$ 0.77
 \\

SE
& $+0.147$
& 91.71 $\pm$ 0.26 & 77.36 $\pm$ 0.48
& 82.41 $\pm$ 0.47 & 60.35 $\pm$ 0.81
& 68.08 $\pm$ 0.04 & 31.99 $\pm$ 0.30
 \\

\scse
& $+0.153$
& 91.94 $\pm$ 0.27 & 77.78 $\pm$ 0.61
& 82.94 $\pm$ 0.18 & 61.23 $\pm$ 0.42
& 67.49 $\pm$ 1.02 & 35.67 $\pm$ 1.11
 \\

ECA
& $+0.000$
& 92.23 $\pm$ 0.34 & 78.53 $\pm$ 0.94
& 84.07 $\pm$ 0.39 & 63.63 $\pm$ 0.17
& 65.86 $\pm$ 2.52 & 32.41 $\pm$ 3.14
 \\

CA
& $+0.241$
& 92.07 $\pm$ 0.16 & 78.21 $\pm$ 0.65
& 83.61 $\pm$ 0.65 & 63.67 $\pm$ 0.77
& 65.37 $\pm$ 0.71 & 33.06 $\pm$ 0.81
 \\

EMA
& $+0.012$
& 92.53 $\pm$ 0.20 & 79.41 $\pm$ 0.26
& 84.36 $\pm$ 0.15 & 64.77 $\pm$ 0.44
& \underline{68.82} $\pm$ 0.55 & \textbf{40.80} $\pm$ 2.96
 \\

CBAM
& $+0.152$
& 91.58 $\pm$ 0.08 & 77.31 $\pm$ 0.53
& 82.64 $\pm$ 0.21 & 60.68 $\pm$ 1.08
& 63.79 $\pm$ 1.00 & 26.50 $\pm$ 2.71
 \\ \midrule



\mapse (GradCAM)
& $+0.000$
& \textbf{92.97} $\pm$ 0.22 & \textbf{79.75} $\pm$ 0.70
& \underline{85.45} $\pm$ 0.49 & \underline{65.67} $\pm$ 0.36
& \textbf{70.23} $\pm$ 2.58 & 34.98 $\pm$ 4.88
\\

\mapse (DINO)
& $+0.000$
& 92.63 $\pm$ 0.34 & 79.08 $\pm$ 0.71
& \textbf{87.15} $\pm$ 0.44 & \textbf{68.82} $\pm$ 0.53
& 66.40 $\pm$ 2.08 & 32.29 $\pm$ 4.47
\\

\base
& $+0.000$
& 92.02 $\pm$ 0.15 & 77.72 $\pm$ 0.65
& 83.26 $\pm$ 0.54 & 61.88 $\pm$ 1.12
& 65.59 $\pm$ 1.58 & 30.93 $\pm$ 2.38
\\


\spase
& $+0.000$
& 92.38 $\pm$ 0.32 & 78.49 $\pm$ 0.86
& 83.59 $\pm$ 0.48 & 61.76 $\pm$ 0.56
& 68.29 $\pm$ 0.80 & 36.30 $\pm$ 1.59
\\


\please
& $+0.293$
& \underline{92.75} $\pm$ 0.28 & \underline{79.50} $\pm$ 0.03
& 84.71 $\pm$ 0.58 & 64.76 $\pm$ 0.69
& 66.53 $\pm$ 0.29 & \underline{36.40} $\pm$ 2.37
\\
\bottomrule
\end{tabular}
\end{adjustbox}

\end{table*}

\subsection{Ablations}

We ablate attention location and granularity, the choice of $\phi(\cdot)$, the region count $K$, and residual connections in the supplementary material. From these we adopt the following defaults: modules at U-Net encoder stages E1--E4; ECA-style Conv1D excitation (Eq.~\ref{eq:se_conv}); no residual connection, as in standard channel attention; and $K=8$ regions for every method except \mapse (GradCAM), which uses $K=128$ top-scoring classes, and \base, which uses a $2 \times 2$ grid. We use $K=8$ as a common default wherever the supplementary ablations separate settings by less than run-to-run variation, and depart from it only where the ablation is decisive. \mapse (GradCAM) is the one such case: its $K$ indexes a stack of top-scoring class activations rather than a partition whose regions must each own a share of the image, so it stays informative at far larger $K$.

\subsection{Model Interpretability} 

Table~\ref{tab:please_condensed_interpretability_visuals} illustrates the spatial interpretability that emerges during \please's forward pass. It visualizes the learned soft-assignment maps at each U-Net encoder layer (E1--E4) using Argmax, which assigns each pixel its highest-probability region, and PCA, which projects the full $K$-dimensional assignment vectors into RGB. Similar PCA colors indicate similar region-membership patterns. More visualizations are provided in the supplementary material.

Successful segmentations show a clear progression: early PLEASE layers produce broad regions driven by low-level appearance, which deeper layers reorganize toward the task objective. The ECSSD bee first splits into bee, flower, and background before later layers merge the first two into one foreground concept; the ADE20K building sharpens from a coarse building/skyline split into meaningful scene structure. The failures show little of this, their PCA maps staying fragmented and never converging. These are qualitative observations on a few examples rather than a measured effect and inform a hypothesis for future work, namely that strong structural priors (as in \spase) help most in early layers and flexibility (in \please) in deeper ones.

\begin{table*}[t!]
\centering

\begin{minipage}[t]{0.58\textwidth}
\centering
\scriptsize
\caption{Semantic U-Net comparison on PASCAL VOC 2012 and ADE20K. All values in $\%$. Best results are in \textbf{bold}, 2nd best \underline{underlined}.}

\label{tab:voc-semantic-unet-results}
\setlength{\tabcolsep}{3pt}
\renewcommand{\arraystretch}{1.15}
\begin{adjustbox}{max width=\linewidth}
\begin{tabular}{lccccc}
\toprule
& & \multicolumn{2}{c}{VOC2012} & \multicolumn{2}{c}{ADE20K} \\
\cmidrule(lr){3-4} \cmidrule(lr){5-6}
Model & Added params & mIoU & Boundary F1 & mIoU & Boundary F1 \\
\midrule
Baseline
& $0.000$
& 73.44 $\pm$ 0.48 & 30.13 $\pm$ 0.36
& 30.45 $\pm$ 0.42 & \underline{44.25} $\pm$ 0.32
\\

SE
& $+0.048$
& 73.92 $\pm$ 0.56 & \underline{30.21} $\pm$ 0.42
& 32.04 $\pm$ 0.17 & 43.94 $\pm$ 0.17
\\

\scse
& $+0.050$
& 73.05 $\pm$ 0.79 & 29.92 $\pm$ 0.04
& 31.81 $\pm$ 0.74 & 43.88 $\pm$ 0.27
\\

ECA
& $+0.000$
& 73.61 $\pm$ 0.21 & 29.97 $\pm$ 0.75
& 32.41 $\pm$ 0.38 & 43.35 $\pm$ 0.03
\\

CA
& $+0.080$
& 73.53 $\pm$ 0.42 & 29.61 $\pm$ 0.22
& 32.28 $\pm$ 0.29 & \textbf{44.31} $\pm$ 0.25
\\

EMA
& $+0.004$
& 73.80 $\pm$ 0.54 & 29.97 $\pm$ 0.29
& \underline{32.44} $\pm$ 0.13 & 42.92 $\pm$ 0.58
\\

CBAM
& $+0.050$
& 74.06 $\pm$ 0.74 & \textbf{30.47} $\pm$ 0.38
& 31.80 $\pm$ 0.31 & 43.44 $\pm$ 0.35
\\ \midrule


\mapse (GradCAM)
& $+0.000$
& 73.26 $\pm$ 0.28 & 30.21 $\pm$ 0.18
& 32.43 $\pm$ 0.29 & 43.98 $\pm$ 0.44
\\

\mapse (DINO)
& $+0.000$
& \underline{74.20} $\pm$ 0.61 & 29.71 $\pm$ 0.40
& \textbf{32.91} $\pm$ 0.23 & 43.19 $\pm$ 0.13
\\

\base
& $+0.000$
& 73.19 $\pm$ 0.49 & 29.74 $\pm$ 0.30
& 32.16 $\pm$ 0.15 & 43.99 $\pm$ 0.28
\\


\spase
& $+0.000$
& 73.30 $\pm$ 0.40 & 29.73 $\pm$ 0.19
& 31.63 $\pm$ 0.32 & 43.99 $\pm$ 0.35
\\


\please
& $+0.096$
& \textbf{75.05} $\pm$ 0.25 & 29.73 $\pm$ 0.37
& 32.19 $\pm$ 0.68 & 43.50 $\pm$ 0.23
\\
\bottomrule
\end{tabular}
\end{adjustbox}
\end{minipage}
\hfill
\begin{minipage}[t]{0.40\textwidth}
\centering
\scriptsize
\caption{ResNet-50 comparison on Mini-ImageNet. All values in $\%$. Best results are in \textbf{bold}, 2nd best \underline{underlined}.}

\label{tab:mini-imagenet-results}
\setlength{\tabcolsep}{3pt}
\renewcommand{\arraystretch}{1.15}
\begin{adjustbox}{max width=\linewidth}
\begin{tabular}{lccc}
\toprule
Model & Added params & Top-1 & Top-5 \\
\midrule
Baseline
& $0.000$
& 72.09 $\pm$ 2.00
& 89.95 $\pm$ 0.78
\\

SE
& $+10.673$
& 76.17 $\pm$ 0.58
& 92.03 $\pm$ 0.44
\\

\scse
& $+10.737$
& 73.03 $\pm$ 0.77
& 90.51 $\pm$ 0.31
\\

ECA
& $+0.000$
& 75.81 $\pm$ 0.61
& 91.76 $\pm$ 0.42
\\

CA
& $+16.044$
& 77.46 $\pm$ 0.27
& 92.49 $\pm$ 0.38
\\

EMA
& $+0.837$
& 75.47 $\pm$ 0.45
& 91.83 $\pm$ 0.08
\\

CBAM
& $+10.680$
& 75.05 $\pm$ 0.47
& 91.73 $\pm$ 0.12
\\
\midrule

\mapse (GradCAM)
& $+0.000$
& \textbf{80.53} $\pm$ 0.28
& \textbf{94.13} $\pm$ 0.44
\\

\mapse (DINO)
& $+0.000$
& 76.05 $\pm$ 0.74
& 91.89 $\pm$ 0.38
\\

\base
& $+0.000$
& 77.25 $\pm$ 0.47
& 92.25 $\pm$ 0.51
\\

\spase
& $+0.000$
& \underline{77.87} $\pm$ 0.73
& \underline{92.77} $\pm$ 0.37
\\

\please
& $+21.340\%$
& 77.11 $\pm$ 0.59
& 91.97 $\pm$ 0.24
\\
\bottomrule
\end{tabular}
\end{adjustbox}
\end{minipage}

\end{table*}

\subsection{Results}

\paragraph{Runtime Analysis}

Table~\ref{tab:unet-complexity} compares complexity and inference latency for binary U-Net segmentation. \spase adds few parameters but incurs SLIC latency, whereas \please replaces SLIC with a learnable soft region assignment block, increasing parameters but reducing latency. For \mapse (GradCAM) and \mapse (DINO), parameter counts include only the trainable student, while GFLOPs include the frozen ResNet-50 or DINOv3 backbone. End-to-end latency also includes CAM generation or DINO feature extraction and PCA assignment construction. The variants therefore occupy complementary points in the cost space rather than a single one: \base and \spase add only the 16-weight \eca kernel at no measurable FLOP cost, \please adds parameters ($+0.293\%$) but matches \base's GPU latency, and \mapse also adds only the 16-weight kernel yet buys its regions with the frozen teacher's forward pass ($+8.3$ GFLOPs). Whichever budget is scarcest --- parameters, FLOPs, or latency --- there is a region-aware variant that spends elsewhere. Runtime results for FCN, Semantic U-Net, and ResNet-50 are provided in the supp. material and show similar trends.


\paragraph{Binary Segmentation}

Table~\ref{tab:main-results} shows that region-aware recalibration consistently improves the U-Net baseline while preserving the lightweight character of \se-style attention: every proposed variant improves over baseline on all three datasets and both metrics, and the proposed modules take five of the six best entries. Because all variants share the same ECA-style channel excitation operator, their differences isolate the effect of region construction. Externally sourced regions are the strongest: \mapse (GradCAM) leads CUB on both metrics and gives the best ECSSD mIoU, while \mapse (DINO) leads DUTS by a clear margin (87.15 mIoU and 68.82 Boundary F1, against 82.06 and 59.54 for baseline). Among the remaining variants, \please is the runner-up on CUB and the stronger of the two on DUTS, while \spase gives the better ECSSD mIoU; and even the fixed-grid \base improves everywhere. The exception is ECSSD Boundary F1, where EMA is strongest and \please follows. The FCN comparison in the supp. material behaves similarly: the proposed modules hold the best entries, but ECSSD is more sensitive to region type.

The proposed modules are most decisive where the backbone already represents the data well: CUB and DUTS reach comparatively high absolute mIoU and are where the region-aware methods win outright, whereas ECSSD is substantially lower across all models and its gains are less consistent, especially at boundaries. Region-aware recalibration is thus best understood as a refinement over already useful features, not a substitute for representational power.

\paragraph{Semantic Segmentation}

Table~\ref{tab:voc-semantic-unet-results} extends the comparison to multiclass semantic segmentation with a pretrained ResNet-34 Semantic U-Net. On VOC2012, \please gives the best mIoU by a clear margin, improving the baseline from 73.44 to 75.05, with \mapse (DINO) second; the less flexible \base and \spase regions land slightly below baseline. On ADE20K, every module improves mIoU, led by \mapse (DINO) at 32.91 ahead of EMA and \mapse (GradCAM), but nearly all of them --- ours and prior work alike --- fall below the baseline Boundary F1 of 44.25, which only CA exceeds. Boundary quality thus trades against region overlap in the multi-class setting: a region-level gate sharpens which class is predicted inside a region more than it sharpens where that region ends.

This reinforces the capacity-dependent reading of the binary results. VOC2012, the semantic setting with substantially higher absolute mIoU, is where a fully learned partition pays off most clearly, whereas ADE20K, like ECSSD, stays much lower and no single module separates cleanly.

\paragraph{Mini-ImageNet Classification}

Table~\ref{tab:mini-imagenet-results} extends the comparison to image classification, where every variant improves over the ResNet-50 baseline: region-specific descriptors help image-level recognition even though classification does not require the pixel-level localization of segmentation. \spase is the strongest variant needing no pretrained teacher, reaching 77.87 top-1 and 92.77 top-5 for 86 added parameters ($+0.000\%$), against 77.46 and 92.49 for the best prior module, CA, at $+16.0\%$. \mapse (GradCAM) scores higher still (80.53 and 94.13), but with a caveat: its teacher is pretrained on ImageNet, whose label space contains all 100 mini-ImageNet classes, so here its regions carry evidence about the very classes being predicted --- an overlap absent from the segmentation experiments. \please stays competitive but is the most expensive variant ($+21.3\%$), since its $1{\times}1$ assignment network repeats at every residual block.


\section{Conclusion}
We introduced region-aware \se modules that replace \se's single global channel descriptor with region-specific ones, in four variants of increasing region flexibility: grid blocks (\base), superpixels (\spase), frozen semantic maps (\mapse), and learned soft assignments (\please). Every variant improves over the U-Net baseline on all binary benchmarks, and the benefit extends to semantic segmentation and classification, at costs that differ by variant rather than being uniformly negligible.

Regions inherited from a frozen general-purpose network are strongest across segmentation --- most binary entries and the best ADE20K mIoU --- suggesting that importing generic semantic structure into the squeeze step beats learning the partition from the task loss alone, while learned assignments pay off where the model has capacity headroom (\please gives the best VOC2012 mIoU) and expose interpretable regions for free. The clearest limitations are boundary quality on hard multi-class data and, for \mapse, a teacher whose label space may overlap the target task. Flexibility alone therefore does not predict performance: the region source is this framework's main design decision.


\FloatBarrier
{\small
\setlength{\bibsep}{2pt}

}
\clearpage
\setcounter{section}{0}
\setcounter{subsection}{0}
\setcounter{figure}{0}
\setcounter{table}{0}
\setcounter{equation}{0}
\renewcommand{\thesection}{S\arabic{section}}
\renewcommand{\thefigure}{S\arabic{figure}}
\renewcommand{\thetable}{S\arabic{table}}
\renewcommand{\theequation}{S\arabic{equation}}
\renewcommand{\theHsection}{supp.\arabic{section}}
\renewcommand{\theHfigure}{supp.\arabic{figure}}
\renewcommand{\theHtable}{supp.\arabic{table}}
\renewcommand{\theHequation}{supp.\arabic{equation}}
\renewcommand{\topfraction}{0.9}
\renewcommand{\bottomfraction}{0.8}
\renewcommand{\textfraction}{0.07}
\renewcommand{\floatpagefraction}{0.7}
\renewcommand{\dbltopfraction}{0.9}
\renewcommand{\dblfloatpagefraction}{0.7}
\setcounter{topnumber}{4}
\setcounter{bottomnumber}{3}
\setcounter{totalnumber}{6}
\setcounter{dbltopnumber}{4}
\lstset{style=pytorchcode,basicstyle=\ttfamily\codefigsize}
\twocolumn[{\centering\Large\bfseries Supplementary Material\par\vspace{1em}}]


\section{Experimental Setup}

\begin{figure*}[!t]
\centering
\includegraphics[width=0.86\textwidth]{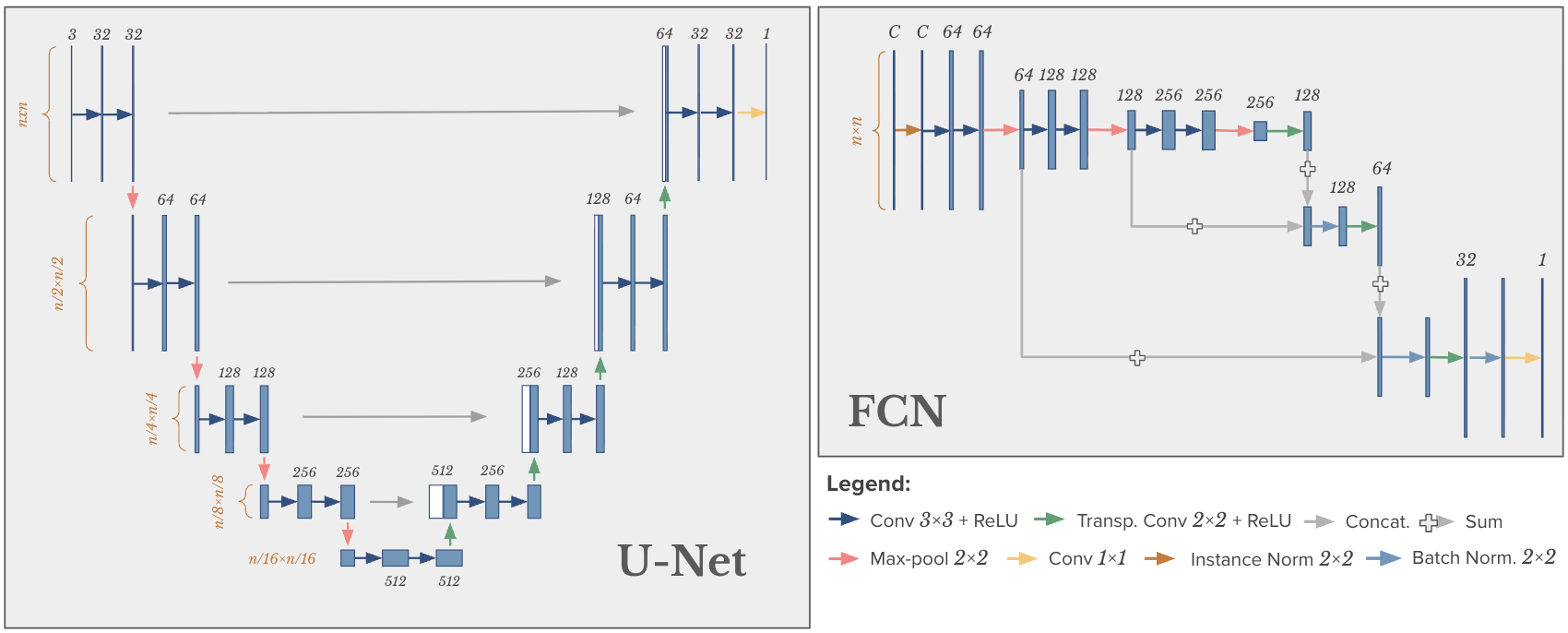}
\caption{Architectural details of the CNNs used for experimentation.}
\label{fig:CNNs}
\end{figure*}

Figure \ref{fig:CNNs} details the designs of the two convolutional architectures used for binary segmentation: the U-Net~\cite{ronneberger2015unet} and the Fully Convolutional Network~\cite{long2015fcn}.

\begin{figure}[!htbp]

\resizebox{\linewidth}{!}{%
\begin{tabular}{cccccc}
& \textbf{CUB} & \textbf{DUTS} & \textbf{ECSSD} & \textbf{VOC} & \textbf{ADE20K} \\[0.3em]

\raisebox{0.9cm}{\textbf{(a)}} &
\includegraphics[width=0.15\linewidth,height=1.8cm]{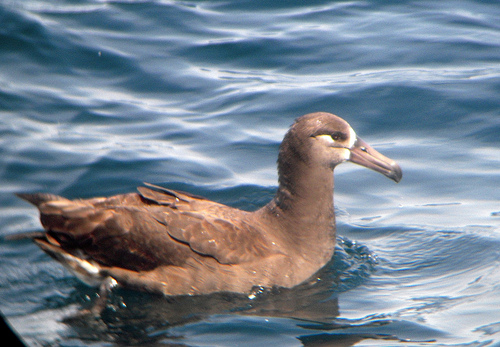} &
\includegraphics[width=0.15\linewidth,height=1.8cm]{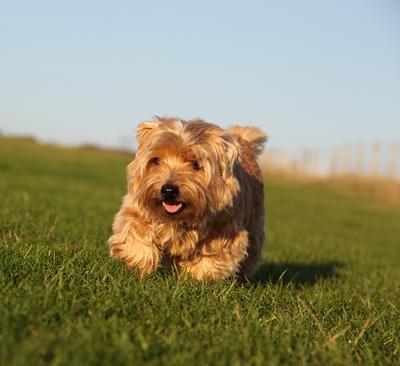} &
\includegraphics[width=0.15\linewidth,height=1.8cm]{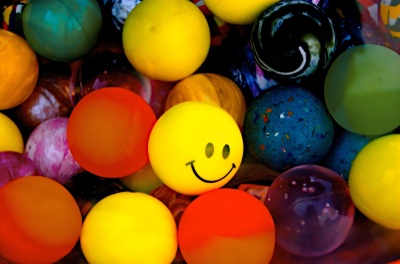} &
\includegraphics[width=0.15\linewidth,height=1.8cm]{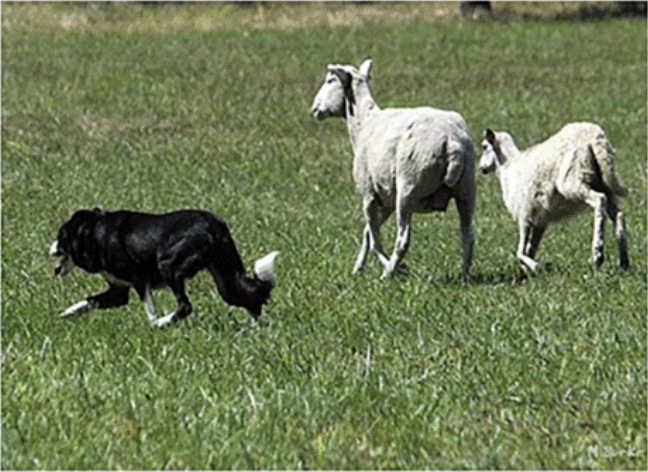} &
\includegraphics[width=0.15\linewidth,height=1.8cm]{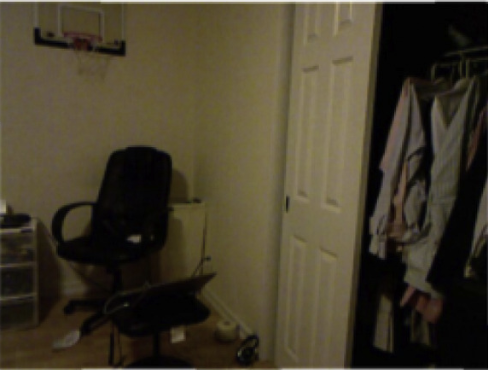} \\

\raisebox{0.9cm}{\textbf{(b)}} &
\includegraphics[width=0.15\linewidth,height=1.8cm]{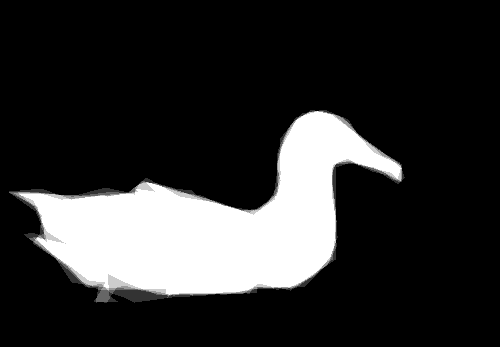} &
\includegraphics[width=0.15\linewidth,height=1.8cm]{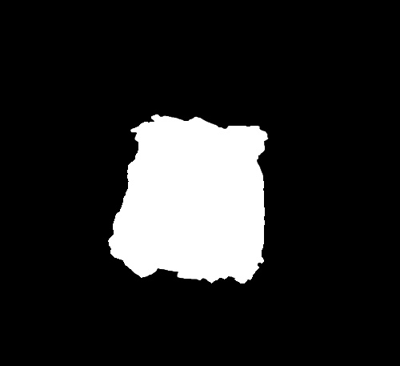} &
\includegraphics[width=0.15\linewidth,height=1.8cm]{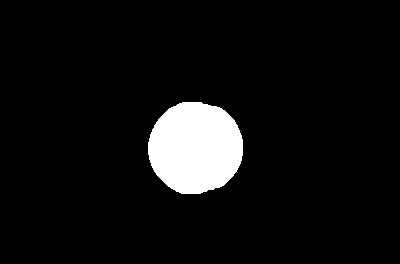} &
\includegraphics[width=0.15\linewidth,height=1.8cm]{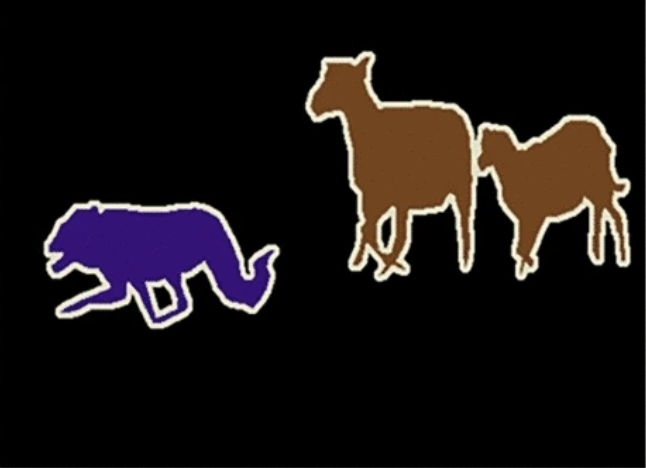} &
\includegraphics[width=0.15\linewidth,height=1.8cm]{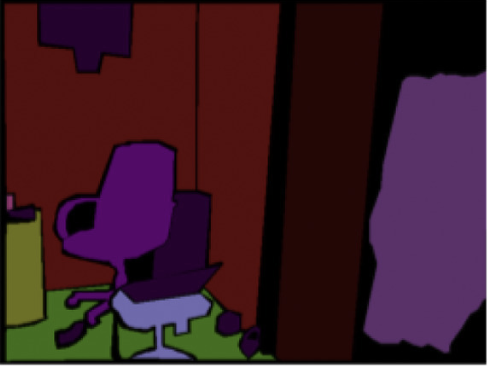} \\

\end{tabular}%
}

\vspace{0.7em}

\caption{Example images and segmentation masks from CUB, DUTS, ECSSD, PASCAL VOC 2012, and ADE20K. (a) Input images. (b) Ground truth masks.}
\label{fig:dataset_examples}
\end{figure}

Figure~\ref{fig:dataset_examples} shows an example input and ground truth image from the five segmentation datasets reported in the Experiments section of the main paper. CUB~\cite{wahCUB2011}, DUTS~\cite{wang2017}, and ECSSD~\cite{shi2015hierarchical} are binary segmentation datasets in which each pixel is assigned either a foreground or background label. PASCAL VOC 2012~\cite{everingham2010pascal} and ADE20K~\cite{zhou2017scene} are semantic segmentation datasets in which each pixel is assigned a semantic class label. PASCAL VOC 2012 contains 20 semantic classes, while ADE20K contains 150 semantic classes.

Figure~\ref{fig:code-modules} provides PyTorch implementations for each of our proposed blocks and demo code to train and evaluate each attention module on the U-Net architecture is also available with the supplementary materials.

\section{Implementation Details}
\label{sec:impl}

This section states the full experimental configuration. It repeats the
settings given in the Experiments section of the main paper so that the
supplement is self-contained, and adds the details of region construction that
the main paper has no space for. Demo code reproducing the DUTS U-Net
experiments accompanies this supplement.

\paragraph{Binary segmentation}
Experiments use the U-Net and FCN of Fig.~\ref{fig:CNNs}. For each of CUB,
DUTS, and ECSSD the image/mask pairs are pooled across the dataset's official
splits and re-partitioned $80/10/10$ into train, validation, and test sets by a
seeded shuffle, so that the partition is identical across methods and seeds.
Models train for 50 epochs on $224 \times 224$ inputs with AdamW (learning rate
$10^{-3}$, weight decay $10^{-4}$, batch size 16), using random crops and
horizontal flips ($p=0.5$); evaluation uses center crops with ImageNet
normalization. The objective is pixelwise binary cross-entropy on the logits,
averaged over all non-ignored pixels; no auxiliary Dice or boundary term is
used.

\paragraph{Semantic segmentation}
Experiments use an ImageNet-pretrained ResNet-34 Semantic U-Net with
dataset-specific prediction heads. VOC models train on the SBD-extended VOC
\textit{trainaug} split~\cite{hariharan2011semantic} and evaluate on the VOC2012
validation split; ADE20K models use the official train/validation splits.
Training uses random scaling in $[0.5, 2.0]$, padding, random $512 \times 512$
crops, horizontal flips, and ImageNet normalization; validation uses
full-resolution images.

\paragraph{Classification}
Experiments use ResNet-50 on the 100-class mini-ImageNet
dataset~\cite{vinyals2016matching}, with attention modules inserted into every
residual module before the residual addition, trained for 100 epochs with SGD
(learning rate $0.1$, momentum $0.9$, weight decay $10^{-4}$, batch size 128)
and a multistep schedule decayed by $0.1$ at epochs 30, 60, and 90. Training
uses random resized $224 \times 224$ crops and horizontal flips; validation and
testing resize to 256 then center-crop to $224 \times 224$, with ImageNet
normalization throughout.

\paragraph{Module placement}
Modules are inserted at the U-Net encoder activations after convolution
and before spatial downsampling; at the FCN skip feature tensors immediately
before decoder fusion; at the feature maps emitted by the pretrained ResNet-34
encoder of the Semantic U-Net prior to decoding; and at the ResNet-50 residual
branch output immediately before the identity addition.

\paragraph{Region construction}
\spase uses \texttt{fast-slic} on the raw RGB image at input resolution
with $K$ components and compactness $m = 10.0$; the resulting label map is
resized to each stage's resolution by nearest-neighbour sampling, and empty
regions are skipped. \mapse (GradCAM) uses a frozen torchvision ResNet-50 with
\texttt{IMAGENET1K\_V2} weights at $224 \times 224$ input. Writing $A$ for the
\texttt{layer4} output activations and $\mathbf{w}_k$ for the classifier row of
the $k$-th highest logit, the $k$-th map is
$\operatorname{ReLU}\!\left(\sum_c \mathbf{w}_{k,c} A_{c}\right)$, so the $K$
maps of a top-$K$ prediction need no backward pass; the stack is then passed
through a softmax over $K$. \mapse (DINO) uses a frozen native DINOv3 ViT-S/16
with the LVD-1689M checkpoint at $224 \times 224$ input. Its final normalized
patch tokens are mean-centred per image and decomposed by SVD; the leading $K$
components are given a deterministic sign (chosen so that each component's
largest-magnitude entry is positive) and the resulting projections are min-max
normalized per component. Both external priors are bilinearly resized to each
stage's resolution, clamped below at zero, and renormalized to sum to one across
the $K$ maps, with a uniform assignment substituted wherever the total mass
falls below $\varepsilon = 10^{-6}$. \please builds its assignment map with
$\operatorname{Conv2d}(c, \max(1, c/4), 1) \rightarrow \operatorname{ReLU}
\rightarrow \operatorname{Conv2d}(c/4, K, 1)$ followed by a softmax over $K$.

\paragraph{Excitation operator}
The Conv1D default follows ECA~\cite{wang2020eca}: a single
$\operatorname{Conv1d}(1, 1, k)$ without bias applied across the channel axis of
each region descriptor, with $k$ set adaptively from the channel count by truncating
$(\log_2 c + 1)/2$ to an integer and incrementing it when even, followed by a
sigmoid. The MLP variant
follows \se~\cite{hu2018squeeze}: a bottleneck $c \rightarrow c/r \rightarrow c$
with reduction ratio $r = 16$, a ReLU between the two layers, and a sigmoid on
the output.

\paragraph{Metrics and seeds}
mIoU is computed from a two-class confusion matrix accumulated over all
non-ignored pixels, and Dice/F1 from the same matrix. Boundary F1 marks every
pixel adjacent to a 4-neighbour label change as a boundary pixel and matches
predicted to ground-truth boundary pixels with a tolerance of 2
pixels~\cite{martin2004learning}. Every configuration is run with three seeds;
tables report the mean $\pm$ the sample standard deviation over those runs.

\section{Ablations}

Several configurations appear in more than one table below, because each ablation carries its own reference setting. These repeated entries are independent sets of three training runs rather than copies of a single run, so they differ from one another by seed-to-seed variation, and the observed differences fall within that variation. Where a repeated entry differs from the corresponding row of the main paper, the main paper reports the runs used for the headline comparison.

\paragraph{Attention Location and Granularity}

To identify the optimal location to insert our proposed attention modules we ran experiments inserting the \spase module into U-Net for binary segmentation of the DUTS dataset. These experiments were run with $K=8$ SLIC~\cite{achanta2012slic} superpixels, keeping regions few while giving SLIC more flexibility than the binary label count. Table~\ref{tab:duts_spase_location} shows that using all four encoder stages gives the highest mIoU and boundary F1 among the tested placements. The gains are not strictly monotonic with each additional shallow block, but the E1--E4 encoder configuration is stronger than the single-stage and partial-encoder variants and stronger than applying the module to all skip connections. If not noted otherwise, the U-Net models in the main paper apply the respective attention modules at the E1--E4 locations.

Table~\ref{tab:duts_spase_k} examines the effect of the $K$ hyperparameter on \spase for the same task. The effect of $K$ is analogous to controlling the number of regions across all methods within our framework, in this case the number of SLIC superpixels. Results show $K=8$ as the ``sweet spot'' for this task. Although the task is binary segmentation, it is intuitive that the model benefits from a slightly finer-grained regional decomposition than just two broad regions, while performance degrades when the attention becomes overly localized.

\begin{table}[t]
\centering
\footnotesize
\caption{DUTS attention location ablation for \spase, $K=8$, $\phi(\cdot) = \mathrm{MLP}$, U-Net.}
\label{tab:duts_spase_location}
\setlength{\tabcolsep}{3pt}
\begin{tabular}{llcc}
\toprule
Attention Location & Blocks & mIoU & Boundary F1 \\
\midrule
Encoder & E1       & 82.07 $\pm$ 0.41 & 59.75 $\pm$ 0.91 \\
Encoder & E1--E2   & 82.16 $\pm$ 0.10 & 60.24 $\pm$ 0.55 \\
Encoder & E1--E3   & 82.63 $\pm$ 0.56 & 60.66 $\pm$ 1.31 \\
Encoder & E1--E4   & \textbf{82.97} $\pm$ 0.70 & \textbf{61.21} $\pm$ 1.43 \\
Skip    & S1   & 81.97 $\pm$ 0.65 & 60.00 $\pm$ 0.95 \\
Skip    & S1--S2   & 82.09 $\pm$ 0.95 & 58.95 $\pm$ 1.28 \\
Skip    & S1--S3   & 81.84 $\pm$ 0.27 & 59.58 $\pm$ 0.54 \\
Skip    & S1--S4   & 82.16 $\pm$ 0.64 & 59.32 $\pm$ 0.70 \\
\bottomrule
\end{tabular}
\end{table}

\begin{table}[t]
\centering
\footnotesize
\caption{DUTS $K$ ablation results for \spase, E1--E4, $\phi(\cdot) = \mathrm{MLP}$, U-Net.}
\label{tab:duts_spase_k}
\setlength{\tabcolsep}{3pt}
\begin{tabular}{lcc}
\toprule
$K$ & mIoU & Boundary F1 \\
\midrule
2   & 82.88 $\pm$ 0.12 & 60.91 $\pm$ 0.55 \\
8   & \textbf{82.95} $\pm$ 0.19 & \textbf{61.02} $\pm$ 0.93 \\
16  & 82.61 $\pm$ 0.60 & 60.37 $\pm$ 1.53 \\
64  & 81.76 $\pm$ 0.38 & 59.09 $\pm$ 0.89 \\
128 & 81.87 $\pm$ 0.29 & 59.01 $\pm$ 0.42 \\
500 & 81.50 $\pm$ 0.76 & 58.11 $\pm$ 1.56 \\
\bottomrule
\end{tabular}
\end{table}

\paragraph{Choice of $\phi(\cdot)$}

We next identified the best method of \textit{excitation}, in particular which choice of $\phi(\cdot)$. The original SE paper~\cite{hu2018squeeze} proposed a fully connected dimensionality reduction layer (MLP) to fully capture the channel-wise dependencies emerging from the \textit{squeeze} operation. ECA-net~\cite{wang2020eca} offered an efficient alternative to $\phi(\cdot)$ proposing a local-cross-channel interaction without dimensionality reduction via a 1D convolution.

Table~\ref{tab:duts_excitation_ablation} compares these two approaches to excitation across our proposed methods via U-Net on the DUTS dataset. The results clearly identify the ECA-style 1D convolution as the superior approach, achieving better segmentation performance while carrying less computational overhead.

\begin{table*}[t]
\centering
\footnotesize
\caption{DUTS $\phi(\cdot)$ ablation for U-Net encoder E1--E4 modules. Metric cells report mIoU / Boundary F1. For the baseline row, the MLP setting is vanilla \se and the 1D conv setting is ECA. The better of the two settings in each row is in \textbf{bold}.}
\label{tab:duts_excitation_ablation}
\setlength{\tabcolsep}{4pt}
\resizebox{\textwidth}{!}{%
\begin{tabular}{lcccccc}
\toprule
& \multicolumn{3}{c}{MLP \cite{hu2018squeeze}} & \multicolumn{3}{c}{Conv1D \cite{wang2020eca}} \\
\cmidrule(lr){2-4}\cmidrule(lr){5-7}
Module & mIoU / Boundary F1 & Added params & Added params (\%) & mIoU / Boundary F1 & Added params & Added params (\%) \\
\midrule
Baseline (\se/\eca) & 82.41 $\pm$ 0.47 / 60.35 $\pm$ 0.81 & $11{,}390$ & $+0.147$ & \textbf{84.07} $\pm$ 0.39 / \textbf{63.63} $\pm$ 0.17 & $16$ & $+0.000$ \\
\base & 82.73 $\pm$ 0.77 / 60.27 $\pm$ 1.67 & $11{,}390$ & $+0.147$ & \textbf{83.28} $\pm$ 0.82 / \textbf{61.97} $\pm$ 1.30 & $16$ & $+0.000$ \\
\spase & 82.95 $\pm$ 0.19 / 61.02 $\pm$ 0.93 & $11{,}390$ & $+0.147$ & \textbf{83.51} $\pm$ 0.24 / \textbf{61.81} $\pm$ 0.83 & $16$ & $+0.000$ \\
\please & 83.08 $\pm$ 0.27 / 61.62 $\pm$ 0.70 & $34{,}142$ & $+0.440$ & \textbf{84.08} $\pm$ 0.63 / \textbf{62.78} $\pm$ 0.09 & $22{,}768$ & $+0.293$ \\
\mapse (GradCAM) & 84.60 $\pm$ 0.31 / 63.71 $\pm$ 0.93 & $11{,}390$ & $+0.147$ & \textbf{85.45} $\pm$ 0.49 / \textbf{65.67} $\pm$ 0.36 & $16$ & $+0.000$ \\
\mapse (DINO) & 86.51 $\pm$ 0.58 / 68.29 $\pm$ 0.68 & $11{,}390$ & $+0.147$ & \textbf{87.15} $\pm$ 0.44 / \textbf{68.82} $\pm$ 0.53 & $16$ & $+0.000$ \\
\bottomrule
\end{tabular}%
}
\end{table*}

\paragraph{Residual Connections}

Finally, we analyzed the choice to allow our modules to directly weight the input tensor or to enforce a residual connection. This is the difference between the learned channel and region-aware weights directly rescaling the input to the module or the output being the sum of this rescaled input and the original input.

\begin{table}[t]
\centering
\footnotesize
\caption{DUTS residual-connection ablation for the U-Net (E1--E4) and FCN (S1--S2). All modules use ECA excitation. Metric cells report mIoU / Boundary F1, and the better of the two settings in each row is in \textbf{bold}.}
\label{tab:duts_residual_ablation}
\setlength{\tabcolsep}{4pt}
\resizebox{\linewidth}{!}{%
\begin{tabular}{llcc}
\toprule
Backbone & Module & Residual False & Residual True \\
\midrule
U-Net & Baseline (ECA)  & \textbf{84.05} $\pm$ 0.35 / \textbf{63.59} $\pm$ 0.24 & 83.20 $\pm$ 0.91 / 62.00 $\pm$ 0.85 \\
U-Net & \base  & \textbf{83.26} $\pm$ 0.54 / \textbf{61.88} $\pm$ 1.12 & 82.53 $\pm$ 0.73 / 60.78 $\pm$ 1.38 \\
U-Net & \spase  & \textbf{83.59} $\pm$ 0.48 / 61.76 $\pm$ 0.56 & 83.36 $\pm$ 0.19 / \textbf{61.78} $\pm$ 0.35 \\
U-Net & \please  & \textbf{84.71} $\pm$ 0.58 / \textbf{64.76} $\pm$ 0.69 & 83.77 $\pm$ 0.64 / 62.67 $\pm$ 0.28 \\
FCN & Baseline (ECA)  & \textbf{73.70} $\pm$ 0.83 / \textbf{41.62} $\pm$ 0.58 & 73.33 $\pm$ 0.45 / 40.39 $\pm$ 0.17 \\
FCN & \base  & \textbf{72.46} $\pm$ 0.64 / \textbf{40.35} $\pm$ 0.68 & 71.54 $\pm$ 0.43 / 38.84 $\pm$ 0.56 \\
FCN & \spase  & \textbf{72.36} $\pm$ 0.34 / \textbf{41.78} $\pm$ 0.48 & 71.76 $\pm$ 0.51 / 40.88 $\pm$ 0.85 \\
FCN & \please  & \textbf{73.48} $\pm$ 0.57 / \textbf{41.46} $\pm$ 0.76 & 73.48 $\pm$ 0.51 / 41.12 $\pm$ 0.45 \\
\bottomrule
\end{tabular}%
}
\end{table}

Table~\ref{tab:duts_residual_ablation} identifies direct reweighting of the input as the stronger default, leaving the underlying architectures to apply residual connections where appropriate. This setting gives the best mIoU for every tested module-backbone pair, with the only boundary-F1 exception being a negligible \spase U-Net difference in favor of the residual variant. This design choice mirrors existing channel-attention formulations: standard \se~\cite{hu2018squeeze}, \eca~\cite{wang2020eca}, CA~\cite{hou2021coordinateattentionefficientmobile}, EMA~\cite{Ouyang_2023}, and CBAM~\cite{woo2018cbam} apply their learned attention weights directly through elementwise multiplication without an additional residual connection inside the attention block itself.

\paragraph{PLEASE-Specific Ablations}

Tables~\ref{tab:duts_please_k8_location} and~\ref{tab:duts_please_k_mlp_ablation} repeat the location and region-count ablations for \please. As with \spase, the E1--E4 encoder configuration is strongest. The region count behaves differently, however: \please peaks at $K=64$ learned soft regions rather than $K=8$, although the two overlap within run-to-run variation ($83.48 \pm 0.28$ against $82.87 \pm 0.81$ mIoU). We therefore report $K=8$ throughout the main paper, for consistency with \spase.

\begin{table}[t]
\centering
\footnotesize
\caption{DUTS attention location ablation using \please with $K=8$,
$\phi=\mathrm{MLP}$, no residual connection, and a U-Net backbone.}
\label{tab:duts_please_k8_location}
\setlength{\tabcolsep}{5pt}
\begin{tabular}{llcc}
\toprule
Location & Blocks & mIoU & Boundary F1 \\
\midrule
Encoder & E1       & 81.36 $\pm$ 0.67 & 58.71 $\pm$ 1.50 \\
Encoder & E1--E2   & 81.83 $\pm$ 0.17 & 60.20 $\pm$ 0.15 \\
Encoder & E1--E3   & 82.05 $\pm$ 0.21 & 60.53 $\pm$ 0.82 \\
Encoder & E1--E4   & \textbf{83.40} $\pm$ 0.63 & \textbf{62.40} $\pm$ 0.44 \\
Skip    & S1       & 81.75 $\pm$ 0.27 & 58.25 $\pm$ 0.65 \\
Skip    & S1--S2   & 82.12 $\pm$ 0.31 & 58.72 $\pm$ 0.23 \\
Skip    & S1--S3   & 81.92 $\pm$ 0.67 & 59.53 $\pm$ 0.68 \\
Skip    & S1--S4   & 82.47 $\pm$ 0.56 & 59.83 $\pm$ 0.67 \\
\bottomrule
\end{tabular}
\end{table}

\begin{table}[t]
\centering
\footnotesize
\caption{DUTS region-count ablation using \please at E1--E4 with
$\phi=\mathrm{MLP}$, no residual connection, and a U-Net backbone.}
\label{tab:duts_please_k_mlp_ablation}
\setlength{\tabcolsep}{6pt}
\begin{tabular}{lcc}
\toprule
$K$ & mIoU & Boundary F1 \\
\midrule
2   & 82.73 $\pm$ 1.09 & 61.38 $\pm$ 1.17 \\
8   & 82.87 $\pm$ 0.81 & 61.70 $\pm$ 0.76 \\
16  & 83.24 $\pm$ 0.82 & 61.52 $\pm$ 0.95 \\
64  & \textbf{83.48} $\pm$ 0.28 & \textbf{62.47} $\pm$ 0.33 \\
128 & 83.03 $\pm$ 0.45 & 61.80 $\pm$ 0.81 \\
500 & 82.71 $\pm$ 0.41 & 61.37 $\pm$ 0.77 \\
\bottomrule
\end{tabular}
\end{table}

The \mapse (GradCAM, TopK) region-count ablation uses class activation maps~\cite{selvaraju2017grad} from the
top-$K$ predictions of a frozen ImageNet-pretrained ResNet-50 teacher.
Table~\ref{tab:duts_gradcam_topk_mlp_k_ablation} reports test-set results over
three seeds for each setting. The $K=128$ setting is numerically best across
all three metrics, although $K=4$ performs similarly within the observed
run-to-run variation.

\begin{table}[t]
\centering
\footnotesize
\caption{DUTS \mapse (GradCAM, TopK) region-count ablation using class activation maps
from a frozen ImageNet-pretrained ResNet-50 teacher at E1--E4,
$\phi=\mathrm{MLP}$, no residual connection, and a U-Net backbone. Results are
mean $\pm$ sample standard deviation over three seeds.}
\label{tab:duts_gradcam_topk_mlp_k_ablation}
\setlength{\tabcolsep}{5pt}
\begin{tabular}{lccc}
\toprule
$K$ & mIoU & Dice/F1 & Boundary F1 \\
\midrule
2   & 83.98 $\pm$ 0.17 & 91.28 $\pm$ 0.10 & 62.49 $\pm$ 0.61 \\
4   & 84.54 $\pm$ 0.73 & 91.61 $\pm$ 0.43 & 63.44 $\pm$ 0.74 \\
8   & 84.28 $\pm$ 0.62 & 91.45 $\pm$ 0.37 & 62.95 $\pm$ 0.76 \\
16  & 83.85 $\pm$ 0.28 & 91.20 $\pm$ 0.17 & 62.15 $\pm$ 1.04 \\
64  & 84.18 $\pm$ 0.12 & 91.40 $\pm$ 0.07 & 63.16 $\pm$ 0.93 \\
128 & \textbf{84.60} $\pm$ 0.31 & \textbf{91.64} $\pm$ 0.19 & \textbf{63.71} $\pm$ 0.93 \\
\bottomrule
\end{tabular}
\end{table}

The \mapse (DINO) region-count ablation uses PCA assignments computed from the final
normalized DINOv3~\cite{simeoni2025dinov3} patch tokens. Table~\ref{tab:duts_dinose_final_mlp_k_ablation}
reports test-set results over three seeds for each setting. The $K=4$ setting
gives the best mIoU and Dice/F1, while $K=8$ gives the best Boundary F1.
Performance drops sharply for $K\geq16$.

\begin{table}[t]
\centering
\footnotesize
\caption{DUTS \mapse (DINO) region-count ablation using final normalized DINOv3
patch tokens at E1--E4, $\phi=\mathrm{MLP}$, no residual connection, and a
U-Net backbone. Results are mean $\pm$ sample standard deviation over three
seeds.}
\label{tab:duts_dinose_final_mlp_k_ablation}
\setlength{\tabcolsep}{5pt}
\begin{tabular}{lccc}
\toprule
$K$ & mIoU & Dice/F1 & Boundary F1 \\
\midrule
2   & 85.20 $\pm$ 0.17 & 91.99 $\pm$ 0.10 & 65.82 $\pm$ 1.10 \\
4   & \textbf{86.54} $\pm$ 0.70 & \textbf{92.77} $\pm$ 0.40 & 67.75 $\pm$ 0.52 \\
8   & 86.51 $\pm$ 0.58 & 92.75 $\pm$ 0.33 & \textbf{68.29} $\pm$ 0.68 \\
16  & 82.82 $\pm$ 0.45 & 90.58 $\pm$ 0.27 & 60.85 $\pm$ 0.54 \\
64  & 83.10 $\pm$ 0.50 & 90.75 $\pm$ 0.30 & 61.22 $\pm$ 0.55 \\
128 & 83.22 $\pm$ 0.24 & 90.82 $\pm$ 0.14 & 61.38 $\pm$ 0.35 \\
\bottomrule
\end{tabular}
\end{table}

The \mapse (DINO) embedding ablation compares PCA region assignments computed from
the final normalized DINOv3 patch tokens with assignments computed from the
key vectors in the final self-attention block. Table~\ref{tab:duts_dinose_embedding_ablation}
reports test-set results over three seeds for each setting. Final patch tokens
with $K=4$ perform best across all three metrics.

\begin{table}[t]
\centering
\footnotesize
\caption{DUTS \mapse (DINO) embedding-source and region-count ablation using
E1--E4, $\phi=\mathrm{Conv1D}$, no residual connection, and a U-Net backbone.
Results are mean $\pm$ sample standard deviation over three seeds.}
\label{tab:duts_dinose_embedding_ablation}
\setlength{\tabcolsep}{4pt}
\resizebox{\linewidth}{!}{%
\begin{tabular}{llccc}
\toprule
Embedding source & $K$ & mIoU & Dice/F1 & Boundary F1 \\
\midrule
Patch tokens & 2 & 85.74 $\pm$ 0.32 & 92.31 $\pm$ 0.19 & 66.54 $\pm$ 0.38 \\
Patch tokens & 4 & \textbf{86.77} $\pm$ 0.40 & \textbf{92.90} $\pm$ 0.23 & \textbf{68.18} $\pm$ 0.84 \\
Key features   & 2 & 85.56 $\pm$ 0.54 & 92.20 $\pm$ 0.32 & 65.47 $\pm$ 0.13 \\
Key features  & 4 & 85.60 $\pm$ 0.06 & 92.23 $\pm$ 0.04 & 66.34 $\pm$ 0.63 \\
\bottomrule
\end{tabular}%
}
\end{table}

\section{FCN Quantitative Results}

Table~\ref{tab:fcn-results} extends the U-Net binary segmentation comparison from the main paper to the FCN architecture, with attention applied to skip connections S1--S2 and all other settings left at the defaults of the main paper.

 The FCN reproduces the U-Net picture on CUB and DUTS and flattens on ECSSD. On CUB and DUTS every module, ours and prior work alike, improves over the attention-free baseline, and the proposed modules take all six best entries in the table: \mapse (GradCAM) and \spase on CUB, \mapse (DINO) on DUTS --- where the margin over the baseline is largest, $76.40$ against $66.39$ mIoU --- and \spase on ECSSD.

 The ECSSD column deserves comment, because several modules there sit below the baseline. Under a Welch $t$-test over the three seeds, no module differs significantly from the baseline at the $5\%$ level except \cbam, which is significantly worse on both metrics, and \mapse (GradCAM), which is significantly worse on Boundary F1; \spase improves both ECSSD metrics, but that improvement is itself within noise. We therefore read this column as evidence that region-aware recalibration neither helps nor hurts in this regime, rather than as evidence that it is harmful.

 Two properties of the setting account for the flattening. First, ECSSD is by far the smallest of the three datasets at $1{,}000$ images, leaving roughly $100$ test images under the 80/10/10 protocol against ${\sim}1{,}200$ for CUB and ${\sim}1{,}600$ for DUTS. Seed variation on ECSSD is correspondingly inflated --- $1.42$ mIoU on average, against $0.55$ on CUB and $0.63$ on DUTS --- which leaves little resolution to separate closely matched methods. Second, the FCN applies attention at the S1--S2 skip tensors, which are shallow and high-resolution; descriptors pooled there summarize appearance rather than semantics, so the region source matters less than it does at the U-Net's E1--E4 encoder stages. Consistent with this, the largest ECSSD losses fall to the modules whose regions are most strongly structured: \please, whose assignment network carries the most added parameters on the smallest dataset, and \mapse (DINO), whose partition is inherited from a teacher operating at a different resolution and semantic level.

\begin{table*}[t]
\centering
\footnotesize
\caption{FCN quantitative comparison across the binary segmentation datasets. All values are percentages. Best results are in \textbf{bold}, second best \underline{underlined}.}
\label{tab:fcn-results}
\setlength{\tabcolsep}{4pt}
\renewcommand{\arraystretch}{1.15}
\resizebox{\textwidth}{!}{%
\begin{tabular}{lccccccc}
\toprule
& & \multicolumn{2}{c}{CUB} & \multicolumn{2}{c}{DUTS} & \multicolumn{2}{c}{ECSSD} \\
\cmidrule(lr){3-4} \cmidrule(lr){5-6} \cmidrule(lr){7-8}
Model & Added params & mIoU & Boundary F1 & mIoU & Boundary F1 & mIoU & Boundary F1 \\
\midrule
Baseline
& $0.000$
& 82.06 $\pm$ 1.08 & 54.12 $\pm$ 0.80
& 66.39 $\pm$ 0.02 & 35.04 $\pm$ 0.78
& 62.54 $\pm$ 0.70 & \underline{29.97} $\pm$ 1.24
\\

SE
& $+0.180$
& 84.04 $\pm$ 1.07 & 56.18 $\pm$ 0.87
& 70.91 $\pm$ 0.70 & 39.30 $\pm$ 0.29
& 61.63 $\pm$ 0.91 & 28.32 $\pm$ 1.49
\\

\scse
& $+0.193$
& 84.18 $\pm$ 0.15 & 56.31 $\pm$ 0.73
& 71.34 $\pm$ 0.89 & 39.35 $\pm$ 1.80
& 61.68 $\pm$ 1.45 & 29.17 $\pm$ 2.31
\\

ECA
& $+0.001$
& 83.93 $\pm$ 0.42 & 55.35 $\pm$ 1.28
& 73.10 $\pm$ 0.77 & 41.34 $\pm$ 0.66
& 61.71 $\pm$ 2.07 & 29.17 $\pm$ 2.12
\\

CA
& $+0.328$
& 84.08 $\pm$ 0.63 & 56.25 $\pm$ 0.55
& 71.20 $\pm$ 0.77 & 37.47 $\pm$ 0.45
& 60.83 $\pm$ 1.54 & 27.72 $\pm$ 1.08
\\

EMA
& $+0.015$
& 84.93 $\pm$ 0.21 & 56.49 $\pm$ 0.55
& 72.17 $\pm$ 1.52 & 38.84 $\pm$ 1.14
& 62.19 $\pm$ 1.06 & 29.38 $\pm$ 1.92
\\

CBAM
& $+0.193$
& 83.58 $\pm$ 0.65 & 55.02 $\pm$ 0.88
& 70.57 $\pm$ 0.46 & 37.96 $\pm$ 1.06
& 58.68 $\pm$ 0.25 & 24.99 $\pm$ 2.33
\\\midrule

\mapse (GradCAM)
& $+0.001$
& \textbf{86.76} $\pm$ 0.10 & 58.14 $\pm$ 0.15
& \underline{73.82} $\pm$ 0.31 & 38.51 $\pm$ 0.24
& \underline{62.94} $\pm$ 2.01 & 25.79 $\pm$ 1.33
\\

\mapse (DINO)
& $+0.001$
& \underline{85.50} $\pm$ 0.83 & \underline{58.75} $\pm$ 0.98
& \textbf{76.40} $\pm$ 0.61 & \textbf{44.34} $\pm$ 0.88
& 60.24 $\pm$ 2.51 & 27.76 $\pm$ 3.00
\\

\base
& $+0.001$
& 84.55 $\pm$ 0.53 & 57.38 $\pm$ 0.65
& 72.46 $\pm$ 0.64 & 40.35 $\pm$ 0.68
& 62.02 $\pm$ 1.17 & 29.01 $\pm$ 1.35
\\

\spase
& $+0.001$
& 85.33 $\pm$ 0.36 & \textbf{59.01} $\pm$ 0.65
& 72.36 $\pm$ 0.34 & \underline{41.78} $\pm$ 0.48
& \textbf{63.29} $\pm$ 1.44 & \textbf{31.77} $\pm$ 1.45
\\

\please
& $+0.361$
& 85.34 $\pm$ 0.57 & 57.55 $\pm$ 0.17
& 73.48 $\pm$ 0.57 & 41.46 $\pm$ 0.76
& 59.77 $\pm$ 1.95 & 28.01 $\pm$ 1.73
\\
\bottomrule
\end{tabular}%
}
\end{table*}

\section{Runtime Analysis}

Tables~\ref{tab:fcn-complexity}, \ref{tab:semantic-unet-complexity}, and \ref{tab:resnet-complexity} provide the runtime analyses for the FCN, pretrained ResNet-34 Semantic U-Net, and ResNet-50 architectures~\cite{DBLP:journals/corr/HeZRS15}. As in the main paper, latency is reported in milliseconds per image over 200 measured iterations after 50 warmup iterations, measured in FP32 on a single NVIDIA RTX 3090 with dual AMD EPYC 7532 CPUs (128 threads). The findings are consistent with the U-Net results reported there: \spase introduces minimal additional parameters but incurs higher inference latency due to the SLIC superpixel computation, while \please replaces the non-differentiable preprocessing step with a fully end-to-end learnable soft region assignment module at the cost of additional parameters and computation. For the \mapse (GradCAM) and \mapse (DINO) variants, parameter counts include only the trainable student model, while GFLOP totals include the respective frozen ResNet-50 and DINOv3 backbones. End-to-end latency additionally includes CAM generation or DINO feature extraction and PCA assignment construction.

\begin{table}[t]
\centering
\footnotesize
\caption{FCN complexity and inference latency at input size $1{\times}3{\times}224{\times}224$, with attention applied to the two skip connections.}
\label{tab:fcn-complexity}
\setlength{\tabcolsep}{4pt}
\renewcommand{\arraystretch}{1.15}
\resizebox{\linewidth}{!}{%
\begin{tabular}{lrrrrrr}
\toprule
Model & Params & Added params & GFLOPs & Added GFLOPs & CPU latency & GPU latency \\
\midrule
Baseline & 1.533M & 0 & 16.399 & 0.000 & 222.07 & 1.29 \\
SE  & 1.536M & +2,764 & 16.400 & +0.001 & 213.37 & 1.33 \\
\scse  & 1.536M & +2,958 & 16.403 & +0.004 & 219.19 & 1.47 \\
ECA  & 1.533M & +8 & 16.400 & +0.001 & 225.06 & 1.35 \\
CA  & 1.538M & +5,024 & 16.400 & +0.001 & 267.28 & 1.94 \\
EMA & 1.533M & +224 & 16.471 & +0.072 & 236.37 & 2.00 \\
CBAM  & 1.536M & +2,960 & 16.404 & +0.005 & 229.22 & 1.94 \\
\mapse (GradCAM) & 1.533M & +8 & 24.649 & +8.250 & 361.83 & 7.72 \\
\mapse (DINO) & 1.533M & +8 & 25.102 & +8.704 & 398.40 & 22.49 \\
\midrule
\base  & 1.533M & +8 & 16.399 & +0.000 & 219.39 & 1.82 \\
\spase  & 1.533M & +8 & 16.399 & +0.000 & 229.97 & 6.48 \\
\please  & 1.539M & +5,528 & 16.494 & +0.095 & 226.43 & 1.75 \\
\bottomrule
\end{tabular}%
}
\end{table}

\begin{table}[t]
\centering
\footnotesize
\caption{Semantic U-Net complexity and inference latency at input size $1{\times}3{\times}512{\times}512$ with a 21-class output head.}
\label{tab:semantic-unet-complexity}
\setlength{\tabcolsep}{4pt}
\renewcommand{\arraystretch}{1.15}
\resizebox{\linewidth}{!}{%
\begin{tabular}{lrrrrrr}
\toprule
Model & Params & Added params & GFLOPs & Added GFLOPs & CPU latency & GPU latency \\
\midrule
Baseline & 24.437M & 0 & 62.733 & 0.000 & 732.67 & 5.10 \\
SE  & 24.448M & +11,808 & 62.739 & +0.006 & 738.71 & 5.26 \\
\scse  & 24.449M & +12,324 & 62.751 & +0.018 & 736.22 & 5.45 \\
ECA  & 24.437M & +16 & 62.739 & +0.006 & 738.05 & 5.23 \\
CA  & 24.456M & +19,536 & 62.736 & +0.003 & 1008.59 & 6.97 \\
EMA & 24.438M & +944 & 63.070 & +0.337 & 889.31 & 6.49 \\
CBAM  & 24.449M & +12,200 & 62.762 & +0.029 & 755.44 & 9.42 \\
\mapse (GradCAM) & 24.437M & +16 & 71.137 & +8.404 & 1144.90 & 14.68 \\
\mapse (DINO) & 24.437M & +16 & 71.591 & +8.858 & 1403.25 & 33.12 \\
\midrule
\base  & 24.437M & +16 & 62.739 & +0.006 & 802.67 & 6.04 \\
\spase  & 24.437M & +16 & 62.733 & +0.000 & 798.21 & 24.01 \\
\please  & 24.460M & +23,600 & 63.188 & +0.455 & 901.64 & 6.00 \\
\bottomrule
\end{tabular}%
}
\end{table}

\begin{table}[t]
\centering
\footnotesize
\caption{ResNet-50 mini-ImageNet~\cite{vinyals2016matching} complexity and inference latency at input size $1{\times}3{\times}224{\times}224$ with a 100-class output head.}
\label{tab:resnet-complexity}
\setlength{\tabcolsep}{4pt}
\renewcommand{\arraystretch}{1.15}
\resizebox{\linewidth}{!}{%
\begin{tabular}{lrrrrrr}
\toprule
Model & Params & Added params & GFLOPs & Added GFLOPs & CPU latency & GPU latency \\
\midrule
Baseline & 23.713M & 0 & 8.207 & 0.000 & 142.65 & 5.20 \\
SE  & 26.244M & +2,530,992 & 8.218 & +0.011 & 147.99 & 9.07 \\
\scse & 26.259M & +2,546,112 & 8.229 & +0.022 & 169.87 & 10.41 \\
ECA & 23.713M & +86 & 8.213 & +0.006 & 156.72 & 8.02 \\
CA & 27.517M & +3,804,512 & 8.316 & +0.109 & 160.00 & 12.59 \\
EMA & 23.911M & +198,368 & 10.151 & +1.943 & 205.79 & 13.88 \\
CBAM & 26.245M & +2,532,560 & 8.231 & +0.024 & 175.17 & 12.52 \\
\mapse (GradCAM) & 23.713M & +86 & 16.597 & +8.389 & 297.15 & 21.69 \\
\mapse (DINO) & 23.713M & +86 & 17.050 & +8.843 & 309.64 & 32.22 \\
\midrule
\base & 23.713M & +86 & 8.214 & +0.006 & 192.01 & 9.13 \\
\spase & 23.713M & +86 & 8.209 & +0.001 & 213.03 & 19.04 \\
\please & 28.773M & +5,060,310 & 10.053 & +1.846 & 194.75 & 12.55 \\
\bottomrule
\end{tabular}%
}
\end{table}

\begin{table*}[p]
\section{Model Interpretability}

Tables~\ref{tab:please_good_interpretability_visuals} and \ref{tab:please_bad_interpretability_visuals} provide additional interpretability visualizations not included in the main paper. These examples further support the observation that successful \please segmentations exhibit progressively more coherent and semantically aligned region structure across encoder depth, while poor segmentations remain fragmented and unstable throughout the network. In particular, the PCA visualizations for successful predictions often evolve from coarse appearance-driven partitions in early layers toward regions that better reflect the supervision objective in deeper layers, whereas poor predictions typically fail to converge to coherent semantic structure.
\medskip
\centering
\scriptsize
\setlength{\tabcolsep}{1.5pt}

\caption{Interpretability visualizations for good segmentation predictions, selected from the top 5\% of examples by mIoU for each dataset. Each dataset is shown with two representative examples. CUB/DUTS/ECSSD use U-Net, while VOC and ADE20K use Semantic U-Net. Columns show the input, ground truth, prediction, and argmax/PCA visualizations for encoder layers 1--4.}
\vspace{10pt}

\resizebox{\textwidth}{!}{%
\begin{tabular}{cccccccccccc}
\toprule
Dataset & Input & GT & Prediction & E1 argmax & E1 PCA & E2 argmax & E2 PCA & E3 argmax & E3 PCA & E4 argmax & E4 PCA \\
\midrule

\multirow{2}{*}{\makebox[0.06\textwidth][c]{CUB}} &
\includegraphics[width=0.075\textwidth,height=0.50in,keepaspectratio]{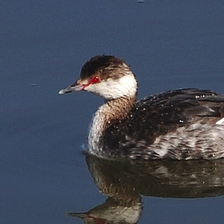} &
\includegraphics[width=0.075\textwidth,height=0.50in,keepaspectratio]{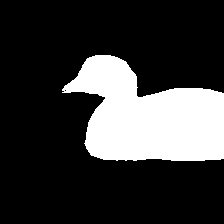} &
\includegraphics[width=0.075\textwidth,height=0.50in,keepaspectratio]{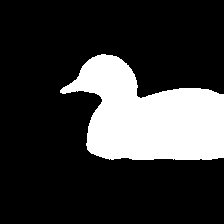} &
\includegraphics[width=0.075\textwidth,height=0.50in,keepaspectratio]{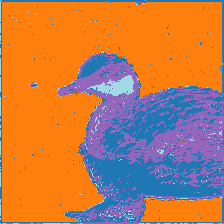} &
\includegraphics[width=0.075\textwidth,height=0.50in,keepaspectratio]{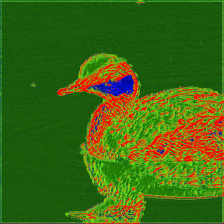} &
\includegraphics[width=0.075\textwidth,height=0.50in,keepaspectratio]{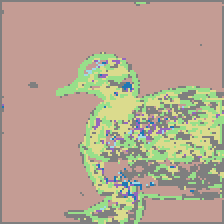} &
\includegraphics[width=0.075\textwidth,height=0.50in,keepaspectratio]{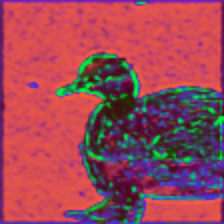} &
\includegraphics[width=0.075\textwidth,height=0.50in,keepaspectratio]{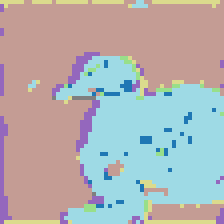} &
\includegraphics[width=0.075\textwidth,height=0.50in,keepaspectratio]{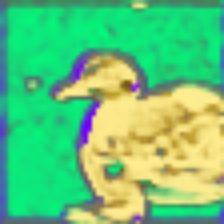} &
\includegraphics[width=0.075\textwidth,height=0.50in,keepaspectratio]{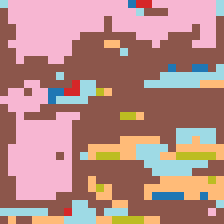} &
\includegraphics[width=0.075\textwidth,height=0.50in,keepaspectratio]{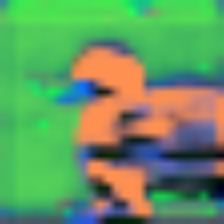} \\
&
\includegraphics[width=0.075\textwidth,height=0.50in,keepaspectratio]{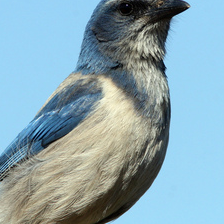} &
\includegraphics[width=0.075\textwidth,height=0.50in,keepaspectratio]{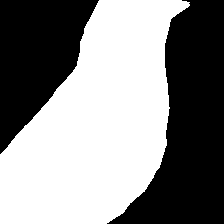} &
\includegraphics[width=0.075\textwidth,height=0.50in,keepaspectratio]{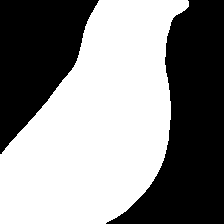} &
\includegraphics[width=0.075\textwidth,height=0.50in,keepaspectratio]{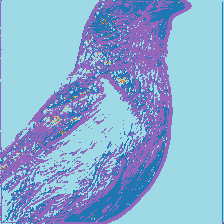} &
\includegraphics[width=0.075\textwidth,height=0.50in,keepaspectratio]{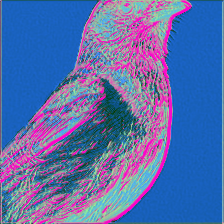} &
\includegraphics[width=0.075\textwidth,height=0.50in,keepaspectratio]{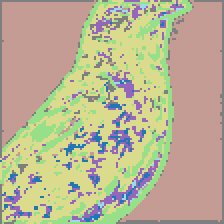} &
\includegraphics[width=0.075\textwidth,height=0.50in,keepaspectratio]{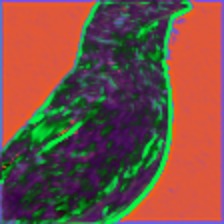} &
\includegraphics[width=0.075\textwidth,height=0.50in,keepaspectratio]{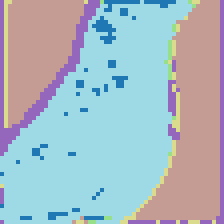} &
\includegraphics[width=0.075\textwidth,height=0.50in,keepaspectratio]{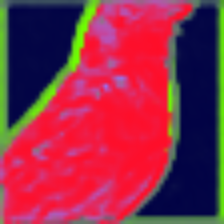} &
\includegraphics[width=0.075\textwidth,height=0.50in,keepaspectratio]{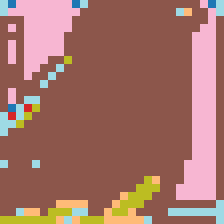} &
\includegraphics[width=0.075\textwidth,height=0.50in,keepaspectratio]{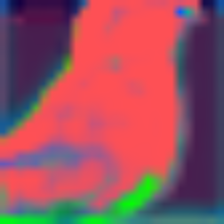} \\
\cmidrule(lr){1-12}

\multirow{2}{*}{\makebox[0.06\textwidth][c]{DUTS}} &
\includegraphics[width=0.075\textwidth,height=0.50in,keepaspectratio]{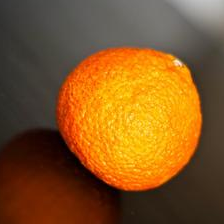} &
\includegraphics[width=0.075\textwidth,height=0.50in,keepaspectratio]{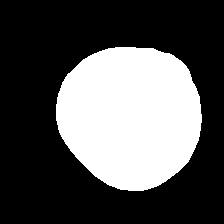} &
\includegraphics[width=0.075\textwidth,height=0.50in,keepaspectratio]{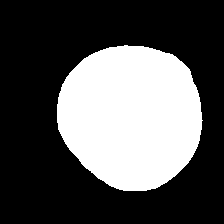} &
\includegraphics[width=0.075\textwidth,height=0.50in,keepaspectratio]{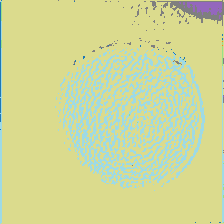} &
\includegraphics[width=0.075\textwidth,height=0.50in,keepaspectratio]{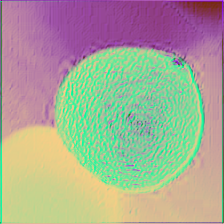} &
\includegraphics[width=0.075\textwidth,height=0.50in,keepaspectratio]{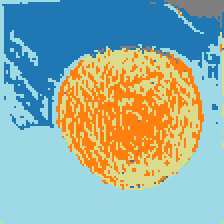} &
\includegraphics[width=0.075\textwidth,height=0.50in,keepaspectratio]{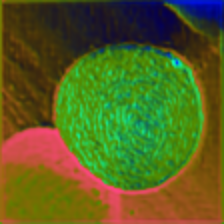} &
\includegraphics[width=0.075\textwidth,height=0.50in,keepaspectratio]{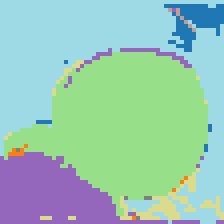} &
\includegraphics[width=0.075\textwidth,height=0.50in,keepaspectratio]{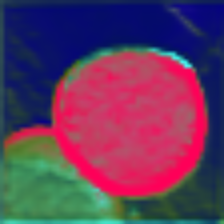} &
\includegraphics[width=0.075\textwidth,height=0.50in,keepaspectratio]{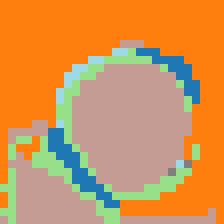} &
\includegraphics[width=0.075\textwidth,height=0.50in,keepaspectratio]{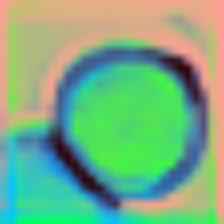} \\
&
\includegraphics[width=0.075\textwidth,height=0.50in,keepaspectratio]{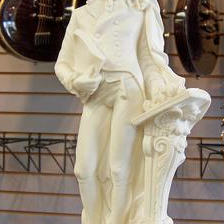} &
\includegraphics[width=0.075\textwidth,height=0.50in,keepaspectratio]{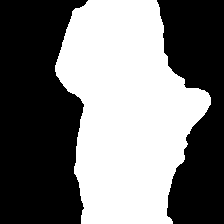} &
\includegraphics[width=0.075\textwidth,height=0.50in,keepaspectratio]{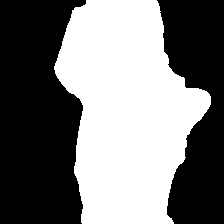} &
\includegraphics[width=0.075\textwidth,height=0.50in,keepaspectratio]{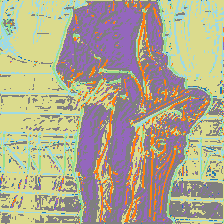} &
\includegraphics[width=0.075\textwidth,height=0.50in,keepaspectratio]{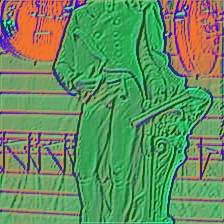} &
\includegraphics[width=0.075\textwidth,height=0.50in,keepaspectratio]{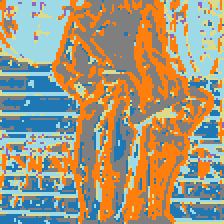} &
\includegraphics[width=0.075\textwidth,height=0.50in,keepaspectratio]{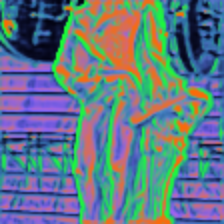} &
\includegraphics[width=0.075\textwidth,height=0.50in,keepaspectratio]{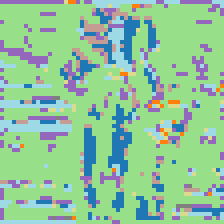} &
\includegraphics[width=0.075\textwidth,height=0.50in,keepaspectratio]{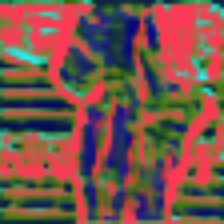} &
\includegraphics[width=0.075\textwidth,height=0.50in,keepaspectratio]{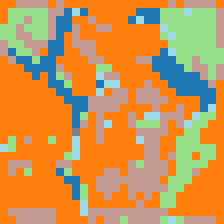} &
\includegraphics[width=0.075\textwidth,height=0.50in,keepaspectratio]{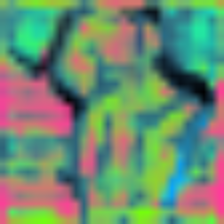} \\
\cmidrule(lr){1-12}

\multirow{2}{*}{\makebox[0.06\textwidth][c]{ECSSD}} &
\includegraphics[width=0.075\textwidth,height=0.50in,keepaspectratio]{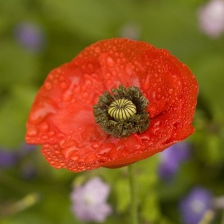} &
\includegraphics[width=0.075\textwidth,height=0.50in,keepaspectratio]{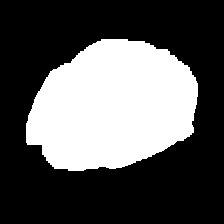} &
\includegraphics[width=0.075\textwidth,height=0.50in,keepaspectratio]{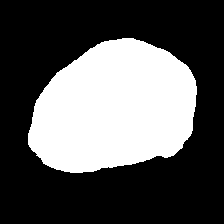} &
\includegraphics[width=0.075\textwidth,height=0.50in,keepaspectratio]{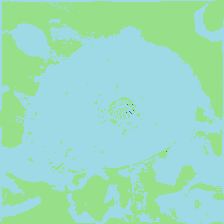} &
\includegraphics[width=0.075\textwidth,height=0.50in,keepaspectratio]{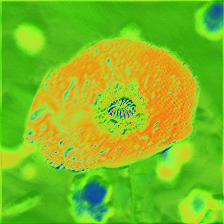} &
\includegraphics[width=0.075\textwidth,height=0.50in,keepaspectratio]{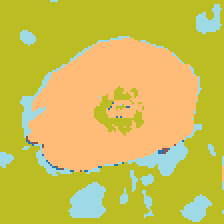} &
\includegraphics[width=0.075\textwidth,height=0.50in,keepaspectratio]{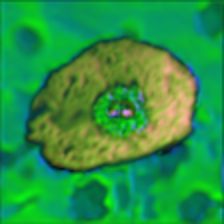} &
\includegraphics[width=0.075\textwidth,height=0.50in,keepaspectratio]{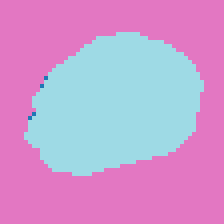} &
\includegraphics[width=0.075\textwidth,height=0.50in,keepaspectratio]{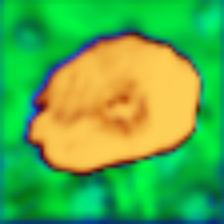} &
\includegraphics[width=0.075\textwidth,height=0.50in,keepaspectratio]{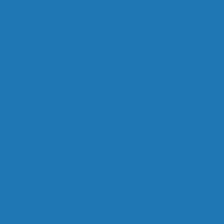} &
\includegraphics[width=0.075\textwidth,height=0.50in,keepaspectratio]{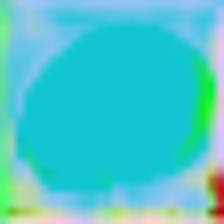} \\
&
\includegraphics[width=0.075\textwidth,height=0.50in,keepaspectratio]{figures/images_please_interp_ecssd_unet_good_sample_2_input.png} &
\includegraphics[width=0.075\textwidth,height=0.50in,keepaspectratio]{figures/images_please_interp_ecssd_unet_good_sample_2_ground_truth.png} &
\includegraphics[width=0.075\textwidth,height=0.50in,keepaspectratio]{figures/images_please_interp_ecssd_unet_good_sample_2_prediction.png} &
\includegraphics[width=0.075\textwidth,height=0.50in,keepaspectratio]{figures/images_please_interp_ecssd_unet_good_sample_2_encoder_1_argmax.png} &
\includegraphics[width=0.075\textwidth,height=0.50in,keepaspectratio]{figures/images_please_interp_ecssd_unet_good_sample_2_encoder_1_pca.png} &
\includegraphics[width=0.075\textwidth,height=0.50in,keepaspectratio]{figures/images_please_interp_ecssd_unet_good_sample_2_encoder_2_argmax.png} &
\includegraphics[width=0.075\textwidth,height=0.50in,keepaspectratio]{figures/images_please_interp_ecssd_unet_good_sample_2_encoder_2_pca.png} &
\includegraphics[width=0.075\textwidth,height=0.50in,keepaspectratio]{figures/images_please_interp_ecssd_unet_good_sample_2_encoder_3_argmax.png} &
\includegraphics[width=0.075\textwidth,height=0.50in,keepaspectratio]{figures/images_please_interp_ecssd_unet_good_sample_2_encoder_3_pca.png} &
\includegraphics[width=0.075\textwidth,height=0.50in,keepaspectratio]{figures/images_please_interp_ecssd_unet_good_sample_2_encoder_4_argmax.png} &
\includegraphics[width=0.075\textwidth,height=0.50in,keepaspectratio]{figures/images_please_interp_ecssd_unet_good_sample_2_encoder_4_pca.png} \\
\cmidrule(lr){1-12}

\multirow{2}{*}{\makebox[0.06\textwidth][c]{VOC}} &
\includegraphics[width=0.075\textwidth,height=0.50in,keepaspectratio]{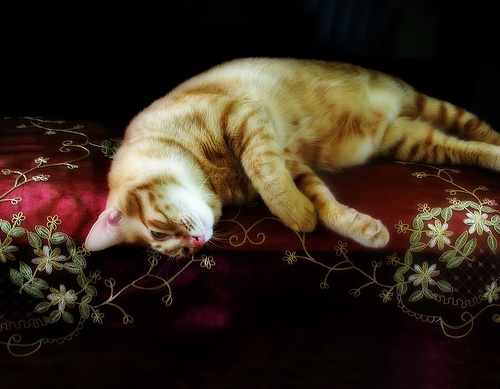} &
\includegraphics[width=0.075\textwidth,height=0.50in,keepaspectratio]{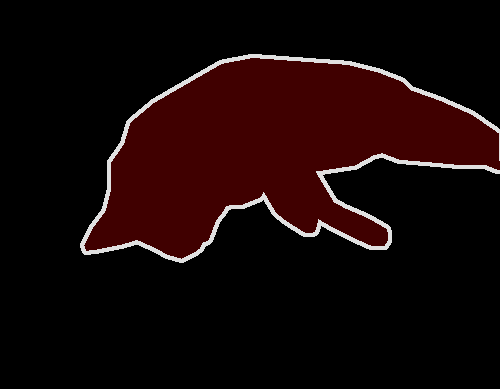} &
\includegraphics[width=0.075\textwidth,height=0.50in,keepaspectratio]{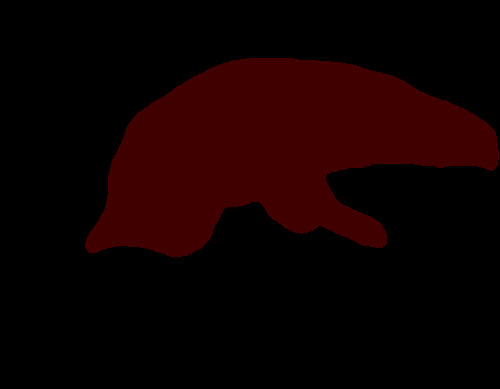} &
\includegraphics[width=0.075\textwidth,height=0.50in,keepaspectratio]{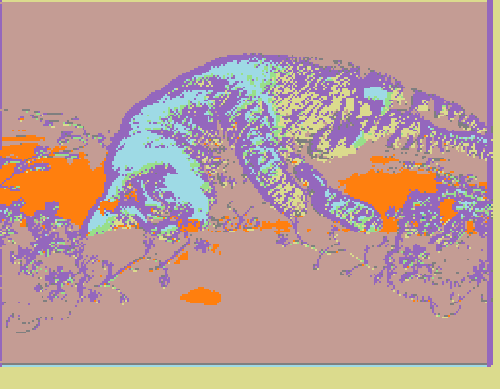} &
\includegraphics[width=0.075\textwidth,height=0.50in,keepaspectratio]{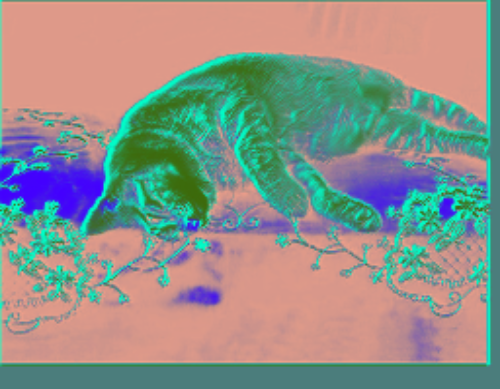} &
\includegraphics[width=0.075\textwidth,height=0.50in,keepaspectratio]{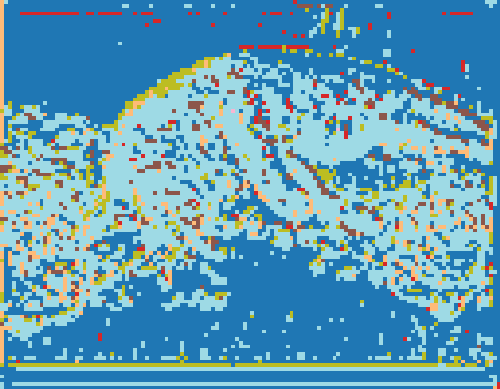} &
\includegraphics[width=0.075\textwidth,height=0.50in,keepaspectratio]{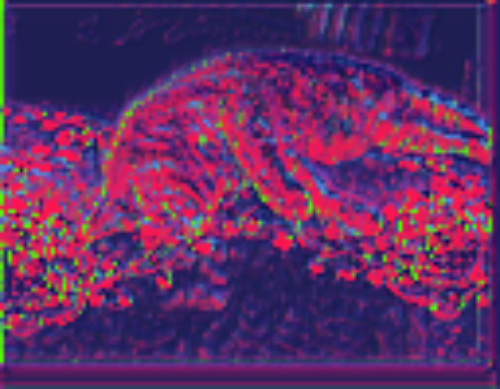} &
\includegraphics[width=0.075\textwidth,height=0.50in,keepaspectratio]{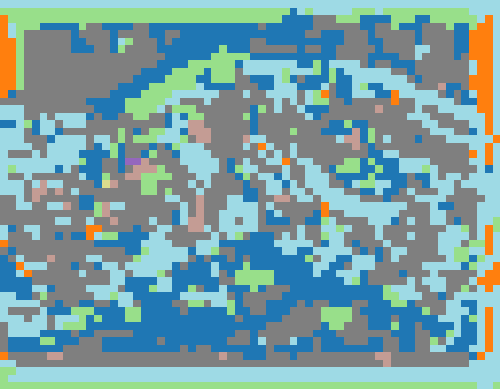} &
\includegraphics[width=0.075\textwidth,height=0.50in,keepaspectratio]{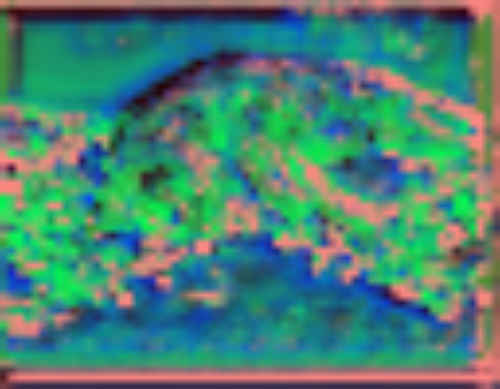} &
\includegraphics[width=0.075\textwidth,height=0.50in,keepaspectratio]{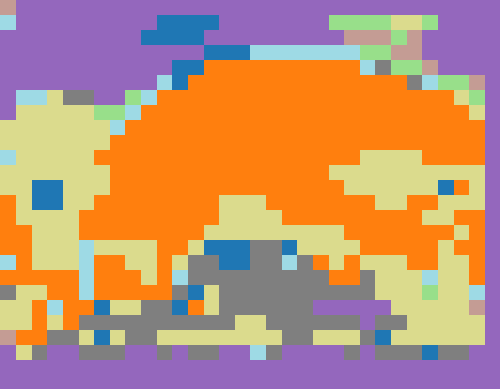} &
\includegraphics[width=0.075\textwidth,height=0.50in,keepaspectratio]{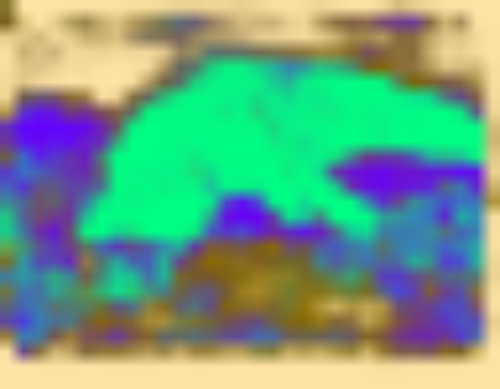} \\
&
\includegraphics[width=0.075\textwidth,height=0.50in,keepaspectratio]{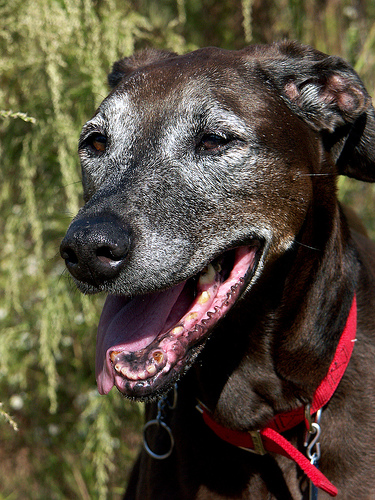} &
\includegraphics[width=0.075\textwidth,height=0.50in,keepaspectratio]{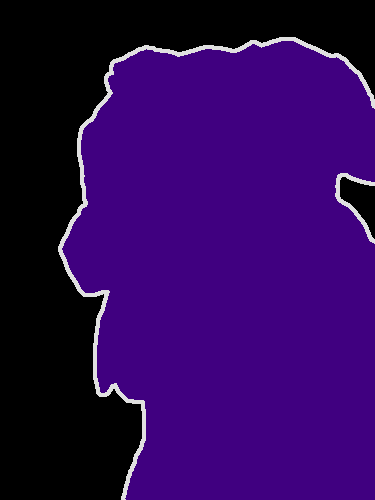} &
\includegraphics[width=0.075\textwidth,height=0.50in,keepaspectratio]{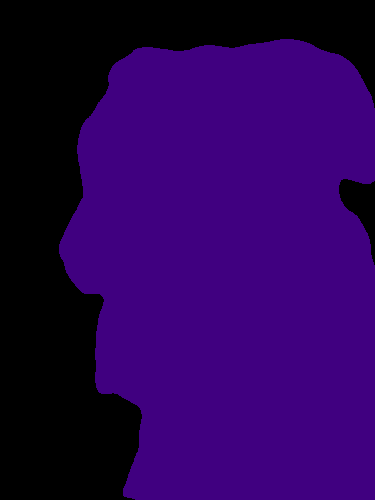} &
\includegraphics[width=0.075\textwidth,height=0.50in,keepaspectratio]{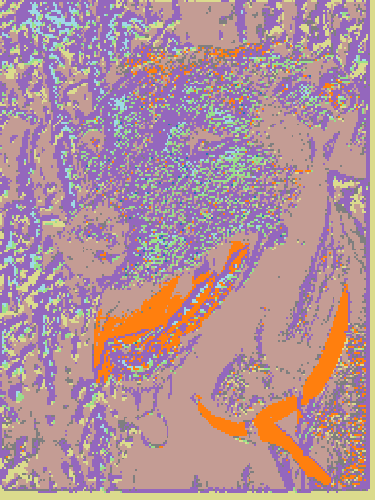} &
\includegraphics[width=0.075\textwidth,height=0.50in,keepaspectratio]{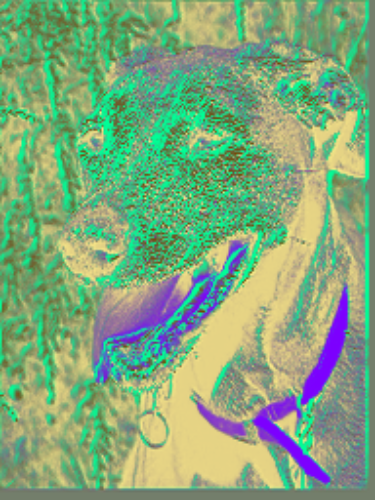} &
\includegraphics[width=0.075\textwidth,height=0.50in,keepaspectratio]{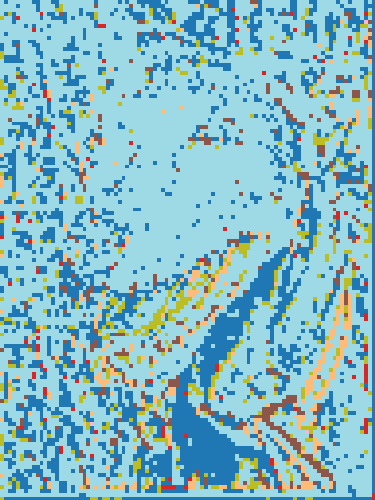} &
\includegraphics[width=0.075\textwidth,height=0.50in,keepaspectratio]{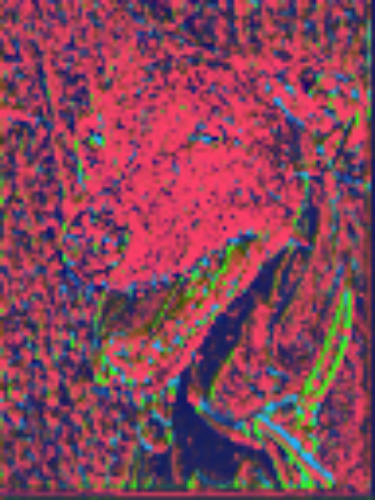} &
\includegraphics[width=0.075\textwidth,height=0.50in,keepaspectratio]{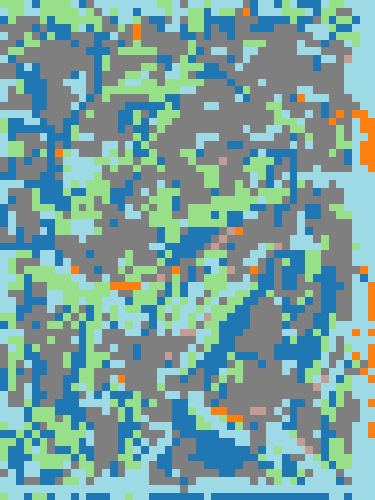} &
\includegraphics[width=0.075\textwidth,height=0.50in,keepaspectratio]{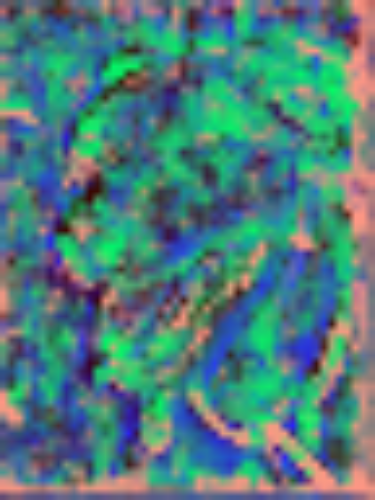} &
\includegraphics[width=0.075\textwidth,height=0.50in,keepaspectratio]{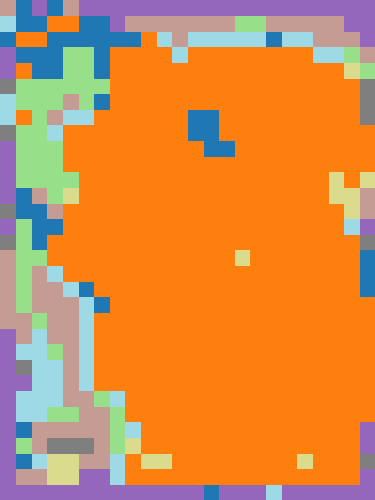} &
\includegraphics[width=0.075\textwidth,height=0.50in,keepaspectratio]{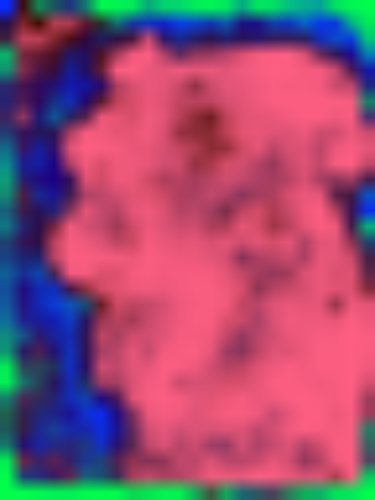} \\
\cmidrule(lr){1-12}

\multirow{2}{*}{\makebox[0.06\textwidth][c]{ADE20K}} &
\includegraphics[width=0.075\textwidth,height=0.50in,keepaspectratio]{figures/images_please_interp_ade20k_unet_good_sample_1_input.png} &
\includegraphics[width=0.075\textwidth,height=0.50in,keepaspectratio]{figures/images_please_interp_ade20k_unet_good_sample_1_ground_truth.png} &
\includegraphics[width=0.075\textwidth,height=0.50in,keepaspectratio]{figures/images_please_interp_ade20k_unet_good_sample_1_prediction.png} &
\includegraphics[width=0.075\textwidth,height=0.50in,keepaspectratio]{figures/images_please_interp_ade20k_unet_good_sample_1_encoder_1_argmax.png} &
\includegraphics[width=0.075\textwidth,height=0.50in,keepaspectratio]{figures/images_please_interp_ade20k_unet_good_sample_1_encoder_1_pca.png} &
\includegraphics[width=0.075\textwidth,height=0.50in,keepaspectratio]{figures/images_please_interp_ade20k_unet_good_sample_1_encoder_2_argmax.png} &
\includegraphics[width=0.075\textwidth,height=0.50in,keepaspectratio]{figures/images_please_interp_ade20k_unet_good_sample_1_encoder_2_pca.png} &
\includegraphics[width=0.075\textwidth,height=0.50in,keepaspectratio]{figures/images_please_interp_ade20k_unet_good_sample_1_encoder_3_argmax.png} &
\includegraphics[width=0.075\textwidth,height=0.50in,keepaspectratio]{figures/images_please_interp_ade20k_unet_good_sample_1_encoder_3_pca.png} &
\includegraphics[width=0.075\textwidth,height=0.50in,keepaspectratio]{figures/images_please_interp_ade20k_unet_good_sample_1_encoder_4_argmax.png} &
\includegraphics[width=0.075\textwidth,height=0.50in,keepaspectratio]{figures/images_please_interp_ade20k_unet_good_sample_1_encoder_4_pca.png} \\
&
\includegraphics[width=0.075\textwidth,height=0.50in,keepaspectratio]{figures/images_please_interp_ade20k_unet_good_sample_3_input.png} &
\includegraphics[width=0.075\textwidth,height=0.50in,keepaspectratio]{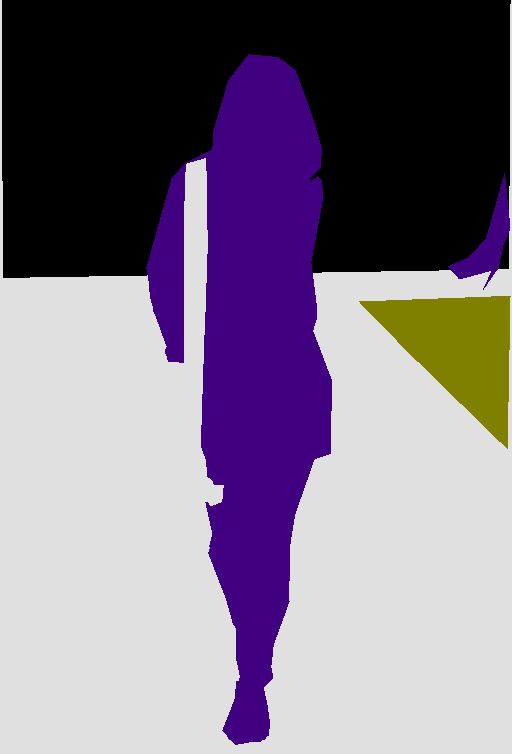} &
\includegraphics[width=0.075\textwidth,height=0.50in,keepaspectratio]{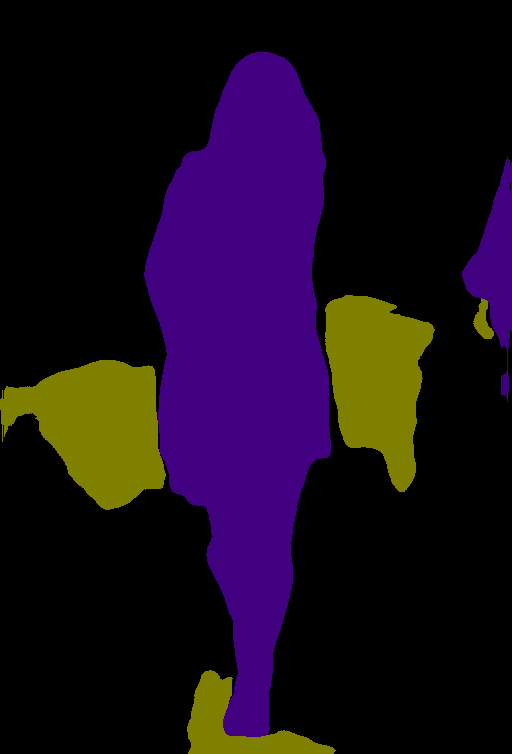} &
\includegraphics[width=0.075\textwidth,height=0.50in,keepaspectratio]{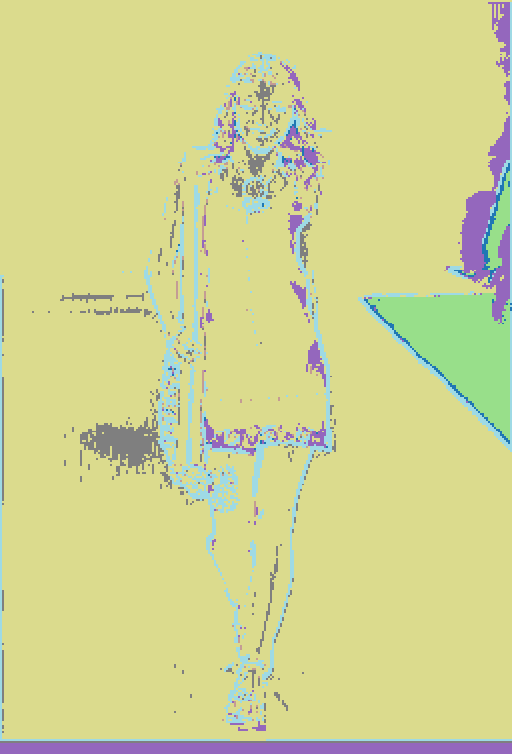} &
\includegraphics[width=0.075\textwidth,height=0.50in,keepaspectratio]{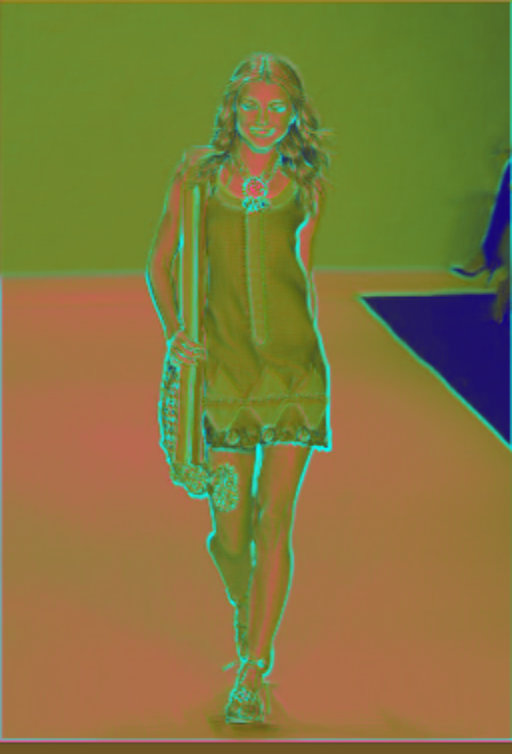} &
\includegraphics[width=0.075\textwidth,height=0.50in,keepaspectratio]{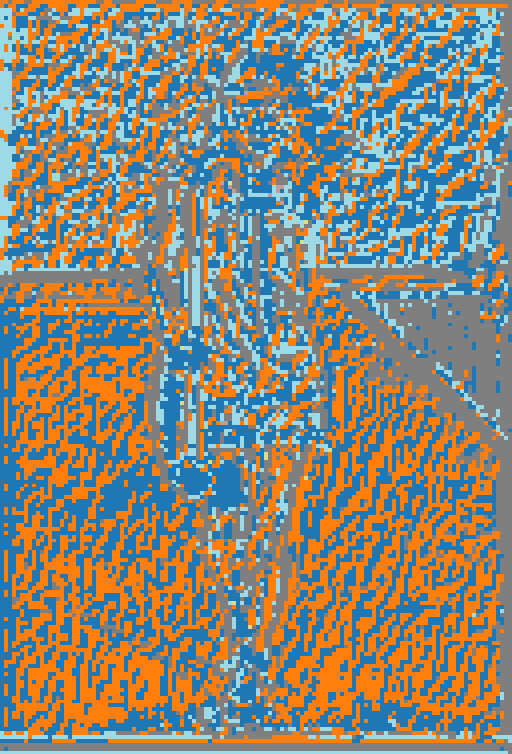} &
\includegraphics[width=0.075\textwidth,height=0.50in,keepaspectratio]{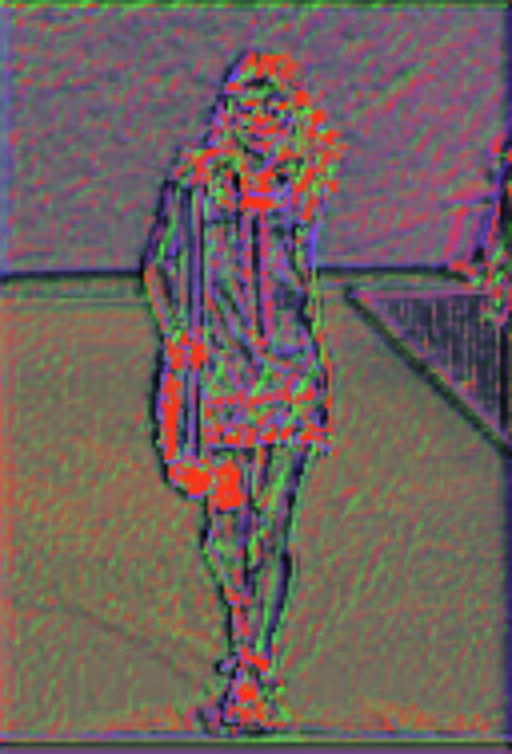} &
\includegraphics[width=0.075\textwidth,height=0.50in,keepaspectratio]{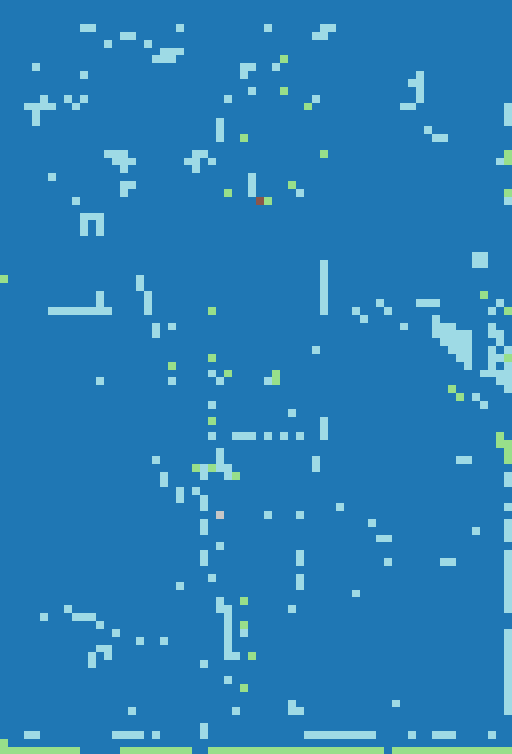} &
\includegraphics[width=0.075\textwidth,height=0.50in,keepaspectratio]{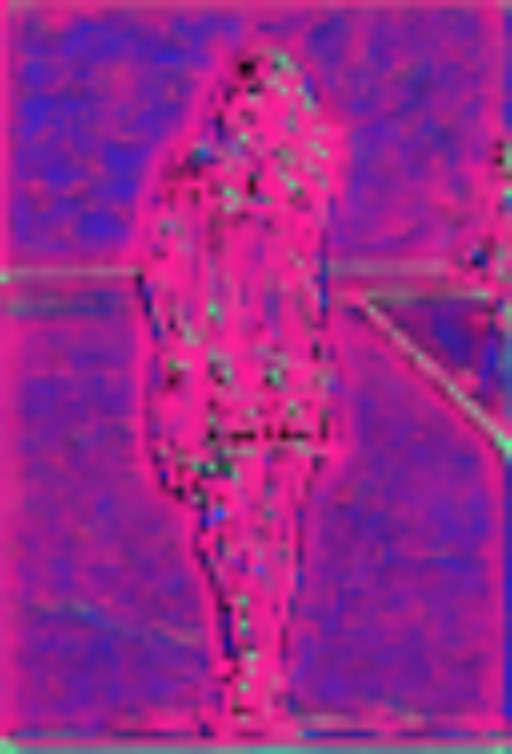} &
\includegraphics[width=0.075\textwidth,height=0.50in,keepaspectratio]{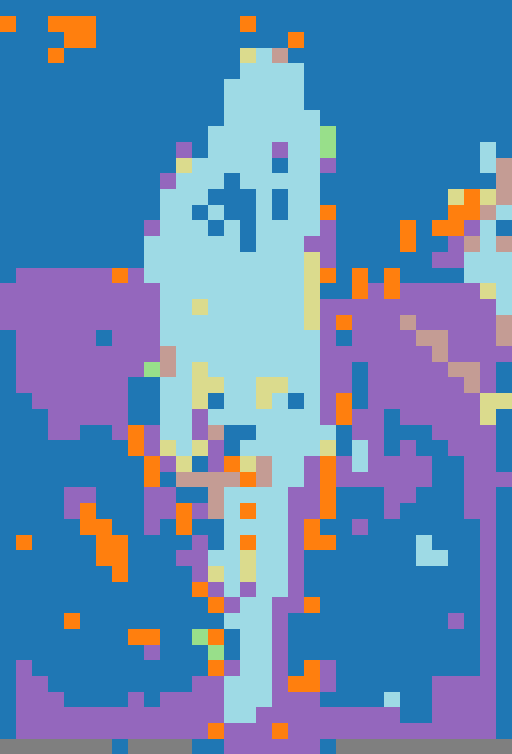} &
\includegraphics[width=0.075\textwidth,height=0.50in,keepaspectratio]{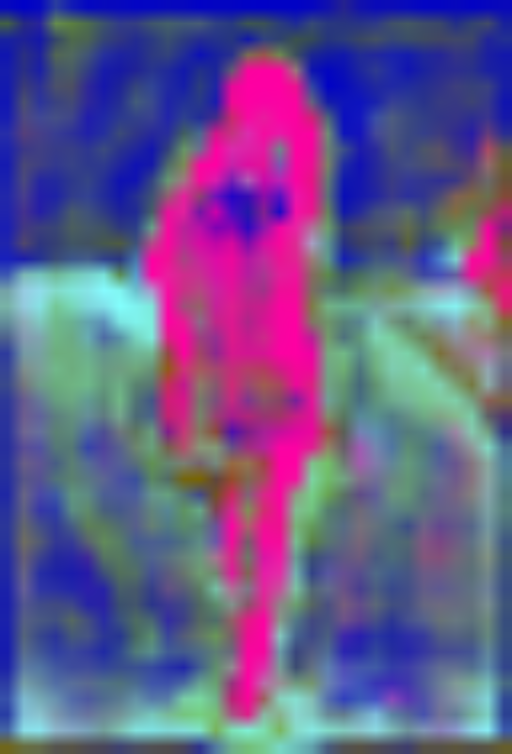} \\

\bottomrule
\end{tabular}%
}
\label{tab:please_good_interpretability_visuals}
\end{table*}

\begin{table*}[!htbp]
\centering
\scriptsize
\setlength{\tabcolsep}{1.5pt}

\caption{Interpretability visualizations for bad segmentation predictions, selected from the bottom 5\% of examples by mIoU for each dataset. Each dataset is shown with two representative examples. CUB/DUTS/ECSSD use U-Net, while VOC and ADE20K use Semantic U-Net. Columns show the input, ground truth, prediction, and argmax/PCA visualizations for encoder layers 1--4.}
\vspace{10pt}

\resizebox{\textwidth}{!}{%
\begin{tabular}{cccccccccccc}
\toprule
Dataset & Input & GT & Prediction & E1 argmax & E1 PCA & E2 argmax & E2 PCA & E3 argmax & E3 PCA & E4 argmax & E4 PCA \\
\midrule

\multirow{2}{*}{\makebox[0.06\textwidth][c]{CUB}} &
\includegraphics[width=0.075\textwidth,height=0.50in,keepaspectratio]{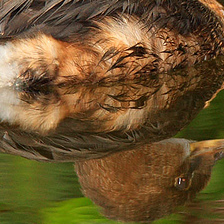} &
\includegraphics[width=0.075\textwidth,height=0.50in,keepaspectratio]{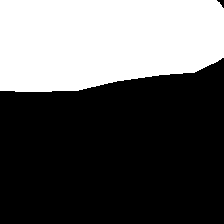} &
\includegraphics[width=0.075\textwidth,height=0.50in,keepaspectratio]{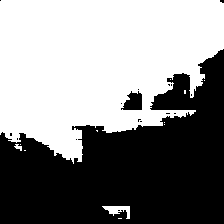} &
\includegraphics[width=0.075\textwidth,height=0.50in,keepaspectratio]{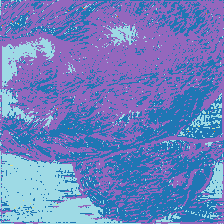} &
\includegraphics[width=0.075\textwidth,height=0.50in,keepaspectratio]{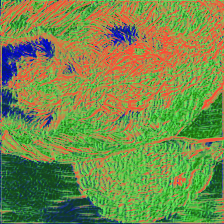} &
\includegraphics[width=0.075\textwidth,height=0.50in,keepaspectratio]{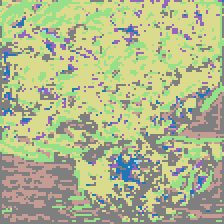} &
\includegraphics[width=0.075\textwidth,height=0.50in,keepaspectratio]{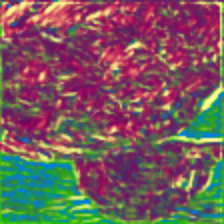} &
\includegraphics[width=0.075\textwidth,height=0.50in,keepaspectratio]{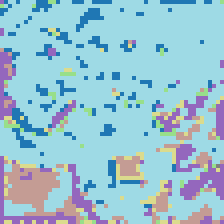} &
\includegraphics[width=0.075\textwidth,height=0.50in,keepaspectratio]{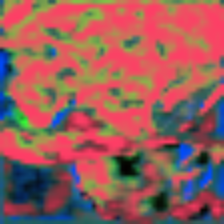} &
\includegraphics[width=0.075\textwidth,height=0.50in,keepaspectratio]{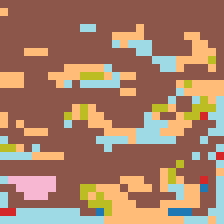} &
\includegraphics[width=0.075\textwidth,height=0.50in,keepaspectratio]{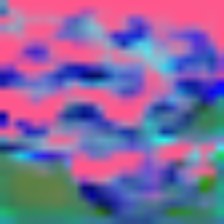} \\
&
\includegraphics[width=0.075\textwidth,height=0.50in,keepaspectratio]{figures/images_please_interp_cub_unet_bad_sample_2_input.png} &
\includegraphics[width=0.075\textwidth,height=0.50in,keepaspectratio]{figures/images_please_interp_cub_unet_bad_sample_2_ground_truth.png} &
\includegraphics[width=0.075\textwidth,height=0.50in,keepaspectratio]{figures/images_please_interp_cub_unet_bad_sample_2_prediction.png} &
\includegraphics[width=0.075\textwidth,height=0.50in,keepaspectratio]{figures/images_please_interp_cub_unet_bad_sample_2_encoder_1_argmax.png} &
\includegraphics[width=0.075\textwidth,height=0.50in,keepaspectratio]{figures/images_please_interp_cub_unet_bad_sample_2_encoder_1_pca.png} &
\includegraphics[width=0.075\textwidth,height=0.50in,keepaspectratio]{figures/images_please_interp_cub_unet_bad_sample_2_encoder_2_argmax.png} &
\includegraphics[width=0.075\textwidth,height=0.50in,keepaspectratio]{figures/images_please_interp_cub_unet_bad_sample_2_encoder_2_pca.png} &
\includegraphics[width=0.075\textwidth,height=0.50in,keepaspectratio]{figures/images_please_interp_cub_unet_bad_sample_2_encoder_3_argmax.png} &
\includegraphics[width=0.075\textwidth,height=0.50in,keepaspectratio]{figures/images_please_interp_cub_unet_bad_sample_2_encoder_3_pca.png} &
\includegraphics[width=0.075\textwidth,height=0.50in,keepaspectratio]{figures/images_please_interp_cub_unet_bad_sample_2_encoder_4_argmax.png} &
\includegraphics[width=0.075\textwidth,height=0.50in,keepaspectratio]{figures/images_please_interp_cub_unet_bad_sample_2_encoder_4_pca.png} \\
\cmidrule(lr){1-12}

\multirow{2}{*}{\makebox[0.06\textwidth][c]{DUTS}} &
\includegraphics[width=0.075\textwidth,height=0.50in,keepaspectratio]{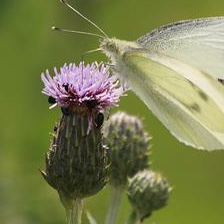} &
\includegraphics[width=0.075\textwidth,height=0.50in,keepaspectratio]{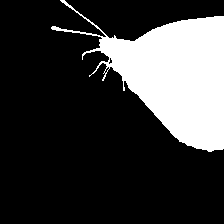} &
\includegraphics[width=0.075\textwidth,height=0.50in,keepaspectratio]{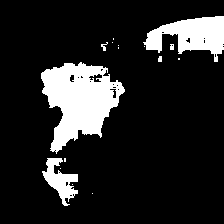} &
\includegraphics[width=0.075\textwidth,height=0.50in,keepaspectratio]{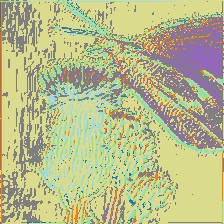} &
\includegraphics[width=0.075\textwidth,height=0.50in,keepaspectratio]{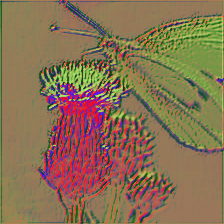} &
\includegraphics[width=0.075\textwidth,height=0.50in,keepaspectratio]{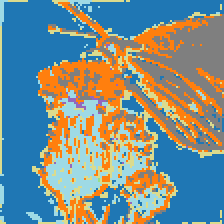} &
\includegraphics[width=0.075\textwidth,height=0.50in,keepaspectratio]{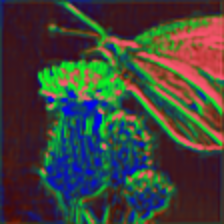} &
\includegraphics[width=0.075\textwidth,height=0.50in,keepaspectratio]{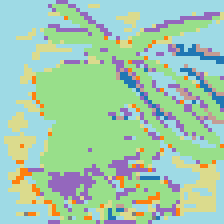} &
\includegraphics[width=0.075\textwidth,height=0.50in,keepaspectratio]{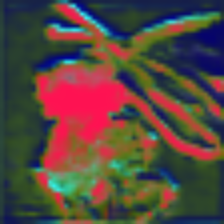} &
\includegraphics[width=0.075\textwidth,height=0.50in,keepaspectratio]{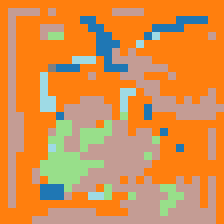} &
\includegraphics[width=0.075\textwidth,height=0.50in,keepaspectratio]{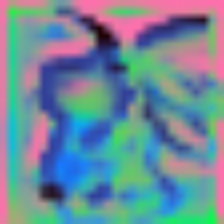} \\
&
\includegraphics[width=0.075\textwidth,height=0.50in,keepaspectratio]{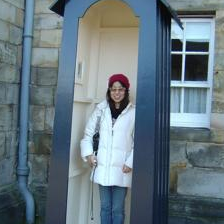} &
\includegraphics[width=0.075\textwidth,height=0.50in,keepaspectratio]{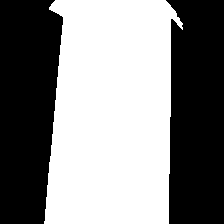} &
\includegraphics[width=0.075\textwidth,height=0.50in,keepaspectratio]{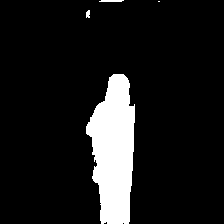} &
\includegraphics[width=0.075\textwidth,height=0.50in,keepaspectratio]{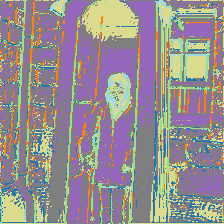} &
\includegraphics[width=0.075\textwidth,height=0.50in,keepaspectratio]{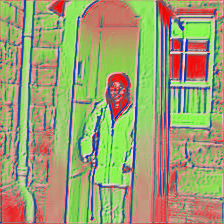} &
\includegraphics[width=0.075\textwidth,height=0.50in,keepaspectratio]{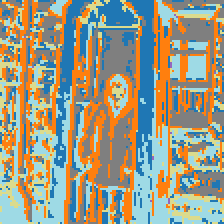} &
\includegraphics[width=0.075\textwidth,height=0.50in,keepaspectratio]{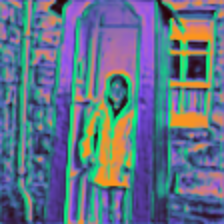} &
\includegraphics[width=0.075\textwidth,height=0.50in,keepaspectratio]{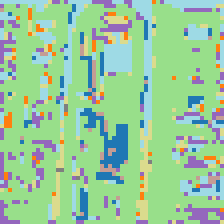} &
\includegraphics[width=0.075\textwidth,height=0.50in,keepaspectratio]{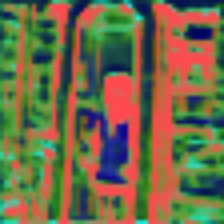} &
\includegraphics[width=0.075\textwidth,height=0.50in,keepaspectratio]{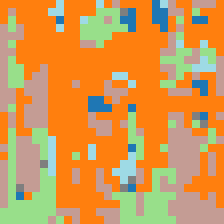} &
\includegraphics[width=0.075\textwidth,height=0.50in,keepaspectratio]{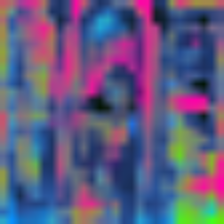} \\
\cmidrule(lr){1-12}

\multirow{2}{*}{\makebox[0.06\textwidth][c]{ECSSD}} &
\includegraphics[width=0.075\textwidth,height=0.50in,keepaspectratio]{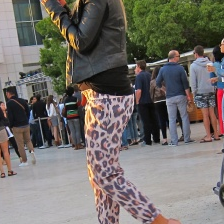} &
\includegraphics[width=0.075\textwidth,height=0.50in,keepaspectratio]{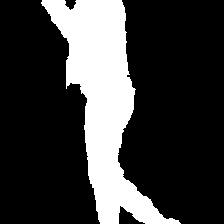} &
\includegraphics[width=0.075\textwidth,height=0.50in,keepaspectratio]{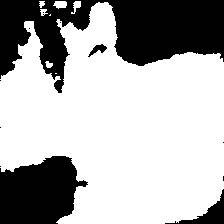} &
\includegraphics[width=0.075\textwidth,height=0.50in,keepaspectratio]{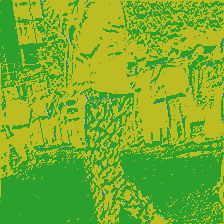} &
\includegraphics[width=0.075\textwidth,height=0.50in,keepaspectratio]{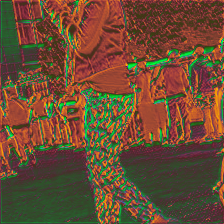} &
\includegraphics[width=0.075\textwidth,height=0.50in,keepaspectratio]{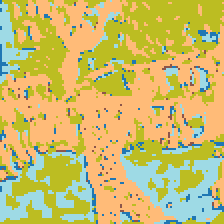} &
\includegraphics[width=0.075\textwidth,height=0.50in,keepaspectratio]{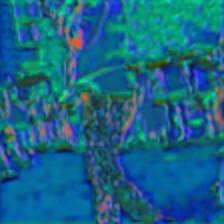} &
\includegraphics[width=0.075\textwidth,height=0.50in,keepaspectratio]{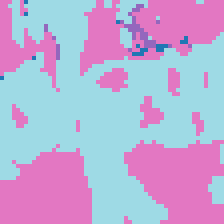} &
\includegraphics[width=0.075\textwidth,height=0.50in,keepaspectratio]{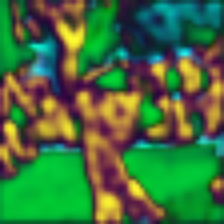} &
\includegraphics[width=0.075\textwidth,height=0.50in,keepaspectratio]{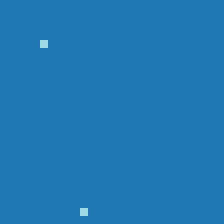} &
\includegraphics[width=0.075\textwidth,height=0.50in,keepaspectratio]{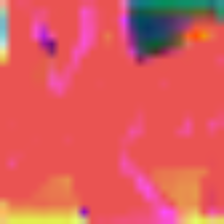} \\
&
\includegraphics[width=0.075\textwidth,height=0.50in,keepaspectratio]{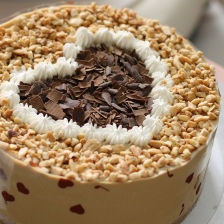} &
\includegraphics[width=0.075\textwidth,height=0.50in,keepaspectratio]{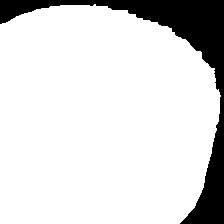} &
\includegraphics[width=0.075\textwidth,height=0.50in,keepaspectratio]{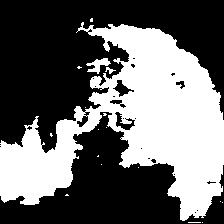} &
\includegraphics[width=0.075\textwidth,height=0.50in,keepaspectratio]{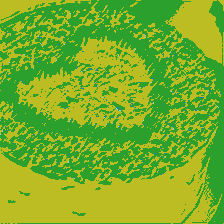} &
\includegraphics[width=0.075\textwidth,height=0.50in,keepaspectratio]{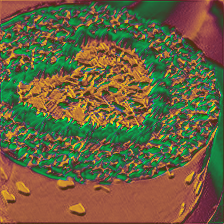} &
\includegraphics[width=0.075\textwidth,height=0.50in,keepaspectratio]{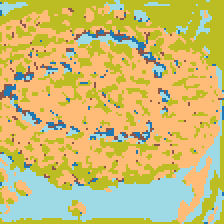} &
\includegraphics[width=0.075\textwidth,height=0.50in,keepaspectratio]{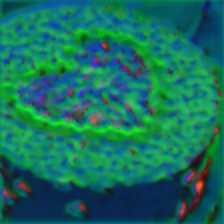} &
\includegraphics[width=0.075\textwidth,height=0.50in,keepaspectratio]{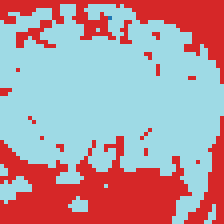} &
\includegraphics[width=0.075\textwidth,height=0.50in,keepaspectratio]{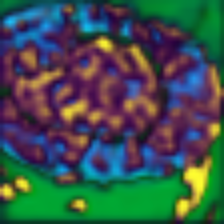} &
\includegraphics[width=0.075\textwidth,height=0.50in,keepaspectratio]{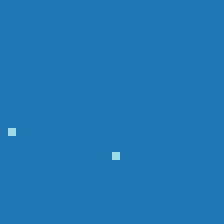} &
\includegraphics[width=0.075\textwidth,height=0.50in,keepaspectratio]{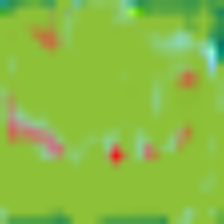} \\
\cmidrule(lr){1-12}

\multirow{2}{*}{\makebox[0.06\textwidth][c]{VOC}} &
\includegraphics[width=0.075\textwidth,height=0.50in,keepaspectratio]{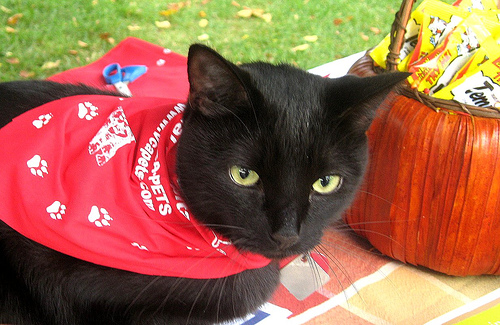} &
\includegraphics[width=0.075\textwidth,height=0.50in,keepaspectratio]{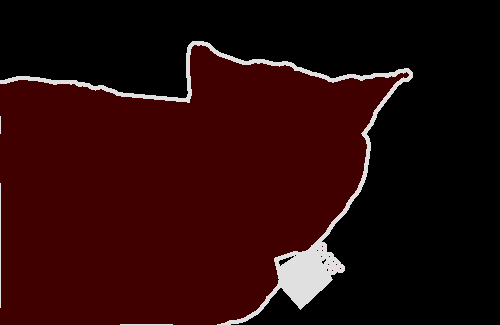} &
\includegraphics[width=0.075\textwidth,height=0.50in,keepaspectratio]{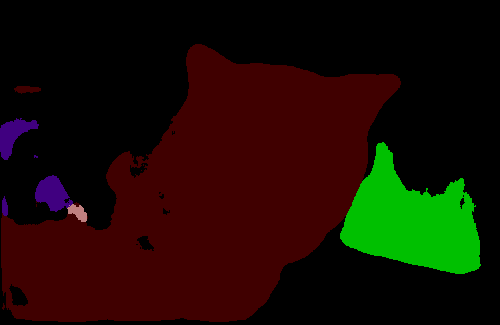} &
\includegraphics[width=0.075\textwidth,height=0.50in,keepaspectratio]{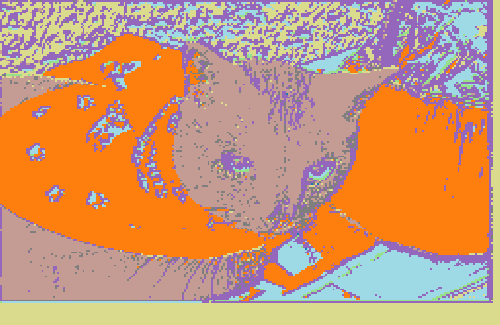} &
\includegraphics[width=0.075\textwidth,height=0.50in,keepaspectratio]{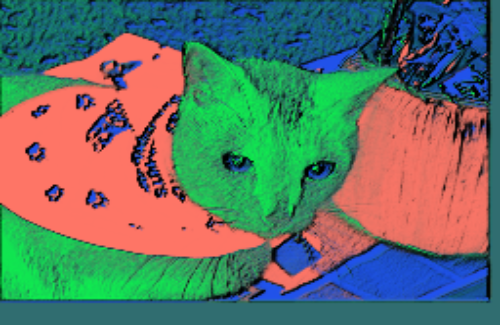} &
\includegraphics[width=0.075\textwidth,height=0.50in,keepaspectratio]{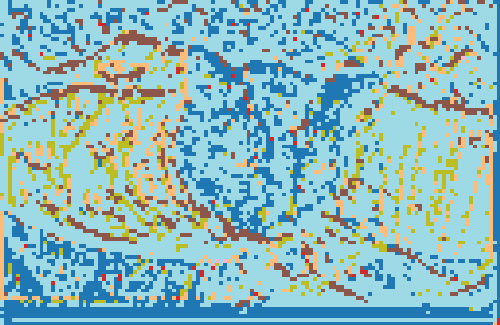} &
\includegraphics[width=0.075\textwidth,height=0.50in,keepaspectratio]{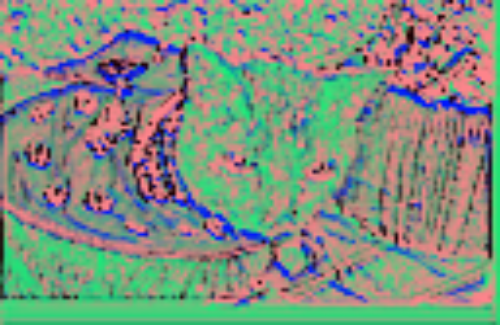} &
\includegraphics[width=0.075\textwidth,height=0.50in,keepaspectratio]{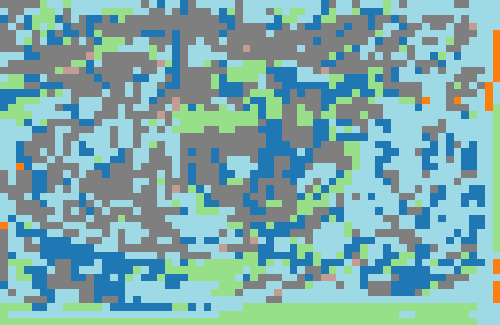} &
\includegraphics[width=0.075\textwidth,height=0.50in,keepaspectratio]{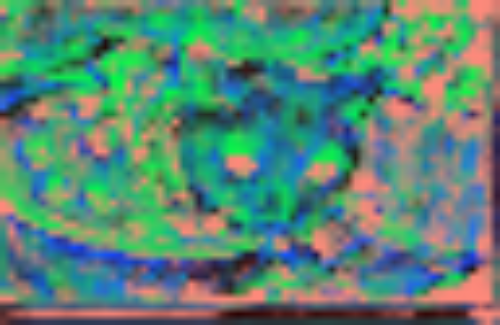} &
\includegraphics[width=0.075\textwidth,height=0.50in,keepaspectratio]{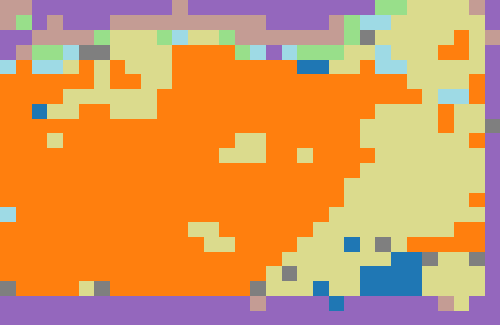} &
\includegraphics[width=0.075\textwidth,height=0.50in,keepaspectratio]{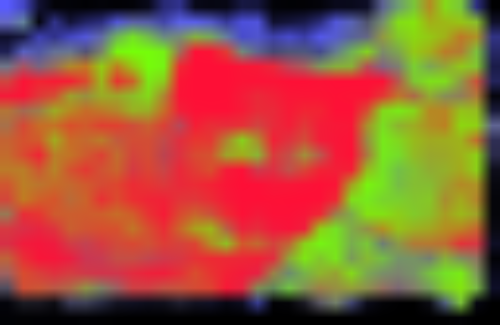} \\
&
\includegraphics[width=0.075\textwidth,height=0.50in,keepaspectratio]{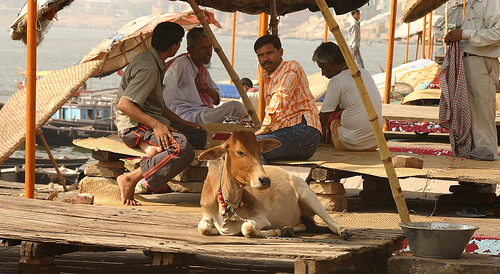} &
\includegraphics[width=0.075\textwidth,height=0.50in,keepaspectratio]{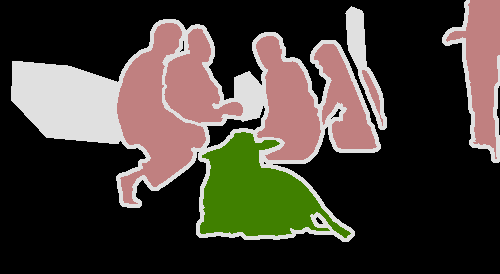} &
\includegraphics[width=0.075\textwidth,height=0.50in,keepaspectratio]{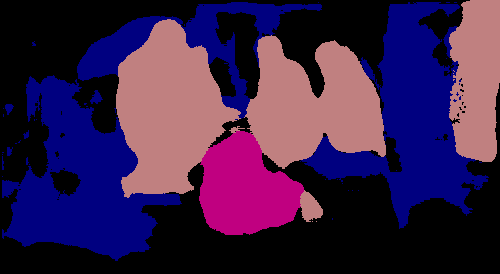} &
\includegraphics[width=0.075\textwidth,height=0.50in,keepaspectratio]{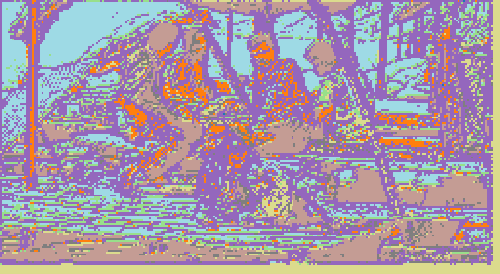} &
\includegraphics[width=0.075\textwidth,height=0.50in,keepaspectratio]{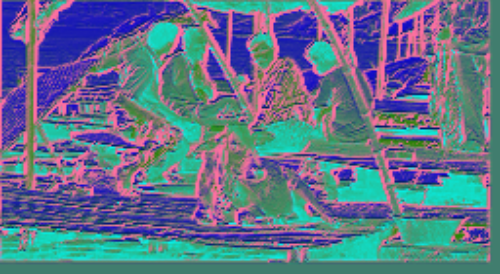} &
\includegraphics[width=0.075\textwidth,height=0.50in,keepaspectratio]{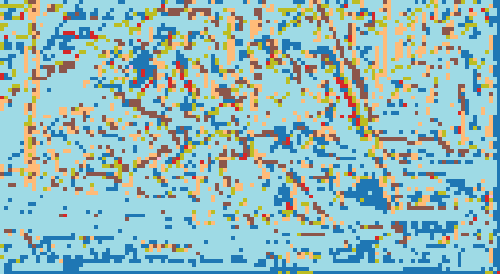} &
\includegraphics[width=0.075\textwidth,height=0.50in,keepaspectratio]{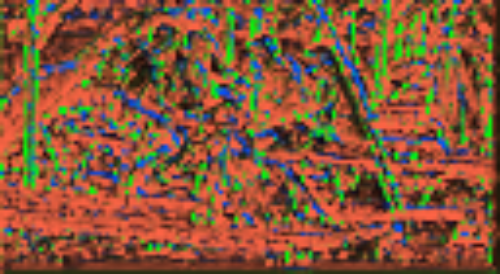} &
\includegraphics[width=0.075\textwidth,height=0.50in,keepaspectratio]{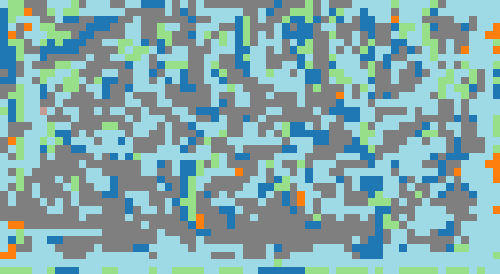} &
\includegraphics[width=0.075\textwidth,height=0.50in,keepaspectratio]{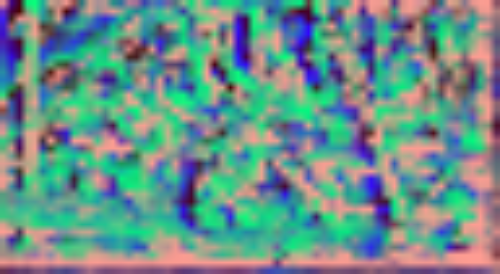} &
\includegraphics[width=0.075\textwidth,height=0.50in,keepaspectratio]{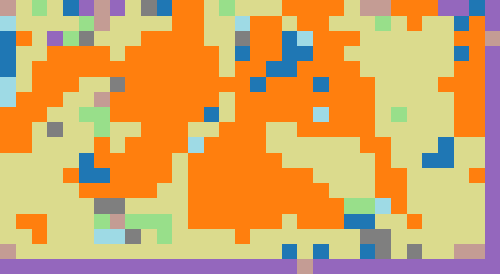} &
\includegraphics[width=0.075\textwidth,height=0.50in,keepaspectratio]{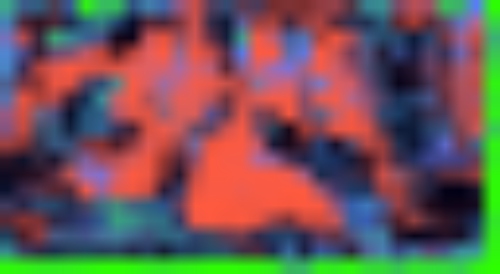} \\
\cmidrule(lr){1-12}

\multirow{2}{*}{\makebox[0.06\textwidth][c]{ADE20K}} &
\includegraphics[width=0.075\textwidth,height=0.50in,keepaspectratio]{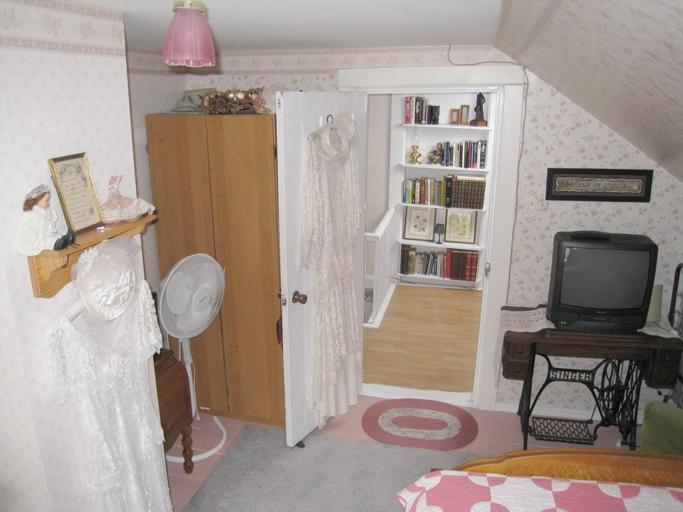} &
\includegraphics[width=0.075\textwidth,height=0.50in,keepaspectratio]{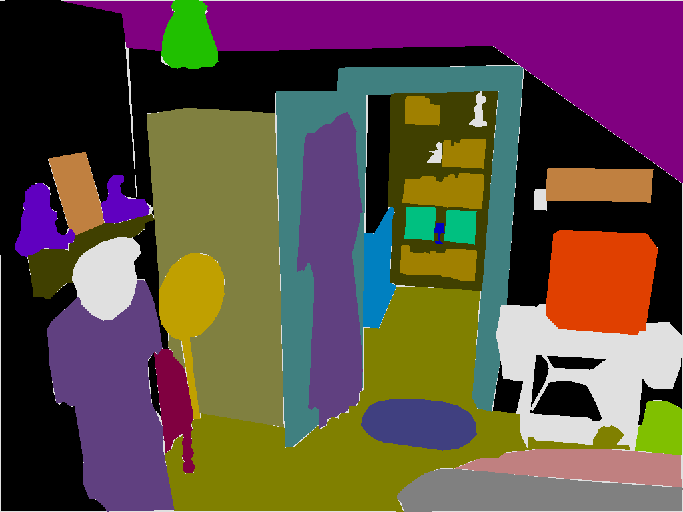} &
\includegraphics[width=0.075\textwidth,height=0.50in,keepaspectratio]{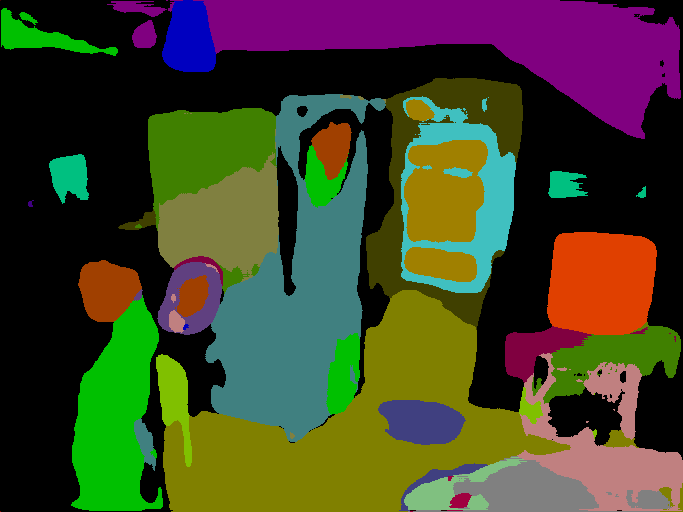} &
\includegraphics[width=0.075\textwidth,height=0.50in,keepaspectratio]{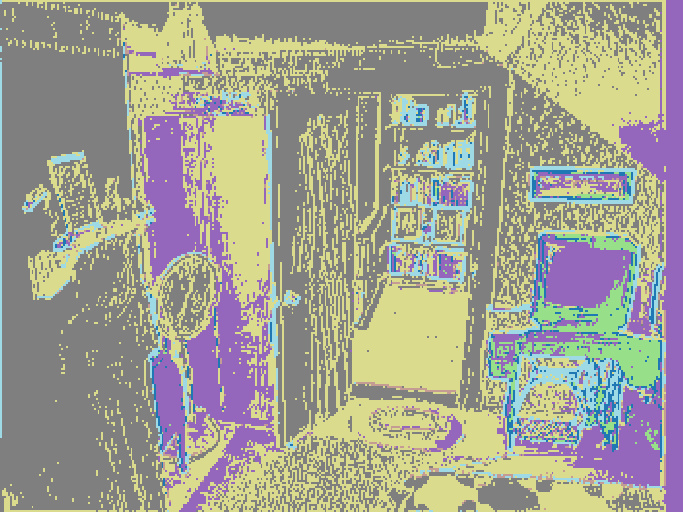} &
\includegraphics[width=0.075\textwidth,height=0.50in,keepaspectratio]{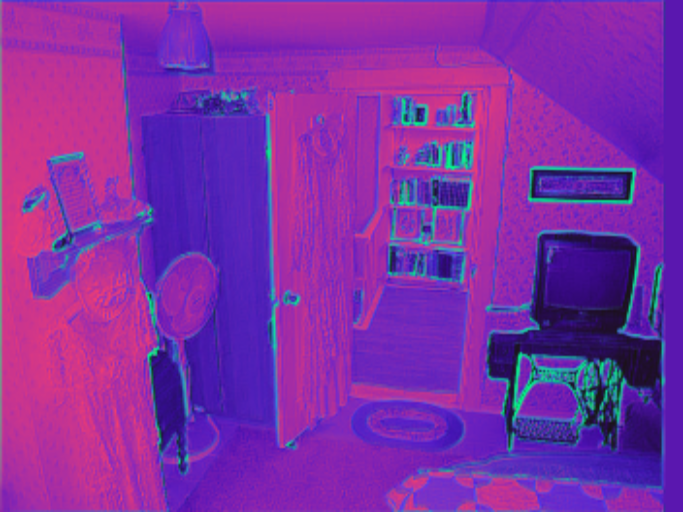} &
\includegraphics[width=0.075\textwidth,height=0.50in,keepaspectratio]{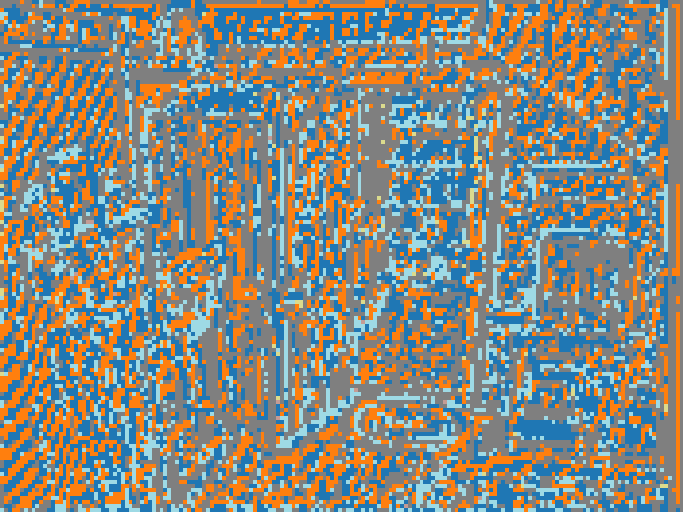} &
\includegraphics[width=0.075\textwidth,height=0.50in,keepaspectratio]{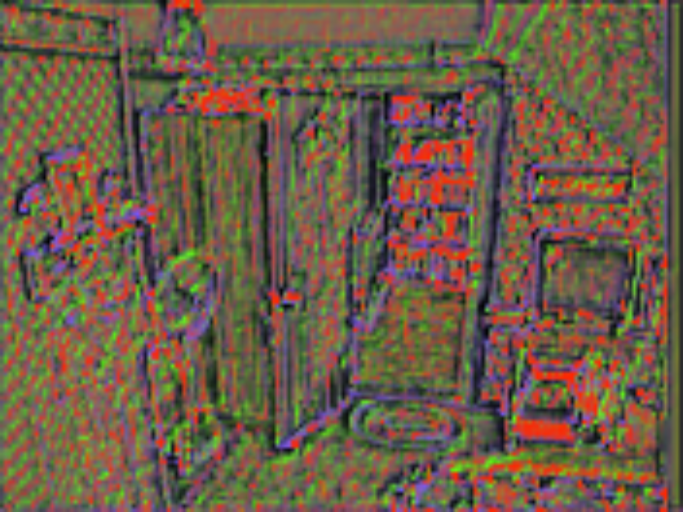} &
\includegraphics[width=0.075\textwidth,height=0.50in,keepaspectratio]{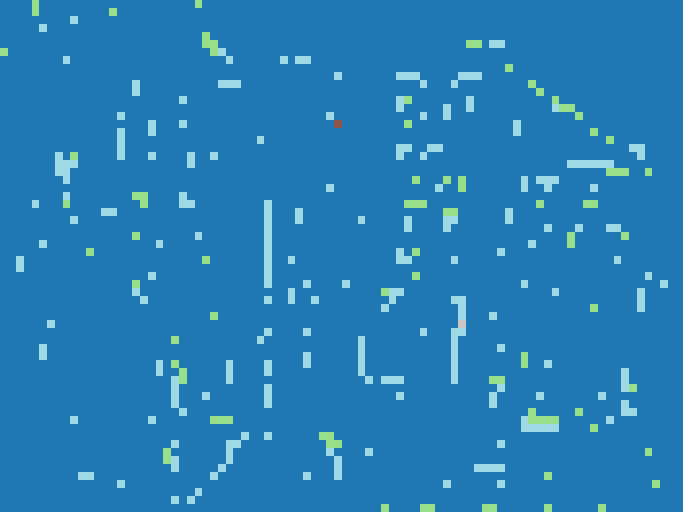} &
\includegraphics[width=0.075\textwidth,height=0.50in,keepaspectratio]{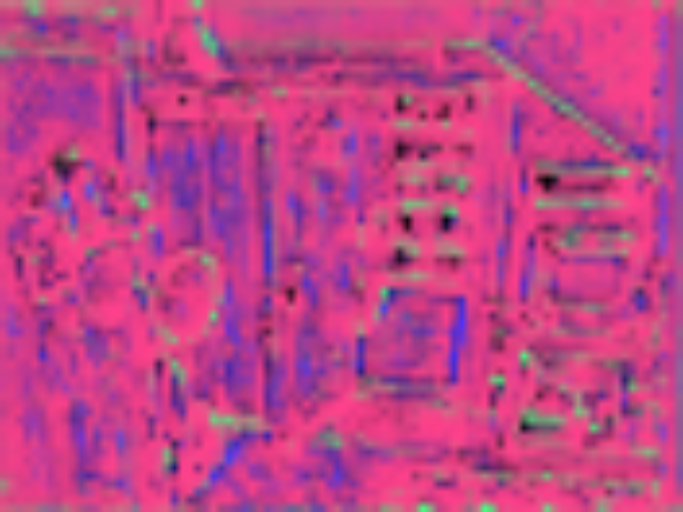} &
\includegraphics[width=0.075\textwidth,height=0.50in,keepaspectratio]{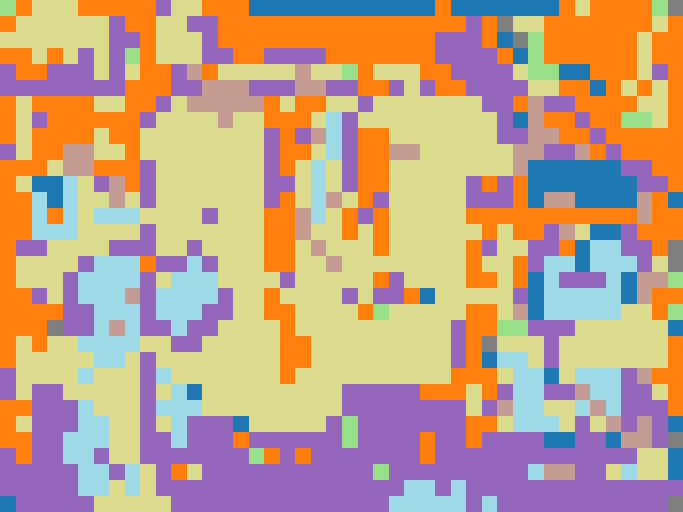} &
\includegraphics[width=0.075\textwidth,height=0.50in,keepaspectratio]{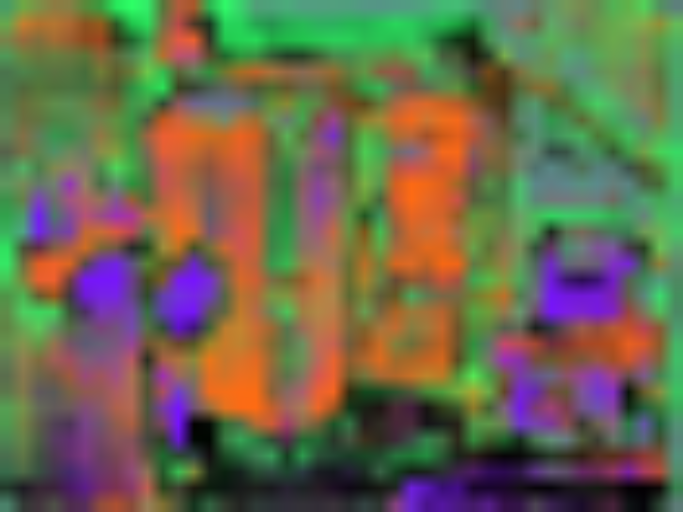} \\
&
\includegraphics[width=0.075\textwidth,height=0.50in,keepaspectratio]{figures/images_please_interp_ade20k_unet_bad_sample_3_input.png} &
\includegraphics[width=0.075\textwidth,height=0.50in,keepaspectratio]{figures/images_please_interp_ade20k_unet_bad_sample_3_ground_truth.png} &
\includegraphics[width=0.075\textwidth,height=0.50in,keepaspectratio]{figures/images_please_interp_ade20k_unet_bad_sample_3_prediction.png} &
\includegraphics[width=0.075\textwidth,height=0.50in,keepaspectratio]{figures/images_please_interp_ade20k_unet_bad_sample_3_encoder_1_argmax.png} &
\includegraphics[width=0.075\textwidth,height=0.50in,keepaspectratio]{figures/images_please_interp_ade20k_unet_bad_sample_3_encoder_1_pca.png} &
\includegraphics[width=0.075\textwidth,height=0.50in,keepaspectratio]{figures/images_please_interp_ade20k_unet_bad_sample_3_encoder_2_argmax.png} &
\includegraphics[width=0.075\textwidth,height=0.50in,keepaspectratio]{figures/images_please_interp_ade20k_unet_bad_sample_3_encoder_2_pca.png} &
\includegraphics[width=0.075\textwidth,height=0.50in,keepaspectratio]{figures/images_please_interp_ade20k_unet_bad_sample_3_encoder_3_argmax.png} &
\includegraphics[width=0.075\textwidth,height=0.50in,keepaspectratio]{figures/images_please_interp_ade20k_unet_bad_sample_3_encoder_3_pca.png} &
\includegraphics[width=0.075\textwidth,height=0.50in,keepaspectratio]{figures/images_please_interp_ade20k_unet_bad_sample_3_encoder_4_argmax.png} &
\includegraphics[width=0.075\textwidth,height=0.50in,keepaspectratio]{figures/images_please_interp_ade20k_unet_bad_sample_3_encoder_4_pca.png} \\

\bottomrule
\end{tabular}%
}
\label{tab:please_bad_interpretability_visuals}
\end{table*}

\begin{figure*}[t]
    \centering
         \begin{lstlisting}[style=pytorchcode]
import torch
import torch.nn as nn
import torch.nn.functional as F

class BASE(nn.Module):
    def __init__(self, k, phi):
        # k:   grid size for block average pooling
        # phi: channel excitation operator
        
        super().__init__()
        self.k = k
        self.phi = phi
        self.avg_pool = nn.AdaptiveAvgPool2d(k)
 
    def forward(self, x):
        B, C, H, W = x.size()
        z_block = self.avg_pool(x)                    
        w_block = self.phi(z_block.view(B, C, self.k ** 2)) 
        w_block = w_block.view(B, C, self.k, self.k)
        w = F.interpolate(w_block, size=(H, W), mode='nearest')  
        return x * w


class SPASE(nn.Module):
    def __init__(self, phi):
        # phi: channel excitation operator
        
        super().__init__()
        self.phi = phi
 
    def forward(self, x, S):    
        B, C, H, W = x.size()
        K = S.max().item() + 1
        S_onehot = F.one_hot(S, K).float()          
        S_onehot = S_onehot.permute(0, 3, 1, 2)      
        counts = S_onehot.sum(dim=(2, 3))            
        z_sp = torch.einsum('bchw,bnhw->bcn', x, S_onehot)                 
        z_sp = z_sp / counts.unsqueeze(1)                  
        w_sp = self.phi(z_sp)                 
        w = torch.einsum('bcn,bnhw->bchw', w_sp, S_onehot)   
        return x * w         


class MAPSE(nn.Module):
    def __init__(self, phi, eps=1e-6):
        # phi: channel excitation operator

        super().__init__()
        self.phi = phi
        self.eps = eps

    def forward(self, x, M):
        B, C, H, W = x.size()
        M = F.interpolate(M, size=(H, W), mode='bilinear', align_corners=False)
        mass = M.sum(dim=(2, 3)).clamp_min(self.eps)
        z_map = torch.einsum('bchw,bkhw->bkc', x, M) / mass.unsqueeze(-1)
        w_map = self.phi(z_map)
        w = torch.einsum('bkc,bkhw->bchw', w_map, M)
        return x * w


class PLEASE(nn.Module):
    def __init__(self, c, K, phi, eps=1e-6):
        # c:   number of input channels
        # K:   number of soft regions
        # phi: channel excitation 
        
        super().__init__()
        self.phi = phi
        self.eps = eps
        r = max(1, c // 4)
        self.pi = nn.Sequential(
            nn.Conv2d(c, r, 1), nn.ReLU(), nn.Conv2d(r, K, 1), nn.Softmax(dim=1)
        )
 
    def forward(self, x):
        P = self.pi(x)
        mass = P.sum(dim=(2, 3)).clamp_min(self.eps)
        z_soft = torch.einsum('bchw,bkhw->bkc', x, P) / mass.unsqueeze(-1)
        w_soft = self.phi(z_soft)
        w = torch.einsum('bkc,bkhw->bchw', w_soft, P)
        return x * w
        \end{lstlisting}

    \label{fig:code_modules}

\vspace{10pt}
\caption{
    PyTorch-style implementations of the proposed
    region-aware \se modules. Each module supports
    both MLP and 1D-convolution excitation through
    the shared excitation operator $\phi(\cdot)$. In the code above, we assume $X \in \mathbb{R}^{B \times C \times H \times W}$ and $S \in [N]^{B \times H \times W}$, where $B$ is the batch size.  These listings are a pedagogical simplification of the released implementation: they show the region-aware squeeze and excitation itself, but omit supporting details such as the renormalization of externally sourced region maps and the numerical guards described in Sec.~\ref{sec:impl}. Readers reproducing the modules should follow the accompanying demo code, which is the reference implementation.}
\label{fig:code-modules}
\end{figure*}

\clearpage
\end{document}